\documentclass[manuscript,screen]{acmart}
\usepackage{multirow}
\usepackage{graphics}
\usepackage{color}
\usepackage{amsthm,amsmath,amssymb}
\AtBeginDocument{%
  \providecommand\BibTeX{{%
    \normalfont B\kern-0.5em{\scshape i\kern-0.25em b}\kern-0.8em\TeX}}}

\setcopyright{acmcopyright}
\copyrightyear{2019}
\acmYear{2019}
\acmDOI{}

\begin{document}

\title{A Survey of Document Grounded Dialogue Systems (DGDS)}

\author{Longxuan Ma}
\email{lxma@ir.hit.edu.cn}
\orcid{0000-0003-1431-8485}
\affiliation{%
  \institution{Research Center for Social Computing and Information Retrieval, Harbin Institute of Technology}
  \streetaddress{92, Xidazhi Road, Nangang Qu}
  \city{Harbin}
  \state{Heilongjiang}
  \country{China}
}

\author{Wei-Nan Zhang}
\email{wnzhang@ir.hit.edu.cn}
\orcid{}
\affiliation{  
  \institution{Research Center for Social Computing and Information Retrieval, Harbin Institute of Technology}
  \streetaddress{92, Xidazhi Road, Nangang Qu}
  \city{Harbin}
  \state{Heilongjiang}
  \country{China}
}

\author{Mingda Li}
\email{mdli@ir.hit.edu.cn}
\orcid{}
\affiliation{%
  \institution{Research Center for Social Computing and Information Retrieval, Harbin Institute of Technology}
  \streetaddress{92, Xidazhi Road, Nangang Qu}
  \city{Harbin}
  \state{Heilongjiang}
  \country{China}
}

\author{Ting Liu}
\email{tliu@ir.hit.edu.cn}
\orcid{}
\affiliation{%
  \institution{Research Center for Social Computing and Information Retrieval, Harbin Institute of Technology}
  \streetaddress{92, Xidazhi Road, Nangang Qu}
  \city{Harbin}
  \state{Heilongjiang}
  \country{China}
}


\begin{abstract}
Dialogue system (DS) attracts great attention from industry and academia because of its wide application prospects. Researchers usually divide the DS according to the function. However, many conversations require the DS to switch between different functions. For example, movie discussion can change from chit-chat to QA, the conversational recommendation can transform from chit-chat to recommendation, etc. Therefore, classification according to functions may not be enough to help us appreciate the current development trend. We classify the DS based on background knowledge. Specifically, study the latest DS based on the unstructured document(s). We define Document Grounded Dialogue System (DGDS) as \textbf{the DS that the dialogues are centering on the given document(s)}. The DGDS can be used in scenarios such as talking over merchandise against product Manual, commenting on news reports, etc. We believe that extracting unstructured document(s) information is the future trend of the DS because a great amount of human knowledge lies in these document(s). The research of the DGDS not only possesses a broad application prospect but also facilitates AI to better understand human knowledge and natural language. We analyze the classification, architecture, datasets, models, and future development trends of the DGDS, hoping to help researchers in this field.
\end{abstract}



\maketitle

\section{Introduction}

For a long time, researchers have been devoting themselves to develop a Dialogue System (DS) that can communicate with human beings naturally. Early DS such as Eliza \cite{DBLP:journals/cacm/Weizenbaum66}, Parry \cite{DBLP:journals/ai/ColbyWH71}, and Alice \cite{DBLP:books/daglib/0072558} attempted to simulate human behaviors in conversations, and challenged various forms of the Turing Test \cite{turing2009computing}. They worked well but only in constrained environments, an open-domain DS remained an elusive task until recently \cite{DBLP:journals/corr/abs-1905-05709}. Then works focused on the task-oriented DS such as DARPA \citep{DBLP:conf/interspeech/WalkerABBGHLLNPPPPPRSSSW01,DBLP:conf/interspeech/WalkerRABGHLPPPPRSSS02,DBLP:conf/acl/WalkerPB01} arose, they performed well only within domains that have well-defined schemas. 

Although task-oriented DS and open-domain DS are originally developed for different purposes, \citet{DBLP:journals/ftir/GaoGL19} regarded both of them can be designed as an optimal decision-making process, whose goal is to maximize the expected reward. The reward of the former is easier to define and optimize than that of the latter. In past years, some researchers \cite{DBLP:journals/corr/DodgeGZBCMSW15,DBLP:conf/acl/AkasakiK17,DBLP:conf/aaai/YanDCZZL17,DBLP:conf/sigdial/ZhaoLLE17} have begun to explore technologies to integrate them. \citet{DBLP:journals/corr/DodgeGZBCMSW15} investigated $5$ different tasks (QA, Dialogue, Recommendation, etc.) with a Memory Network \cite{DBLP:journals/corr/WestonCB14}. \citet{DBLP:conf/acl/AkasakiK17} proposed a dataset to distinguish whether the user is having a chat or giving a request to the chatting machine. \citet{DBLP:conf/aaai/YanDCZZL17} presented a general solution to the task-oriented DS for online shopping. The goal is to help users complete various purchase-related tasks, such as searching products and answering questions, just like the dialogue between normal people. \citet{DBLP:conf/sigdial/ZhaoLLE17} proposed a task-oriented dialogue agent based on the encoder-decoder model with chatting capability. Subsequently, \citet{DBLP:conf/aaai/GhazvininejadBC18} presented a fully data-driven and knowledge-grounded neural conversation model aimed at producing more contentful responses without slot filling. These works represented steps that build end-to-end DS in scenarios beyond a single function.

In the past few years, a great effort has been made to develop virtual assistants such as Apple’s Siri\footnote{https://www.apple.com/ios/siri/}, Microsoft’s Cortana\footnote{https://www.microsoft.com/en-us/cortana/}, Amazon’s Alexa\footnote{https://developer.amazon.com/alexa/} and Google Assistant\footnote{https://assistant.google.com/}. These applications are capable of answering a wide range of questions on mobile devices. In addition to passively responding to user requests, they also proactively predict users' demands and provide in-time assistance such as reminding of an upcoming event or recommending a useful service without receiving explicit commands \cite{DBLP:journals/spm/Sarikaya17}. Meanwhile, social bots designed to meet the users' emotional needs showed great development potential. Since its launch in 2014, Microsoft's XiaoIce System\footnote{https://www.msXiaoIce.com/} has attracted millions of people on various topics for long time interlocution \cite{DBLP:journals/jzusc/ShumHL18,DBLP:journals/corr/abs-1812-08989}. Inspired by the Turing test, it is designed to test the ability of social bots to provide coherent, relevant, interesting and interactive communication and to keep user's participation within the possible range. In 2016, the Alexa Prize challenge was proposed to advance the research and development of social bots that are able to converse coherently and engagingly with humans on popular topics such as sports, politics, and entertainment, for at least $20$ minutes \cite{DBLP:journals/corr/abs-1801-03604}. The virtual assistants and social bots usually consist a natural hierarchy: a top-level process selecting what the agent is about to activate for a particular subtask (e.g., answering a question, scheduling a meeting, providing a recommendation or just having a casual chat), and a low-level process choosing primitive actions to complete the subtask. However, due to the difficulty of natural language understanding (NLU) and natural language generation (NLG), the universal intelligence embodied in these systems still lags behind human beings. 


As we introduced, the modern DS pays more attention to the integration of multiple functions to improve the interactive experience, which makes the function-based classification of the DS insufficient to reflect the current progress. In this paper, we divide the DS by background knowledge which is defined as unstructured text knowledge besides the content of the conversation and can be used during the dialogue. Specifically, we focus on the DS based on the unstructured document(s), namely Document Grounded Dialogue System (DGDS). The DGDS tries to establish a conversation mode in which relevant information can be obtained from the given document(s). Despite the increasing efforts with introducing external structured or unstructured knowledge into the conversation to generate more informative replies in the DS, the DGDS is different from works \cite{DBLP:conf/aaai/GhazvininejadBC18,DBLP:conf/emnlp/WestonDM18,DBLP:conf/cikm/0005HQQGCLSL19} that first retrieve a set of candidate facts or responses and then generate the response by treating the retrieval results as additional knowledge, because the background information the DGDS used is the document(s) and the internal structure of the text need to be considered. We also do not take into account the process of initial selection of candidate documents from a larger knowledge environment (e.g. The Web), because it usually adopts common IR technology (e.g. Keywords, TF-IDF) to find some candidate document(s) \cite{DBLP:conf/acl/ChenFWB17,long2017knowledge,DBLP:conf/starsem/TalmorGB17,DBLP:conf/naacl/TalmorB18,DBLP:journals/corr/abs-1811-01241}, and assumes that ideally, the top-N results cover the most related knowledge. The dialogue history of the DGDS must be based on the same pre-determined document(s), choosing new document(s) as knowledge sources according to the dialogue history should not be counted. 

Besides chit-chat with external knowledge, the DGDS also shares some common features with the MRC. In fact, if a single round QA is counted as a conversation, the DGDS could cover some popular MRC\footnote{in the broad sense, Cloze Tests, Multiple Choice, Span Extraction, Free Answering, KBQA, CQA, etc., are all belong to the MRC task, while in this article we refer to the MRC as the QA tasks based on document(s).} tasks \cite{DBLP:conf/nips/HermannKGEKSB15,DBLP:conf/emnlp/RajpurkarZLL16,DBLP:conf/nips/NguyenRSGTMD16,DBLP:conf/rep4nlp/TrischlerWYHSBS17,DBLP:journals/corr/HillBCW15,DBLP:journals/tacl/KwiatkowskiPRCP19}. Some DGDS models \cite{DBLP:conf/emnlp/MogheABK18,DBLP:conf/emnlp/SaeidiBL0RSB018,DBLP:conf/acl/QinGBLGDCG19} also employ MRC models as baselines \cite{DBLP:journals/corr/SeoKFH16} or component \cite{DBLP:conf/acl/GaoDLS18}. However, single round QA does not need to pay attention to the history of dialogue, resulting in obvious differences with the multi-turn DGDS in the modeling process, memory ability, reasoning ability, evaluation methods, etc. We only include multi-turn Conversational Reading Comprehension (CRC) in the DGDS because it is consistent with our definition. Besides, the single-turn MRC review research has been systematic and comprehensive \cite{gao2019neural,DBLP:journals/corr/abs-1907-01118,DBLP:journals/corr/abs-1906-03824,DBLP:journals/corr/abs-1907-01686}, but the survey on the multi-turn CRC is scarce.

\begin{figure}[h]
\centering
\includegraphics[width=5in]{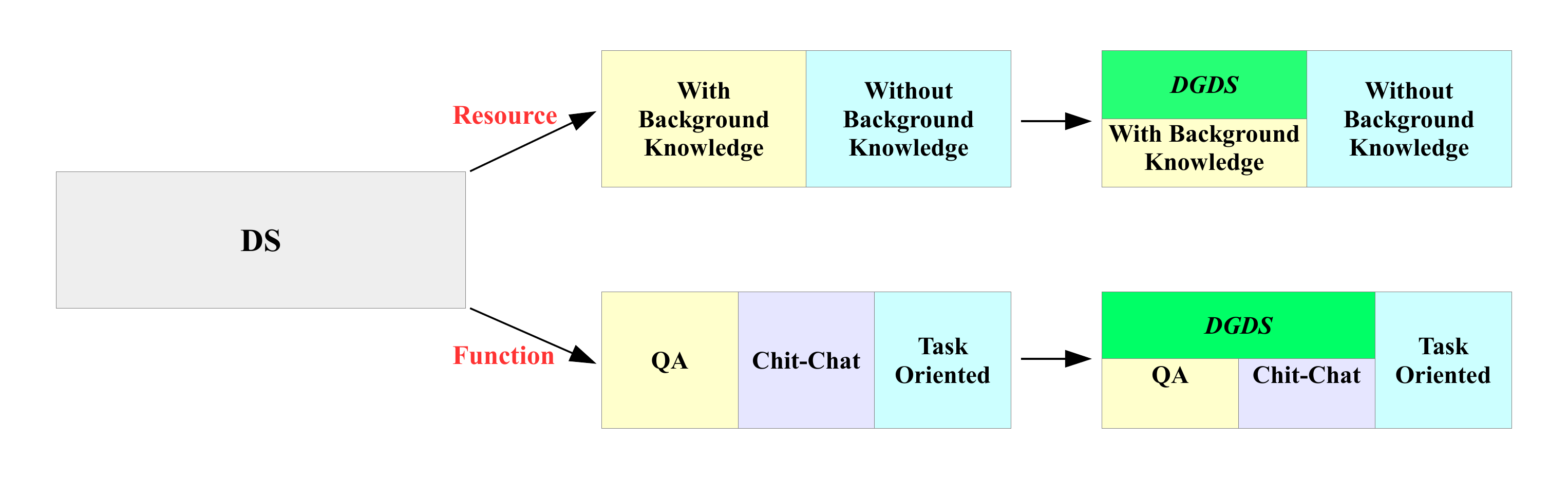}
\caption{The position of the DGDS in Dialogue System.}
\label{DS_3}
\end{figure}

In this paper, we studied the multi-turn DGDS, including Conversational Reading Comprehension (CRC) where the conversations are QA mode and Document-Based Dialogue (DBD) where the dialogues are chit-chat form. The scope of the DGDS in DS is demonstrated in Figure \ref{DS_3}. Since most of the DGDS datasets are released from $2018$ to $2019$, this paper mainly focuses on the relevant deep learning models. As far as we know, we are the first to make a systematic review of this field. We believe that incorporating unstructured document(s) information into response generation is the inevitable trend of the open domain DS because a large number of human knowledge is hidden in these document(s). The research of the DGDS can better assistant AI in using human knowledge and improving AI's understanding of natural language.

The structure of the paper is as follows:
\begin{itemize}
\item In Chapter \textbf{Introduction}, we give a brief history introduction of the DS and the DGDS.
\item In Chapter \textbf{Comparison}, we analyze the difference between the DGDS and three classifications (task-oriented, chit-chat, and QA) of the DS from different perspectives then compare the CRC with the DBD.
\item In Chapter \textbf{Architecture}, the main architecture of the DGDS models are outlined. 
\item In Chapter \textbf{Datasets}, we review the DGDS datasets that have been released.
\item In Chapter \textbf{CRC Models} and \textbf{DBD Models}, we discuss the CRC and the DBD approaches based on the pre-defined architecture respectively. 
\item In Chapter \textbf{Future Work}, we put forward the promising research directions in both fundamental and technical aspects.
\end{itemize}

\section{Comparison}
In this chapter, we demonstrate the distinctions among the DGDS and the traditional DS categories and analyze the differences between the CRC and the DBD in detail. We first define some concepts frequently used in this research field in Table \ref{define_words} to avoid confusion.

\begin{table}[h]
\caption{Related concepts in DGDS.}
\label{define_words}
\begin{tabular}{l|l}
\toprule
Concept & Definition \\
\midrule
Document(s) & The text used for discussion in a conversation, the content of which has an  inseparable coherent\\
     &logical relationship.\\
\midrule
QA & One interlocutor asks questions, the other one provides answers if exist.\\
\midrule
Turn & In QA, one turn is one question or one answer, in chit-chat, one turn is consecutive utterances\\
 &from one speaker. \\
\midrule
Exchange &In QA, each exchange is one QA pair. In chit-chat, any two consecutive turns make one exchange.  \\
\midrule
Dialogue/     &Multiple turns/exchanges constitute a dialogue/conversation.\\
Conversation &  \\
\midrule
Utterance & A sequence of sentences from one speaker in one turn. \\
\midrule
Context & All the utterances but the current one. Context usually means the conversation history. \\
\midrule
Evidence & Text segment(s) containing the information for response generation.\\
\midrule
Agent/Bot & Model trained as dialogue partner.\\
\midrule
User & Real human participating in the dialogue.\\
\bottomrule
\end{tabular}
\end{table}

\subsection{Differences Between the DGDS and the other DS}
\begin{figure}[h]
\centering
\includegraphics[width=5in]{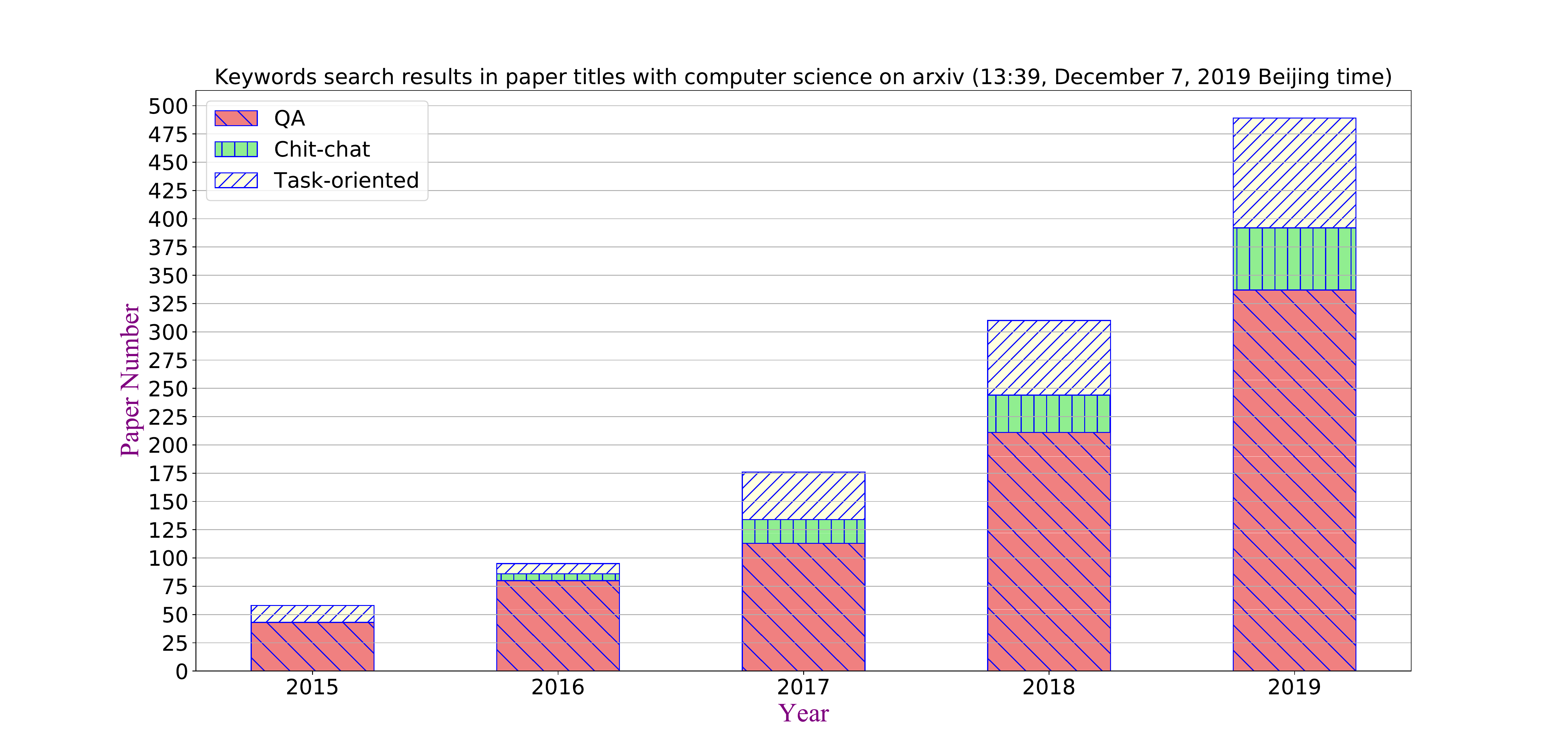}
\caption{The search results using keywords in paper titles on arxiv.org. "Question answering, Machine Reading Comprehension, QA, MRC", "chatbot, chat bot, chit chat, social bot, socialbot" and "task/goal-oriented, task/goal-completion, task/goal-driven" are used respectively.}
\label{DS_bar}
\end{figure}

Accompanied by the rapid development, surveys of the DS nowadays attempt to analyze from different perspectives. In view of task or non-task based DS, \citet{DBLP:journals/sigkdd/ChenLYT17} analyzed the retrieval, generation and hybridization models. \citet{DBLP:journals/spm/Sarikaya17} summarized the technology behind personal digital assistants, such as the system architecture and key components. \citet{DBLP:journals/dad/SerbanLHCP18} classified datasets for building DS from a data-driven perspective. \citet{DBLP:journals/jzusc/ShumHL18} outlined the advantages and disadvantages of the current social chatbots. \citet{DBLP:conf/ijcai/Yan18} summarized the non-task-oriented chit-chat bots. \citet{DBLP:journals/corr/abs-1905-05709} analyzed open-domain DS from three main challenges (semantics, consistency, and interactivity). \citet{DBLP:journals/corr/abs-1909-03409,DBLP:journals/corr/abs-1906-00500} interpreted the DS from the perspective of conditional text generation technology. \citet{DBLP:journals/corr/abs-1905-04071} surveyed the methods and concepts developed for the evaluation of dialogue systems. \citet{DBLP:journals/ftir/GaoGL19} grouped conversational systems into three categories: (1) question answering agents, (2) task-oriented dialogue agents, and (3) social bots. Many later works \cite{DBLP:journals/corr/abs-1811-01241,radlinski2019coached,DBLP:journals/corr/abs-1905-04071} follow this classification. Figure \ref{DS_bar} shows the research trend of these $3$ kinds of DS in the past five years.

There are obvious differences in dialogue patterns among the $3$ types of DS. Dialogue state tracking (DST) and dialogue policy management (DPM) play an important role in the task-oriented DS which helps users accomplish tasks ranging from meeting scheduling to vacation planning. The dialogue focuses on filling predefined semantic slots with respect to special application scenarios. 

The chit-chat bot can converse seamlessly and appropriately with humans about current events and popular topics, and often plays the role of chat companion or recommender. The goal of the system is to act as an emotional partner, and the main concern in the conversation is fluency and coherence. Generally, the more turns a conversation maintains, the more successful the role of an emotional partner becomes.

The QA systems can be categorized into answer selection and answer generation. The former is a retrieving or ranking problem between question and answer candidates, sometimes involving text as background. The later one is a task to select a span or generate an answer based-on the given document(s). The QA-based MRC technology has been greatly developed in the past few years. Equipped with rich knowledge drawn from various data sources including Web document(s) and pre-compiled knowledge graphs (KG’s), the QA agent is able to provide concise direct answers to user's queries. The main concern of the QA systems is the accuracy of the answers.

The DGDS tries to establish a real-world conversation mode as mentioned by former researchers \cite{DBLP:journals/ftir/GaoGL19,DBLP:journals/corr/abs-1811-01241}, where the interlocutors all ask and answer questions and provide each other with a mixture of facts and personal feelings during their general discussion. In other words, the DGDS maintains a dialog pattern in which relevant information can be obtained from the document(s). The similarities between the DGDS and the QA is that they both perform informative replies. The DGDS resembles the Chit-Chat in free form responses.

\subsection{Differences Between the CRC and the DBD}

The DGDS is composed of the CRC and the DBD. In the CRC, users put forward a series of questions for the given document(s), some words of these questions have mutual referential relations. Therefore, the bot needs to consider the history of dialogue to understand the current question. The CRC has gradually become the research hotspot \cite{DBLP:journals/corr/abs-1907-01118} nowadays. The task of the CRC could be formulated as below: Given the document D, the conversation history with previous questions and answers $C = \{q_1, a_1, \dots, q_{n-1}, a_{n-1}\}$ and the current question $q_n$, the goal is to predict the right answer $a_n$ by maximizing the conditional probability $P(a_n|D, C, q_n)$\footnote{Except in the ShARC, the question $q_0$ needed to be answered is asked by User in the beginning of the entire conversation, and the bot needs to ask the follow-up questions based on this $q_0$, then give an answer $a_0$ to close the dialog.}. The main difference between the CRC and the other MRC task is the dialogue history. In order to understand the current question accurately, agents need to solve the problems of coreference and ellipsis in the history of dialogue.

While in DBD, the conversation history is not QA pairs but utterances $U = \{u_1, u_2, \dots, u_n\}$, the target is to predict $u_{n+1}$ with $P(u_{n+1}|D, U)$. Therefore, the common ground of the CRC and the DBD is that they both need to jointly model document(s), historical conversations and last utterance, and they both need to pick out the required evidence from the document(s) and dialog history. For example, the Holl-E \cite{DBLP:conf/emnlp/MogheABK18} dataset constructed by copying or modifying the existing information from a document which is relevant to the conversation, This setup is very similar to CoQA \cite{DBLP:journals/tacl/ReddyCM19} where the answer response is extracted from a document as a span.

The distinctions between the CRC and the DBD lie in dialog patterns, evaluation methods, etc. The DBD takes advantage of document information to generate a reply based on understanding historical dialogue. The CRC tasks usually need to find out the location of the answer based on the understanding of the current question with the help of dialogue history. When encountering unanswerable questions, the CRC (e.g. QuAC) usually gives a "CANNOTANSWER", while DBD can response more freely, the reaction can be a direct answer as "I don't know", a rhetorical question or a changing subject. When evaluating the performance, the CRC tasks care about accuracy more, while the DBD tasks pay more attention to fluency, informative, coherent, etc.

\begin{table}[h]
\caption{Differences among dialog systems. * means QA contains other task besides CRC and MRC, but we only compare these two here. In fact, the MRC includes the CRC, but we use MRC to represent single-turn QA in this table. \# means we only compare with multi-turn chit-chat. }
\label{diff_of_DS}
\begin{tabular}{ll|c|c|c|c|c|c}
\hline
\multicolumn{2}{c|}{\multirow{2}{*}{Characteristic}}& \multicolumn{2}{c|}{\multirow{1}{2cm}{\centering DGDS}} & \multicolumn{2}{c|}{\multirow{1}{2cm}{\centering QA*}}
& \multirow{2}{*}{Task-oriented}
& \multirow{2}{*}{Chit-chat\#}
\\
\cline{3-6}
&& \multicolumn{1}{c|}{\multirow{1}{1.2cm}{\centering DBD}} & \multicolumn{2}{c|}{\multirow{1}{1.2cm}{\centering CRC}} & \multicolumn{1}{c|}{\multirow{1}{1.2cm}{\centering MRC}} & & \\
\hline
\multirow{4}{*}{Function}&\multicolumn{1}{|l|}{Task-completion} & &\multicolumn{2}{c|}{}& & {\color{blue}\checkmark} & \\
&\multicolumn{1}{|l|}{Emotional Partner}& &\multicolumn{2}{c|}{}& & &{\color{blue}\checkmark} \\
&\multicolumn{1}{|l|}{Providing Information}&{\color{blue}\checkmark}& \multicolumn{2}{c|}{\color{blue}\checkmark} &{\color{blue}\checkmark} & &\\
&\multicolumn{1}{|l|}{Exchange Information} &{\color{blue}\checkmark}&\multicolumn{2}{c|}{}&& & \\
\hline
\multirow{4}{*}{Knowledge}&\multicolumn{1}{|l|}{Utterance History} &{\color{blue}\checkmark}&\multicolumn{2}{c|}{\color{blue}\checkmark}&& &{\color{blue}\checkmark}\\
&\multicolumn{1}{|l|}{Background} &{\color{blue}\checkmark}& \multicolumn{2}{c|}{\color{blue}\checkmark} &{\color{blue}\checkmark}& {\color{blue}\checkmark} & \\
&\multicolumn{1}{|l|}{Knowledgebase} & &\multicolumn{2}{c|}{}&& {\color{blue}\checkmark} & \\
&\multicolumn{1}{|l|}{Document(s)-based}&{\color{blue}\checkmark}&\multicolumn{2}{c|}{\color{blue}\checkmark} &{\color{blue}\checkmark}& & \\
\hline
\multirow{6}{*}{Evaluation}&\multicolumn{1}{|l|}{Fluency} &{\color{blue}\checkmark} &\multicolumn{2}{c|}{}&& &{\color{blue}\checkmark} \\
&\multicolumn{1}{|l|}{Coherence} &{\color{blue}\checkmark} &\multicolumn{2}{c|}{}&& &{\color{blue}\checkmark} \\
&\multicolumn{1}{|l|}{Informative} &{\color{blue}\checkmark} &\multicolumn{2}{c|}{}&&{\color{blue}\checkmark}& \\
&\multicolumn{1}{|l|}{Diversity} &{\color{blue}\checkmark}&\multicolumn{2}{c|}{}&& &{\color{blue}\checkmark} \\
&\multicolumn{1}{|l|}{Turns} &{\color{blue}\checkmark} &\multicolumn{2}{c|}{}& & {\color{blue}\checkmark} &{\color{blue}\checkmark} \\
&\multicolumn{1}{|l|}{Accuracy} &&\multicolumn{2}{c|}{\color{blue}\checkmark} &{\color{blue}\checkmark} & {\color{blue}\checkmark} & \\
\hline
\end{tabular}
\end{table}

\subsection{Summarization}
To sum up, the DGDS is distinguished from other conversation scenarios in terms of dialog characteristic, the main concern, knowledge source, etc. In Table \ref{diff_of_DS}, we list the differences not only between the DGDS and the $3$ DS categories, but also between the CRC, the DBD, and the MRC.

\section{Architecture}
\begin{figure}[h]
\centering
\includegraphics[width=4.0in]{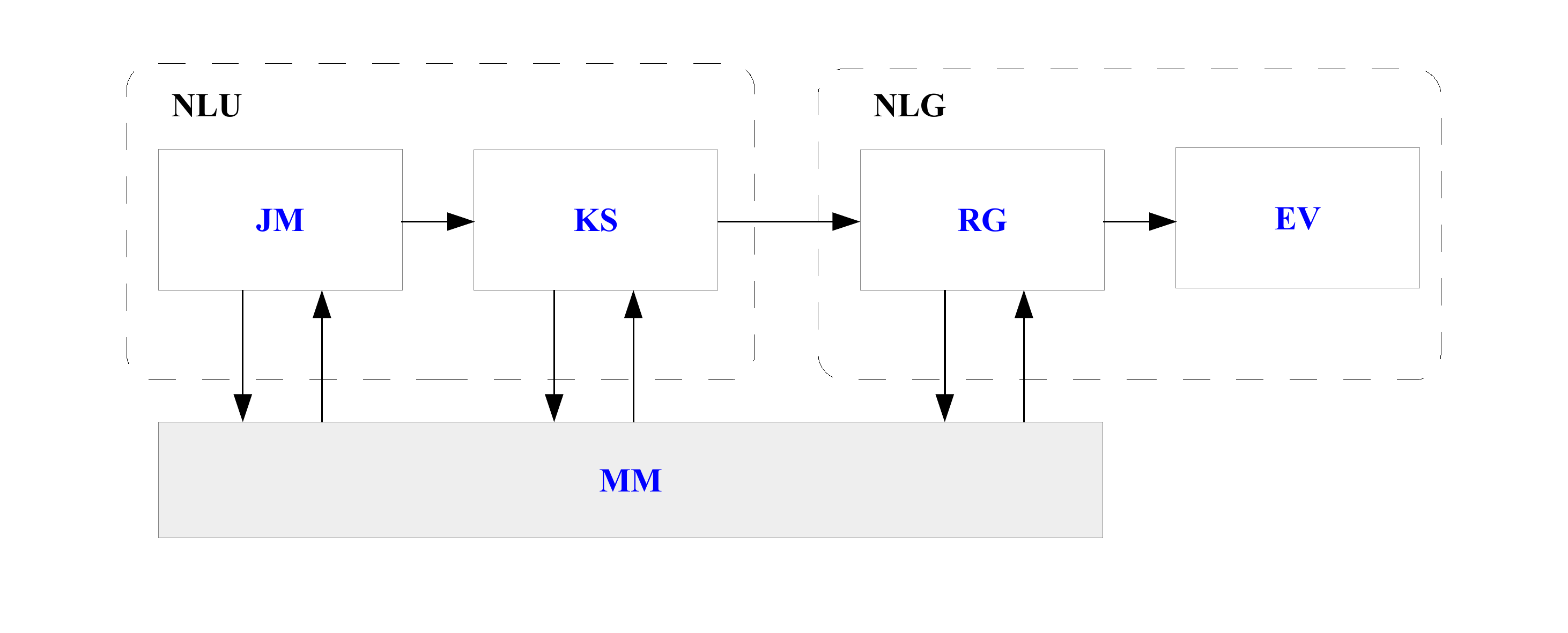}
\caption{The general architecture of the DGDS and the DBD.}
\label{DBD-ALL}
\end{figure}

Most recently, a number of DGDS datasets \citep{DBLP:conf/emnlp/ChoiHIYYCLZ18,DBLP:journals/tacl/ReddyCM19,DBLP:journals/corr/abs-1902-00821,DBLP:conf/emnlp/ZhouPB18,DBLP:conf/emnlp/MogheABK18,DBLP:conf/acl/QinGBLGDCG19,gopalakrishnan2019topical,DBLP:conf/emnlp/SaeidiBL0RSB018} and DGDS models \citep{DBLP:conf/naacl/Yatskar19,DBLP:journals/corr/abs-1909-10772,DBLP:journals/corr/abs-1810-06683,DBLP:journals/corr/abs-1812-03593,DBLP:journals/corr/abs-1905-12848,DBLP:journals/corr/abs-1908-00059,DBLP:conf/cikm/QuYQZCCI19,DBLP:journals/corr/abs-1908-05117,DBLP:conf/sigir/Qu0QCZI19,DBLP:journals/corr/abs-1909-10743,DBLP:conf/aaai/SuGFLZC19,DBLP:conf/acl/ZhongZ19,DBLP:journals/corr/abs-1909-03759,DBLP:journals/access/GuGL19,DBLP:conf/ijcai/ZhaoTWX0Y19,DBLP:conf/acl/LiNMFLZ19,DBLP:conf/ksem/TangH19,DBLP:journals/corr/abs-1906-06685,DBLP:journals/corr/abs-1908-09528,DBLP:journals/corr/abs-1908-06449,DBLP:journals/corr/abs-1903-10245,DBLP:conf/naacl/AroraKR19} have been proposed to mine unstructured document information in conversation. According to the characteristics of the task, current approaches normally consist of $5$ parts: \textbf{joint modeling (JM)}, \textbf{knowledge selection (KS)}, \textbf{response generation (RG)}, \textbf{evaluation (EV)} and \textbf{memory (MM)}. We define the JM and the KS as the NLU problem and reckon the RG and the EV as the NLG problem. The general architecture of the DGDS is in Figure \ref{DBD-ALL}, we will introduce JM, KS, RG, and EV in this chapter. The memory (MM) module that the researchers haven't studied in depth in the DGDS untill now will be discussed in the Future Work chapter.

\subsection{Joint Modeling} Joint modeling refers to the integration of all input information. Joint modeling in the DGDS integrates the background document(s), current utterance and the dialogue history. Factors such as temporal relationship, connections among context, current utterance, and document(s), etc. need to be considered. When it comes to the differences in modeling, compared with the MRC model, the DGDS needs to consider conversation history, and compared with chit-chat, the DGDS needs to take background knowledge into account. 

In the neural DS, \citet{lowe2015incorporating} first presented a method for incorporating unstructured external textual information for predicting the next utterance. Their model is an extension of the dual-encoder model \cite{DBLP:conf/sigdial/LowePSP15}. There is a gap between the model they proposed and the actual application scenario. \citet{DBLP:conf/sigir/YangQQGZCHC18} incorporated external knowledge with pseudo-relevance feedback and QA correspondence knowledge distillation for a response ranking task. They built matrices on the word-level and semantic vector-level similarity. These methods treated the utterance history as a parallel relationship. Many DGDS models \cite{DBLP:journals/tacl/ReddyCM19,DBLP:journals/corr/abs-1812-03593,DBLP:conf/sigir/Qu0QCZI19,DBLP:conf/cikm/QuYQZCCI19,DBLP:journals/corr/abs-1909-10772,DBLP:conf/ijcai/ZhaoTWX0Y19,DBLP:conf/naacl/AroraKR19,DBLP:conf/ksem/TangH19,DBLP:journals/corr/abs-1906-06685,DBLP:journals/corr/abs-1908-06449,DBLP:journals/corr/abs-1908-09528} followed this parallel setting.

\begin{figure}[h]
\centering
\includegraphics[width=4.5in]{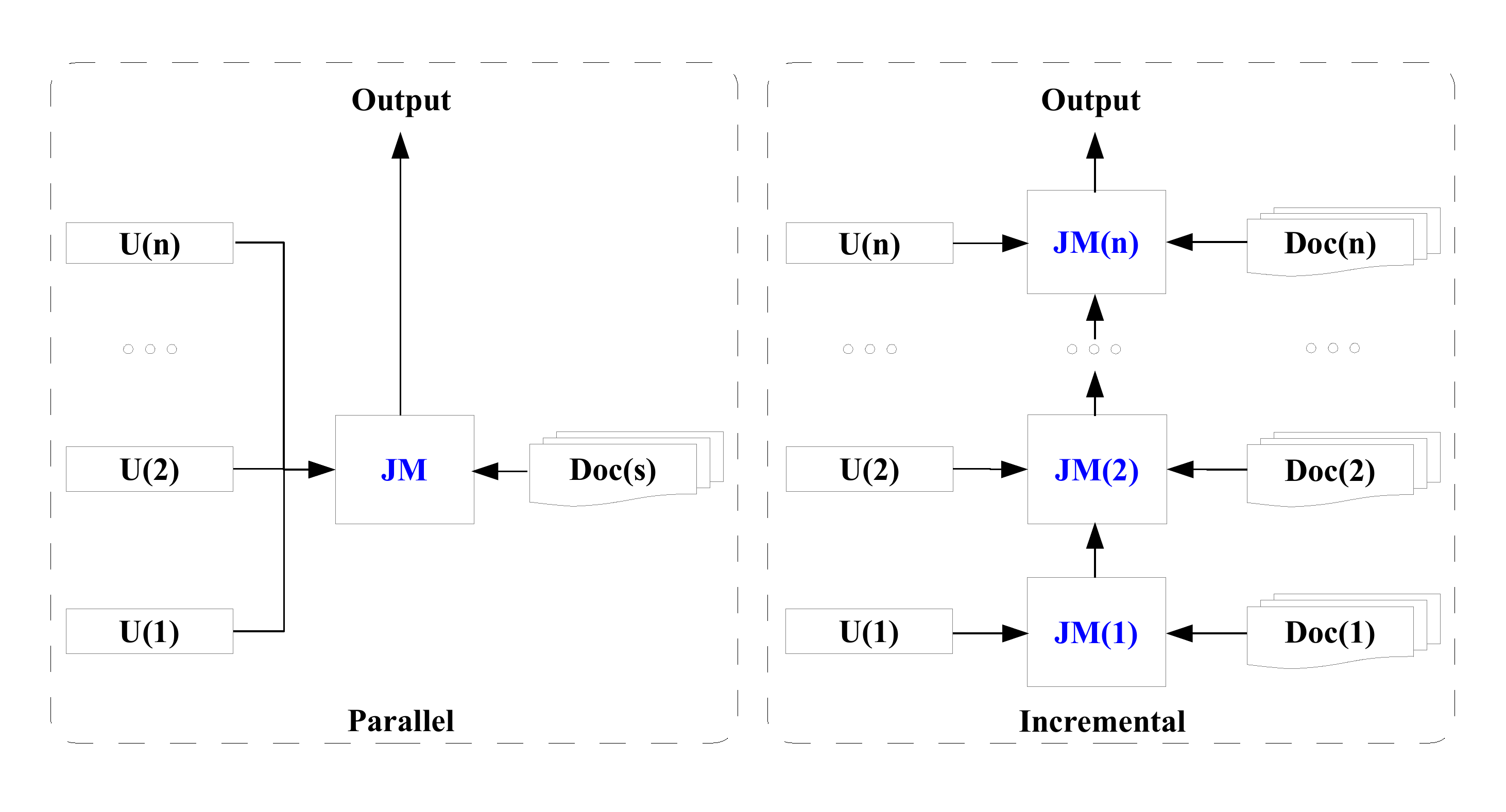}
\caption{The comparison of Parallel Modeling and Incremental Modeling. U is short for utterance, Doc is short for document.}
\label{Incremental}
\end{figure}

\citet{DBLP:conf/coling/ZhangLZZL18} weighted previous conversations with a turns-aware aggregation design. \citet{DBLP:conf/acl/WuLCZDYZL18} constructed the historical dialogue into 3-D tensor to extract the features between different rounds of dialogue and response. These works treated the dialogue history differently according to temporal relationships, which we reckoned an incremental setting. Some DGDS \cite{DBLP:journals/corr/abs-1810-06683,DBLP:journals/corr/abs-1908-05117,DBLP:journals/corr/abs-1908-00059,DBLP:journals/access/GuGL19,DBLP:conf/acl/LiNMFLZ19} models employed the similar idea.

According to the way the previous DS model and the DGDS model handled historical conversations and text, we classify the JM into \textbf{Parallel Modeling} and \textbf{Incremental Modeling} as Figure \ref{Incremental}. The parallel modeling means that the document(s), the historical dialogue and the last utterance are processed as a parallel relationship. The incremental modeling preserves the temporal relationship between the historical conversation and the current conversation and models them successively with the document(s).


\subsection{Knowledge Selection}
The knowledge selection (KS) in the DGDS is defined as the information seeking process in the document(s) and the utterance history that can be used to construct the response. According to whether there is an interpretable reasoning path when selecting, we divide the KS into \textbf{selection} and \textbf{reasoning}. In terms of the processed objects, we could divide the KS into \textbf{context-based} and \textbf{document(s)-based}.

The selection method in the DGDS normally extracts text segments \cite{DBLP:journals/corr/abs-1812-03593,DBLP:conf/emnlp/MogheABK18,DBLP:conf/acl/QinGBLGDCG19} or selects some history utterances \cite{DBLP:conf/sigir/Qu0QCZI19,DBLP:conf/cikm/QuYQZCCI19}. While the reasoning method tries to build an interpretable reasoning path to the evidence in document(s) \cite{DBLP:journals/corr/abs-1810-06683,DBLP:journals/corr/abs-1908-00059}. 

The DGDS task parallels a growing interest in developing datasets that test specific reasoning abilities: algebraic reasoning \cite{DBLP:conf/aaai/Clark15}, logical reasoning \cite{DBLP:journals/corr/WestonBCM15}, commonsense reasoning \cite{DBLP:conf/semeval/OstermannRMTP18} and multi-fact reasoning \cite{DBLP:conf/aclnut/WelblLG17,DBLP:conf/naacl/KhashabiCRUR18,DBLP:conf/naacl/TalmorB18}. In MRC, \citet{DBLP:conf/naacl/DuaWDSS019} propose a task for discrete reasoning over the content of paragraphs. They define reasoning as subtraction, comparison, selection, addition, count, coreference resolution, etc. In the multi-turn DS, \citet{DBLP:conf/naacl/WuMK18} employ a multi-hop attention with termination mechanism \cite{DBLP:conf/kdd/ShenHGC17} for reasoning. 

In the CRC, \citet{DBLP:conf/emnlp/SaeidiBL0RSB018} reckoned reasoning ability as a fundamental challenge. An example is showed in table \ref{ShARC}. According to the input content, the system needs to infer the missing conditions and put forward the corresponding questions to answer the initial question. Another example is given from CoQA in Table \ref{coQA_Rs}. The understanding of Question $4$ and the reasoning of Answer $4$ are based on the document(s) and the former QA pairs. In DBD, \citet{DBLP:journals/corr/abs-1903-10245} performed reasoning on a knowledge augmented graph. 

\begin{table}[h]
\caption{One example in the ShARC dataset.}
\label{ShARC}
\begin{tabular}{l|l|l}
\toprule
Speaker & Text Type & Text content \\
\midrule
&Rule Text&You’ll carry on paying National Insurance for the first 52 weeks you’re \\
 &&abroad if you’re working for an employer outside the EEA.\\
\midrule
User&Scenario&I am working for an employer in Canada.\\
\midrule
User&\textcolor{red}{Target Question}&\textcolor{red}{Do I need to carry on paying UK National Insurance?}\\
\midrule
Bot&Follow-up Question & Have you been working abroad 52 weeks or less?\\
\midrule
User&Follow-up Answer &Yes\\
\midrule
Bot&\textcolor{red}{Target Answer} & \textcolor{red}{Yes}\\
\bottomrule
\end{tabular}
\end{table}

\subsection{Response Generation} 
In the DGDS, there are three ways to generate replies: extraction \cite{DBLP:conf/naacl/AroraKR19}, retrieval \cite{DBLP:conf/ijcai/ZhaoTWX0Y19}, and generation \cite{DBLP:conf/acl/LiNMFLZ19}. Extraction needs to determine the beginning and end of the text, retrieval needs to select one from the candidates the highest score among the given candidates, and generation needs to generate words in turn to form a complete reply. Based on whether generating all words, we consider the extraction and retrieval to be \textbf{indirect}, while generation to be \textbf{direct}. Some works try to combine the indirect and direct are defined as \textbf{hybrid} \cite{DBLP:conf/acl/ZhongZ19,DBLP:journals/corr/abs-1908-06449}. 

Language models \cite{DBLP:conf/nips/BengioDV00,DBLP:conf/interspeech/MikolovKBCK10} can predict the next word given the sequence of the previous word, so they are widely used in the task of text generation. For example, encoder-decoder model \cite{DBLP:conf/emnlp/ChoMGBBSB14}, sequence-to-sequence model \cite{DBLP:conf/nips/SutskeverVL14} and attention based models \cite{DBLP:conf/naacl/PetersNIGCLZ18,DBLP:conf/nips/VaswaniSPUJGKP17,radford2018improving,DBLP:conf/naacl/DevlinCLT19,DBLP:journals/corr/abs-1906-08237} greatly improve the text generation tasks. We observed that the attention-based generation are often adopted in the DGDS.

\subsection{Evaluation} EV is used to judge whether the generated text meets the DGDS requirements. In the CRC, we normally adopt to some mature metrics of information retrieval (IR) such as accuracy. The DBD models are usually measured with the word overlap metrics (F1, BLEU, etc), which are insufficient for dialog scenario \cite{DBLP:conf/acl/LoweNSABP17,DBLP:conf/aaai/TaoMZY18}, hence human evaluations are usually employed paralleled with auto evaluations. We still need to establish an auto evaluation metric highly correlated with human evaluations for dialog quality measuring.

Generally speaking, DS with good performance needs to have the characteristics of high semantic relevance, rich information, and diverse expressions. There are many works addressed the evaluation methods of the DS. \citet{DBLP:conf/sigdial/Paek01} studied what purpose dialogue measurement serves, and then propose an empirical method to evaluate the system that meets that purpose. \citet{DBLP:conf/emnlp/LiuLSNCP16} introduced embedding-based relevance evaluation metrics (the Greedy Matching, the Embedding Average, The Vector Extreme). \citet{DBLP:conf/acl/LoweNSABP17} introduced ADEM for mimic human evaluation. \citet{DBLP:journals/corr/KannanV17} discussed the adversarial evaluation in the DS. \citet{DBLP:conf/naacl/XuJLRWWWW18} proposed mean diversity score (MDS) and probabilistic diversity score (PDS) to evaluate the diversity of responses generated when multiple reference responses are given. \citet{DBLP:conf/aaai/TaoMZY18} evaluated a reply by taking into consideration both a ground-truth reply and a query. \citet{DBLP:conf/aaai/SaiGKS19} pointed out 
the drawback of the ADEM and outlined we still have a long way to go in the automatic evaluation of DS. Similar to other DS, the DGDS is also in an era of lack of automatic evaluation indicators.

\subsection{Summarization}
At present, the models used to solve the CRC problems are mainly extractive and generative, while retrieval and generative models are usually adopted to address the DBD. We summarize the differences between the CRC and the DBD in dealing with the $4$ model components in Table \ref{architecture}. 

\begin{table*}[h]
\caption{Main architecture in the DGDS. "Acc." means accuracy.}
\label{architecture}
\begin{tabular}{|c|c|c|c|c|}
\hline
\multirow{2}{*}{architecture} &
\multicolumn{2}{c|}{CRC} &
\multicolumn{2}{c|}{DBD} \\
\cline{2-5}
& \multirow{1}{1.5cm}{\centering Extractive} & \multirow{1}{1.8cm}{\centering Generative} & \multirow{1}{1.5cm}{\centering Retrieval } & \multirow{1}{1.8cm}{\centering Generative} \\
\hline
JM & Parallel / Incremental & Parallel & Parallel & Parallel / Incremental \\
\hline
KS & Selection & Reasoning & Selection & Selection / Reasoning \\
\hline
RG & Indirect & Direct / Hybrid & Indirect & Direct / Hybrid \\
\hline
EV & Acc./ Word overlap & Acc./ Word overlap & Acc. & Lack of standards* \\
\hline
\end{tabular}
\end{table*}

\section{Datasets}
We introduce the DGDS related datasets of the CRC and the DBD respectively.

\subsection{CRC Datasets}
Recently, a series of CRC datasets are proposed by researchers. For instance, \citet{DBLP:journals/tacl/ReddyCM19} released CoQA, a dataset with 8,000 conversations about given passages from seven different domains. We present a dialogue example in the CoQA in Table \ref{coQA_Rs}. The two interlocutors conduct a question and answer dialogue according to the given text.

\begin{table}[h]
\caption{The CoQA dialogue example. The same text color corresponds to the QA pair and the evidence.}
\label{coQA_Rs}
\begin{tabular}{l|l|l|l}
\toprule
Passage: & \multicolumn{3}{l}{Jessica went to sit in her rocking chair. \textcolor{red}{Today was her birthday} and \textcolor{blue}{she was turning 80}.} \\
& \multicolumn{3}{l}{\textcolor{magenta}{Her granddaughter Annie} was coming over in the afternoon and Jessica was very excited}\\
&\multicolumn{3}{l}{to see her. \textcolor{magenta}{Her daughter Melanie and Melanie’s} \textcolor{magenta}{husband Josh} were coming as well.}\\
\midrule
Question 1: & \textcolor{red}{Who had a birthday?} & Answer 1 : & \textcolor{red}{Jessica} \\
\midrule
Question 2: & \textcolor{blue}{How old would she be?} & Answer 2 : & \textcolor{blue} {80} \\
\midrule
Question 3: & \textcolor{green}{Did she plan to have any visitors?} & Answer 3 : & \textcolor{green}{Yes} \\
\midrule
Question 4: & \textcolor{cyan}{How many?} & Answer 4 : & \textcolor{cyan}{Three} \\
\midrule
Question 5: & \textcolor{magenta}{Who?} & Answer 5 : & \textcolor{magenta}{Annie, Melanie and Josh} \\
\bottomrule
\end{tabular}
\end{table}

At the same time, \citet{DBLP:conf/emnlp/ChoiHIYYCLZ18} introduced QuAC, Compared with the CoQA, document(s) are only given to the answerer, whereas the questioner asks the questions based on the title of passages. The answerer replies to the question with a subsequence of the original passage and determines whether the questioner can ask a follow-up question. Hence the dialogs often switch topics compared with CoQA's dialogs including more queries for details. The CoQA answers are less than $3$ tokens long on average, while QuAC’s are over $14$. Both CoQA and QuAC have a question type prediction subtask. The training set of the QuAC has one reference answer, while dev and test set questions have multiple references each. We present an instance of the dev set of QuAC in Table \ref{QuAC_dev}. We only keep the reference answers with obvious differences, Compared with the DBD's multiple references example in Table \ref{one2many}, the diversity of the CRC's multiple references is insufficient. This is suffered from extracting fragments from the document(s) as reference.

\begin{table}[h]
\caption{The multiple reference example of the QuAC dev set.}
\label{QuAC_dev}
\begin{tabular}{l|l}
\toprule
Category & Text \\
\midrule
Background & "Anna Vissi ... also known as Anna Vishy, is a Greek Cypriot singer, ..." \\
\midrule
Document & "In May 1983, she married Nikos Karvelas, a composer, with whom she ... " \\
\midrule
question 1/2/3 & "what happened in 1983?" / "did they have any children?" / "did she have any other children?" \\
\midrule
question 4 & "what collaborations did she do with Nikos?" \\
answers & \textcolor{magenta}{"After their marriage, she started a close collaboration with Karvelas ..."} \\
& \textcolor{magenta}{"a composer, with whom she collaborated in 1975"} \\
& \textcolor{magenta}{"Thelo Na Gino Star"} \\
& \textcolor{magenta}{"Since 1975, all her releases have become gold or platinum and ..."} \\
\midrule
question 5/6/7 & "what influences does he have ..." / "what were some of the songs?" / "how famous was it?" \\
\midrule
question 8 & "did she have any other famous songs?" \\
answers & \textcolor{magenta}{"In 1986 I Epomeni Kinisi (The Next Move) was released."} \\
& \textcolor{magenta}{"Epomeni Kinisi (The Next Move) was released."} \\
& \textcolor{magenta}{"Pragmata (Things)"} \\
& \textcolor{magenta}{"The album included the hit Pragmata (Things) and went platinum,"} \\

\bottomrule
\end{tabular}
\end{table}

Sequence QA attracted attention in recent years. The RC2 dataset \cite{DBLP:journals/corr/abs-1902-00821} leverages knowledge from reviews to answer multi-turn questions from customers, which is an open-domain CRC task. The QBLink \cite{DBLP:conf/emnlp/ElgoharyZB18} is a sequential QA dataset asking multiple related questions about a Wikipedia page. The questions are designed primarily to challenge human players in Quiz Bowl tournaments. However, due to the continuous change of the document(s) for the question, they cannot be classified into the DGDS. \citet{DBLP:conf/naacl/MaJC18} also introduced a CRC dataset. We did not include it here for that it both treats dialogues as documents \cite{DBLP:journals/tacl/SunYCYCC19} and limits task to entity completion.
Unlike other CRC datasets where conversations are led by the users, \citet{DBLP:conf/emnlp/SaeidiBL0RSB018} introduced SHARC, a sequential QA task based on regulation texts given by government websites and the bot lead the dialogue interaction. When users ask a question about laws and regulations based on a certain scenario, the bot will give accurate answers through a series of queries to users. The bot needs to understand complex decision rules based on the conversation history and certain scenarios to answer the initial question raised by users, while other CRC tasks usually extract answers directly from texts. Besides, the SHARC task needs to develop free-form follow-up questions to determine whether the user meets the decision rules, while other CRC tasks do not.

\citet{DBLP:conf/acl/IyyerYC17} initiated sequential QA on a structured knowledge table. \citet{DBLP:conf/aaai/SahaPKSC18} and \citet{DBLP:conf/nips/GuoTDZY18} both introduced the task of Complex Sequential QA based on KG. Most recently, \citet{DBLP:conf/cikm/ChristmannRASW19} proposed ConvQuestions datasets which are also based on a KG. These sequential QA tasks based on structured data are not included in the DGDS. There are other interesting tasks such as constructing QA pairs from a document \cite{DBLP:conf/acl/KrishnaI19}, asking right sequential questions about a document \cite{DBLP:conf/naacl/FuCD19}, decomposing complex questions into a sequence of simple questions \cite{DBLP:conf/naacl/TalmorB18}, which have some potential connections with the CRC.

\subsection{DBD Datasets}
Recently, a number of DBD datasets based on movie domain have been released \cite{DBLP:conf/emnlp/ZhouPB18,DBLP:conf/emnlp/MogheABK18}. The document(s) they discussed come from multi-sources (e.g. Wikipedia), the conversations they used are often collected from crowdsourcing platform Amazon Mechanical Turk (AMT) or Reddit.

The CMUDoG \cite{DBLP:conf/emnlp/ZhouPB18} is a dataset built with AMT where dialog is generated based on the given background text. The Holl-E dataset \cite{DBLP:conf/emnlp/MogheABK18} also addresses the lack of background knowledge in the DS. The difference between CMUDoG and Holl-E are list below:

\begin{itemize}
\item The CMUDoG only uses Wikipedia article, mainly the plot of the movie, which is divided into four sections, one section is the basic information of a movie such as reviewers' comments, introductions, ratings, and the other three sections are the plots of the movie). The Holl-E uses Reddit Comments, IMDB and Wikipedia introductions as the background.
\item The CMUDoG is discussed according to the sections. While the Holl-E can discuss any part of the given document(s) as long as it is relevant to the current conversations, and the author gives the source evidence of each utterance in the dataset.
\item The CMUDoG has two scenarios (whether the interlocutor knows the document(s) or not), while both Holl-E's interlocutors are aware of the content of the document(s).
\item There are two versions of the test set in the Holl-E: one with a single golden reference and the other with multiple golden references, while the CMUDoG only has one test set with one single golden reference for each dialogue. We present a dialog example of Holl-E in Figure \ref{one2many}.
\item The Holl-E collects self-conversations \cite{DBLP:journals/corr/abs-1709-09816} which means the same worker plays the role of both parties in the conversation.
\item The Holl-E has five manual ratings for conversational fluency similar to the CoQA, while CMUDoG uses BLEU and dialog turns to classify the dialogues. The BLEU here measures the overlap of the turns of the conversation with the sections of the document. 
\item The dialogues in the Holl-E allow one speaker to freely organize the language, the second speaker needs to copy or modify the existing information, also similar to the CoQA. But unlike the CoQA, the second speaker can properly add words before or after span to ensure smooth dialogue. To comparison, the CMUDoG placed no limits on dialogues except the two interlocutors respectively play an implicit recommender and recommended role.
\end{itemize}

\begin{table*}
  \caption{The one-to-many examples in Holl-E test set. The movie name is "The Secret Life of Pets", we do not present the documents (comments, review, plot) for saving space.}
   \label{one2many}
   \begin{tabular}{c|c|l}
    \hline
    Speaker & Reference & Utterance  \\
\hline
User & & What do you think about the movie? \\
\hline
Bot && I think it was comical and entertaining. \\
\hline
User && It delivered what was promised. \\
\hline
Bot &A& \textcolor{blue}{I agree! I’m surprised this film got such a low overall score by users.} \\
 &B& \textcolor{blue}{My favorite character was Gidget! She was so much fun and so loyal to her friends!} \\
&C& \textcolor{blue}{Yes! As a Great Dane owner, I often wonder what my dogs are thinking. It was fun } \\
&& \textcolor{blue}{to see this take on it.} \\
&D& \textcolor{red}{It was full of cliches with a predictable story, but with some really funny moments.}\\
    \hline
  \end{tabular}
\end{table*}

To provide the conversation model with relevant long-form text on the fly as a source of external knowledge, \citet{DBLP:conf/acl/QinGBLGDCG19} presented a dataset where the next utterance is generated with web-text. Given the dialogue history obtained from Reddit, the web text is treated as an external source to generate the next dialogue related to the dialogue history and containing external knowledge. Some symbols on the original web page file are retained as annotations, such as "\#" annotating the dialogue, the corresponding keyword or paragraph in the document, "<title>" annotating the position of the title, "<p>" annotating the text structure, etc. 

Topical-Chat \cite{gopalakrishnan2019topical} relies on multiple data sources, including Washington Post articles, Reddit fun facts and Wikipedia articles about pre-selected entities, to enable interactions where the interlocutors have no explicit roles. The external knowledge provided to interlocutors could be the same or not, leading to more diverse conversations.

Reddit conversational dataset is released by Dialog System Technology Challenges 7 (DSTC-7)\footnote{https://github.com/mgalley/DSTC7-End-to-End-Conversation-Modeling/tree/master/data\_extraction} and is extracted from the Reddit website. For each web page of the Reddit website, there is a link below the title, which might provide background knowledge for the current topic. The data can be obtained from Reddit dump\footnote{http://files.pushshift.io/reddit/comments/} and Common Crawl\footnote{http://commoncrawl.org/}. After further filtering, it may be used in the DGDS, but it does not meet our definition at present because the background knowledge is either redundancy or has low relevance with the conversation.

There are some other works trying to incorporate unstructured documents knowledge into DS. \citet{DBLP:conf/coling/VougiouklisHS16} proposed a dataset aligning knowledge from Wikipedia in the form of sentences with sequences of Reddit utterances. \citet{DBLP:journals/corr/abs-1811-01241} released a dataset which constructed with ParlAI \cite{DBLP:conf/emnlp/MillerFBBFLPW17} to provide a supervised learning benchmark task which exhibits knowledgable open dialogue with clear grounding. However, these datasets remove the organization of the document(s) by flattening all the paragraphs into separate sentences. \citet{DBLP:conf/naacl/AkasakiK19} proposed a task to initiate a conversation based on the given document, which has some potential connection with the DBD.

\subsection{Summarization}
Table \ref{diff_of_datasets} summarize the characteristics of above mentioned DGDS datasets in 6 aspects: chat style, domain of dialogue, whether the speaker can see the document(s), whether the dataset is labeled for training, which side leads the conversation.

\begin{table}[h]
\caption{Comparison of Document Grounded Conversation Datasets. "Flu." is short for Fluency.}
\label{diff_of_datasets}
\begin{tabular}{l|c|c|c|c|c|c}
\toprule
Dataset & QA & See Doc & Dialog.Source & Domain & Labeled & Lead by \\
\midrule
CoQA\cite{DBLP:journals/tacl/ReddyCM19} & {\color{blue}\checkmark}& Both & AMT & Open & Span & User\\
QuAC\cite{DBLP:conf/emnlp/ChoiHIYYCLZ18} & {\color{blue}\checkmark}&Bot & AMT & Open & Span & User \\
ShARC\cite{DBLP:conf/emnlp/SaeidiBL0RSB018}&{\color{blue}\checkmark}& Both & AMT & Regulatory & No & Bot \\
RC2\cite{DBLP:journals/corr/abs-1902-00821} & {\color{blue}\checkmark}& Both & Manually & Review & Span & User \\
CMUDoG\cite{DBLP:conf/emnlp/ZhouPB18} & &User/both & AMT & Movie & No & Both \\
Holl-E\cite{DBLP:conf/emnlp/MogheABK18} & &Both & AMT & Movie & Flu./Span & Both \\
CbR\cite{DBLP:conf/acl/QinGBLGDCG19} & &All & Reddit & Open & No & Multi\\
T-Chat\cite{gopalakrishnan2019topical} & &Both & ParlAI/AMT & Open & No & Both\\
\bottomrule
\end{tabular}
\end{table}

Table \ref{statistics_of_datasets} illustrates the statistics of the DGDS datasets with total dialog numbers, turns per dialogue, total number of document(s), average words of document, average word of each utterance. It is worth mentioning that the definition of "Turns" represents a QA pair in the CRC and one utterance in the DBD. We can observe that the DBD datasets have more dialog turns, more words in both document(s) and utterances.

\begin{table}[h]
\caption{Statistics of Document Grounded Conversation Datasets. The statistics with the asterisk are from us.}
\label{statistics_of_datasets}
\begin{tabular}{l|c|c|c|c|c}
\toprule
dataset & Dialogs & Turns/Dialog &Num.of.Doc. & Words/Doc. & Words/Utter. \\
\midrule
coQA\cite{DBLP:journals/tacl/ReddyCM19} & 8,399 & 15.2 & 8,399 & 271 & 4.1 \\
QuAC\cite{DBLP:conf/emnlp/ChoiHIYYCLZ18} & 13,594 & 7.2 & 8,854 & 401 & 10.6 \\
ShARC\cite{DBLP:conf/emnlp/SaeidiBL0RSB018} &32,436&2.7*& 948 &60.6*&5.2*\\
RC2\cite{DBLP:journals/corr/abs-1902-00821} & 1,218* & 3.9* & 1,218* & 108.1* & 4.3* \\
CMUDoG\cite{DBLP:conf/emnlp/ZhouPB18} & 4,112 & 21.4 & 120 & 229 & 18.6 \\
Holl-E\cite{DBLP:conf/emnlp/MogheABK18} & 9,071 & 10.0 & 921 & 727.8 & 15.3 \\
CbR\cite{DBLP:conf/acl/QinGBLGDCG19} & 2.82M & 86.2 & 32.7k & 7,347.4 & 18.7 \\
T-Chat\cite{gopalakrishnan2019topical} & 11,319 & 21.9 & 3064* & 830* & 19.7 \\
\bottomrule
\end{tabular}
\end{table}

\section{CRC models}

In this chapter, we summarize and analyze the current CRC models according to the architecture defined previously. The existence of the historical answers of the CRC entails differences in JM and KS with the DBD. For example, in the KS processing of the CoQA, the corresponding answer position can be predicted for each historical question, which means that in the JM part of the CRC, the historical answers are not necessarily added to the model, but only used as a midterm target for the RS ability training.

\subsection{Joint Modeling in the CRC}
In CRC, joint modeling (JM) aims to integrate the background document(s) and dialogue history (QA pairs) in Table \ref{architecture}.

\subsubsection{Parallel Modeling}
We further classify the parallel modeling into complete parallel and partial parallel depends on whether concatenating all utterances together.

\textbf{Complete Parallel.} \citet{DBLP:journals/tacl/ReddyCM19} proposed a hybrid model, DrQA \cite{DBLP:conf/acl/ChenFWB17} + PGNet \cite{DBLP:conf/acl/GuLLL16,DBLP:conf/acl/SeeLM17}, which combines the sequence-to-sequence model and MRC model together to extract and generate answers. To integrate information of conversational history, they treated previous question-answer pairs as sequence and append them to the document. Similar to \citet{DBLP:journals/tacl/ReddyCM19}, \citet{DBLP:journals/corr/abs-1812-03593} proposed SDNet which appends previous question-answer pairs to the current question, while in order to find out the related conversational history, they employed additional self-attention on previous questions. 

\citet{DBLP:conf/aaai/SuGFLZC19} defined two types of questions, verification ones and knowledge-seeking ones in the CRC and proposed an adaptive framework for the CoQA. They first extracted the rationale for the question from the document(s), then used different components for the corresponding question type. \citet{DBLP:journals/corr/abs-1909-10772} used RoBERTa\cite{DBLP:journals/corr/abs-1907-11692} combining with adversarial training (AT) \cite{DBLP:journals/corr/GoodfellowSS14} and knowledge distillation (KD) \cite{DBLP:conf/icml/FurlanelloLTIA18} to leverage additional training signals from well-trained models. They used parallel encoding, where $Q_k^* = \{Q_1, A_1, \dots ,Q_{k-1}, A_{k-1}, Q_k\}$, and concatenated $Q_k^*$ with document(s) as the input of RoBERTa. Since the answers of CoQA dataset can be free-form text, Yes, No or Unknown, and besides the Unknown answers, each answer has its rationale, an extractive span in the passage, they also add a new task named Rationale Tagging in which the model predicts whether each token of the paragraph should be included in the rationale. In other words, tokens in the rationale will be labeled $1$ and others will be labeled $0$. For unknown questions, they should all be $0$. 

On the ShARC task, \citet{DBLP:conf/acl/ZhongZ19} proposed an Entailment-driven Extracting and Editing (E3) model, \citet{DBLP:journals/corr/abs-1909-03759} presented an UrcaNet model to learn the deep level clues, \citet{DBLP:conf/emnlp/LawrenceKN19} made the sequence generation process bidirectional by employing special placeholder tokens. They all adopted the parallel modeling which treated the rule text, scenario and context equally.

\textbf{Partial Parallel.} \citet{DBLP:conf/naacl/Yatskar19} used BiDAF++ \cite{DBLP:conf/acl/GardnerC18} with ELMo \cite{DBLP:conf/naacl/PetersNIGCLZ18} to answer the question based on the given document and conversational history. Besides encoding previous dialog information to the context representation, they labeled answers to previous questions in the context. They proposed to first make a Yes/No decision, then output an answer span only if Yes/No was not selected.

\citet{DBLP:journals/corr/abs-1905-12848} proposed a fine-tuning BERT (w/k-ctx)\footnote{(k-ctx) means employing previous k QA pairs.} model, which treated questions and answers as independent input, concatenated them to a passage and encode with Bert. \citet{DBLP:conf/cikm/QuYQZCCI19} proposed a general framework for the QuAC in an information-seeking point of view. Some other works only verified on the CoQA development set. 

\subsubsection{Incremental Modeling}
\citet{DBLP:journals/corr/abs-1810-06683} adopted a flow mechanism to better understand conversational history. The FlowQA model employed an alternating parallel processing structure and incorporated intermediate representations generated during the process of answering previous questions. We classify the FlowQA into the incremental modeling as the encoding information of historical dialogue is accumulated from far to near turn. \citet{DBLP:journals/corr/abs-1908-05117} further considered the long-distance historical dialogue information in the process of reasoning. They also proposed a Bert-FlowDelta to investigate the extension of Bert into the multi-turn dialog. These models carried out experiments on both the CoQA and the QuAC.

\citet{DBLP:journals/corr/abs-1908-00059} proposed a GraphFLOW structure, which built a sparse graph from the document and utterances history dynamically then reasoned with the Graph-Flow mechanism which sequentially processed the graphs they constructed. Following \citet{DBLP:conf/emnlp/ChoiHIYYCLZ18}, they leveraged conversation history by concatenating a feature vector encoding previous $N$ answer locations to the document(s) word embeddings. They also prepended the previous $N$ QA pairs to the current question and concatenated a $3$ dimension relative turn marker embedding to each word vector in the augmented question to indicate which turn it belongs to. 

Experimenting on the CoQA development set, \citet{DBLP:journals/access/GuGL19} proposed a TT-Net model on CoQA, which was capable of capturing topic transfer features using the temporal convolutional network (TCN) in the dialog. The TT-Block packaged by the BiLSTM, TCN and Self-attention mechanism was presented to extract topic transfer features between questions, which was also an incremental modeling method. 

\subsubsection{Summarization}
The way of parallel modeling is more conducive to the screening of historical dialogue information, and the way of incremental modeling is more in line with the logical order of reasoning. Task characteristics and design ideas determine the modeling method. In table \ref{Modeling_CRC} we summarize some modeling methods of combining document(s), last question and context in the CRC.

\begin{table*}[h]
\caption{The modeling methods in the CRC. C means $\{Q_1, A_1, Q_2, A_2, \dots , Q_{n-1}, A_{n-1}\}$, i=\{1, 2, \dots, n-1\}.}
\label{Modeling_CRC}
\begin{tabular}{c|c|c}
\hline
Description & Method & Examples \\
\hline
Q and A sequentially & $\{C, Q_n\}, \{Doc\}$ & \cite{DBLP:journals/tacl/ReddyCM19,DBLP:journals/corr/abs-1812-03593,DBLP:journals/corr/abs-1909-10772,DBLP:conf/aaai/SuGFLZC19}\\
\hline
Doc, Q and A separately & $\{Doc\}, \{Q_n\}, \{Q_i\}, \{A_i\}$ &\cite{DBLP:conf/emnlp/ChoiHIYYCLZ18,DBLP:journals/corr/abs-1810-06683,DBLP:journals/corr/abs-1908-05117} \\
\hline
Doc and Q separately & $\{Doc\}, \{Q_i\}$ &
\cite{DBLP:journals/access/GuGL19,DBLP:journals/corr/abs-1908-00059}\\
\hline
Concatenating all & $\{C, Q_n, Doc\}$ & \cite{DBLP:conf/sigir/Qu0QCZI19,DBLP:conf/emnlp/SaeidiBL0RSB018,DBLP:conf/acl/ZhongZ19,DBLP:conf/emnlp/LawrenceKN19,DBLP:journals/corr/abs-1909-03759} \\
\hline
Doc paired with Q and A & $\{Q_n, Doc\}, \{Q_i, Doc\}, \{A_i, Doc\}$ & \cite{DBLP:journals/corr/abs-1905-12848} \\
\hline
Doc, latest Q and $\{Q_i, A_i\}$ & $\{Doc, Q_n, Q_i, A_i\}$ &\cite{DBLP:conf/cikm/QuYQZCCI19} \\
\hline
\end{tabular}
\end{table*}

\subsection{Knowledge Selection in the CRC}
As we defined in the architecture section, the KS in the CRC task can be divided into context-based and document(s)-based.

\subsubsection{Context-based}
\citet{DBLP:conf/sigir/Qu0QCZI19} introduced a general framework containing an utterances history selection module that retrieves a subset of dialog history more useful than others\footnote{Alough they simply select the nearest three utterances.}. They also presented a history answer embedding module to incorporate the utterance history naturally to BERT. \citet{DBLP:conf/cikm/QuYQZCCI19} further improved previous work \cite{DBLP:conf/sigir/Qu0QCZI19} with three aspects. Firstly, a positional history answer embedding (PosHAE) method based on location information was proposed. Secondly, a history attention mechanism (HAM) was used to "soft select" the utterance history. Thirdly, in addition to dealing with conversation history, multi-task learning (MTL) was used to predict the dialog act.

\subsubsection{Document(s)-based} In this section, we introduce the current CRC model of the KS for the document(s) from implicit and explicit reasoning perspectives.

\textbf{Implicit Reasoning.} As mentioned earlier, the FlowQA proposed by \citet{DBLP:journals/corr/abs-1810-06683} was an implicit reasoning module which uses historical dialogues to perform information flow in the document(s). The problem is that the learned representations captured by the FLOW change during multi-turn questions. Whether such changes correlated well with the current answer or not is unclear. To explicitly explore the information gain in FLOW and further relate the current answer to the corresponding context, \citet{DBLP:journals/corr/abs-1908-05117} presented FlowDelta, which focused on modeling the difference between the learned context representations in multi-turn dialogues. They add a subtraction of the previous hidden state into the flow section to mimic the conversation topic change between sequential turn questions. To capture the topic transfer (TT) information in CoQA, \citet{DBLP:journals/access/GuGL19} proposed a TT-Net model using TT-Block packaged by the BiLSTM, TCN, and Self-attention mechanism is presented to extract topic transfer features between questions. 

\textbf{Explicitly Reasoning.}
\citet{DBLP:journals/corr/abs-1908-00059} argued that the Integration-Flow mechanism in FlowQA fails to mimic the human reasoning process since humans do not first perform reasoning in parallel for each question, and then refined the reasoning results across different turns. The reasoning performance at each turn was only slightly improved by the hidden states of the previous reasoning process. So they built a sparse graph from the document and utterances history dynamically, then proposed a GraphFLOW (GF) structure to perform interpretable reasoning. The GF mechanism adopted a GNN to the sparse graph and fused both the original context information and the updated context information at the previous turn. The final prediction results were obtained through multiple rounds of GF operation on dialogues. On the ShARC task which takes reasoning ability as the main challenge, the E3 model \cite{DBLP:conf/acl/ZhongZ19} jointly extracted a set of decision rules from the procedural text while reasoning about which was entailed by the conversational history and which still needed to be edited to create questions for the user. To prevent models from learning the superficial clues, \citet{DBLP:journals/corr/abs-1909-03759} modified the ShARC dataset by automatically generating new instances reducing the occurrences of those patterns. They proved that \citet{DBLP:conf/acl/ZhongZ19} relied more heavily on spurious clues in the dataset and suffered a steeper drop in performance on the ShARC-augmented dataset.

\subsection{Response Generation in the CRC}
In the CRC, the indirect generation usually means determining the probability of each word as the starting and ending position of the answer. While in the direct generation, the decoder of the generative model generates the response word by word.

\subsubsection{Indirect Generation}
On the CoQA and the QuAC dataset, \citet{DBLP:journals/tacl/ReddyCM19} and \citet{DBLP:conf/emnlp/ChoiHIYYCLZ18} use DrQA and BiDAF++ for answer span prediction respectively. Models on CoQA and QuAC \cite{DBLP:journals/corr/abs-1812-03593,DBLP:conf/naacl/Yatskar19,DBLP:journals/corr/abs-1905-12848,DBLP:journals/corr/abs-1909-10772,DBLP:journals/corr/abs-1810-06683,DBLP:journals/corr/abs-1908-05117,DBLP:journals/corr/abs-1908-00059,DBLP:journals/access/GuGL19,DBLP:conf/sigir/Qu0QCZI19,DBLP:conf/cikm/QuYQZCCI19} all followed the span prediction setting.

\subsubsection{Direct Generation}
\citet{DBLP:journals/tacl/ReddyCM19} proposed a Pointer-Generator network (PGNet) model generating the answer using an RNN decoder which attends to the encoder states. They also employ a copy mechanism in the decoder which allows copying a word from the document. \citet{DBLP:conf/aaai/SuGFLZC19} adopted a seq2seq model as the DistillNet to refine the final answer. On the ShARC dataset, the questions to generate are derived from the background text. \citet{DBLP:conf/emnlp/SaeidiBL0RSB018} presented a Combined Model (CM) which is a pipeline that contains a rule-based follow-up question generation model. \citet{DBLP:conf/emnlp/LawrenceKN19} proposed bidirectional sequence generation model and introduced several different sequence generation strategies.

\subsubsection{Hybrid Generation}
\citet{DBLP:conf/acl/ZhongZ19} first selected a rule-span, then used a separate attentive decoder to generate the pre-span and the post-span words, the concatenation of the three parts constituted the final output.

\subsection{Evaluation Method in the CRC}
Since the CRC answers are usually a clear span or "YES/NO/UNANSWERABLE", the evaluation metrics are usually automatic. Besides EM (exact match), former work \cite{DBLP:conf/emnlp/ChoiHIYYCLZ18,DBLP:journals/corr/abs-1909-10772,DBLP:journals/corr/abs-1908-00059} employed a word-level $F1$ similar to \citet{DBLP:conf/emnlp/RajpurkarZLL16} to measure the performance of model. This metric regards the prediction and ground truth as bags of tokens and takes the maximum F1 over all of the ground truth answers for a given question, and then averages over all of the questions.

\citet{DBLP:conf/emnlp/ChoiHIYYCLZ18} proposed human equivalence score (HEQ). The HEQ-Q is the accuracy of each question, where the answer is considered correct when the model’s F1 score is higher than the average human F1 score. Similarly, the HEQ-D is the accuracy of each dialog – it is considered correct if all the questions in the dialog satisfy the HEQ. A system that scores the value of $100$ on the HEQ-D can by definition achieves average human performance over full dialogs. \citet{DBLP:journals/corr/abs-1810-06683,DBLP:journals/corr/abs-1905-12848,DBLP:journals/corr/abs-1908-00059} also used this metric.

On the ShARC task, for all following classification tasks, \citet{DBLP:conf/emnlp/SaeidiBL0RSB018} used micro- and macro- averaged accuracy. For the follow-up generation task, they computed the BLEU scores between the reference and generated follow-up questions. Then \citet{DBLP:conf/acl/ZhongZ19} showed a combined metric ("Comb."), which was the product between the macro-averaged accuracy and the BLEU-4 score on ShARC dataset.

We list the current model performance in the CRC in Table \ref{CoQA_F1}. We only show the single models which are published on the QuAC\footnote{http://quac.ai/} and the CoQA\footnote{https://stanfordnlp.github.io/coqa/} leaderboard. BERT + Answer Verification\footnote{https://github.com/sogou/SMRCToolkit} and XLNet \cite{DBLP:journals/corr/abs-1906-08237} + Augmentation\footnote{https://github.com/stevezheng23/xlnet\_extension\_tf} are added to the table\footnote{The top single model performance of the QuAC leaderboard is F1=73.5, HEQ-Q=69.8, HEQ-D=12.1 untill Beijing time, November 29, 2019 0:00.} for comprison. We also present the model performance on the ShARC\footnote{https://sharc-data.github.io/} dataset.

\begin{table*}
\caption{The model performance on the CoQA, QuAC, and ShARC Test set. * means less training time. }
\label{CoQA_F1}
\begin{tabular}{l|c|c|c|c|c|c}
\hline
\multirow{2}{*}{Model} &
\multicolumn{1}{c|}{CoQA} &
\multicolumn{3}{c|}{QuAC} &
\multicolumn{2}{c}{ShARC} \\
\cline{2-7}
& F1 & F1 & HEQ-Q & HEQ-D & (Micro/Macro)-Acc. & BLEU-(1/4) \\
\hline
DrQA+PGNet \cite{DBLP:journals/tacl/ReddyCM19} & 65.1 &- & -&- && \\
\hline
BiDAF++ w/ 2-ctx \cite{DBLP:conf/emnlp/ChoiHIYYCLZ18} & - & 60.1&54.8 &4.0&&\\
\hline
Bert+HAE* \cite{DBLP:conf/sigir/Qu0QCZI19} &- & 62.4 & 57.8 & 5.1&&\\
\hline
FlowQA \cite{DBLP:journals/corr/abs-1810-06683} & 75.0 & 64.1 & 59.6 & 5.8&&\\
\hline
HAM \cite{DBLP:conf/cikm/QuYQZCCI19} &- & 65.4& 61.8 & 6.7&&\\
\hline
SDNet \cite{DBLP:journals/corr/abs-1812-03593} & 76.6 & -& -& -&&\\
\hline
GraphFlow \cite{DBLP:journals/corr/abs-1908-00059} & 77.3 & 64.9 & 60.3 & 5.1&&\\
\hline
Bert-FlowDelta \cite{DBLP:journals/corr/abs-1908-05117} & 77.7 & 67.8 & 63.6 & 12.1&&\\
\hline
Bert w/ 2-ctx \cite{DBLP:journals/corr/abs-1905-12848} & 78.7 & 64.9 & 60.2 & 6.1 &&\\
\hline
BERT + Ans.Verification & 82.8 &- &- &- &&\\
\hline
XLNet + Augmentation & 89.0 &- &- &- &&\\
\hline
RoBerta + AT + KD \cite{DBLP:journals/corr/abs-1909-10772} & 90.4 &- &- &- &&\\
\hline
CM \cite{DBLP:conf/emnlp/SaeidiBL0RSB018}& & & & &0.619 / 0.689 & 54.4 / 34.4 \\
\hline
E3 \cite{DBLP:conf/acl/ZhongZ19}& & & & &0.676 / 0.733 & 54.1 / 38.7 \\
\hline
BiSon \cite{DBLP:conf/emnlp/LawrenceKN19}& & & & &0.669 / 0.716 & 58.8 / 44.3\\
\hline
UrcaNet \cite{DBLP:journals/corr/abs-1909-03759}& & & & &0.659 / 0.717 & 61.2 / 45.8 \\
\hline
Human & 88.8 & 81.1 & 100 & 100&&\\
\hline
\end{tabular}
\end{table*}

\section{DBD models}
In this chapter, we analyze the current DBD models in accordance with the architecture previously presented.

\subsection{Joint Modeling in the DBD}
Same as the CRC chapter, we introduce the JM of DBD from parallel and incremental as well.

\subsubsection{Parallel Modeling}

\citet{DBLP:conf/ijcai/ZhaoTWX0Y19} focused on connecting dialogue and background knowledge and identifying document information for matching. They used a retrieve-based method DGMN. By hierarchical interacting with the response, the importance of the different parts of the document and context was dynamically determined in the matching step. They conducted their experiments on the CMUDoG and Persona-chat \cite{DBLP:conf/acl/KielaWZDUS18} as retrieval tasks. \citet{DBLP:conf/naacl/AroraKR19} modeled the latest utterance with prober and responder history respectively, then used the enhanced last utterance to filter the document segments. \citet{DBLP:conf/acl/QinGBLGDCG19} adopted all utterance history as query to retrieve related text segments from document(s). \citet{DBLP:conf/ksem/TangH19} also employed a parallel way of encoding the text. They proposed three models: DialogTransformer which consisted of knowledge memory and response generation, DialogTransformer-Plus which used a three-stage multi-head attention mechanism to incorporate dialogue utterances and knowledge representation, and DialogTransformerX which combined different sources in the generation. Models \cite{DBLP:journals/corr/abs-1906-06685,DBLP:journals/corr/abs-1908-09528} focusing on KS all employed the parallel modeling methods. 

\subsubsection{Incremental Modeling}
Incremental means adding historical conversation, document and current utterance layer by layer. \citet{DBLP:conf/acl/LiNMFLZ19} proposed Incremental Transformer with Deliberation Decoder (ITEDD). They designed an incremental transformer to encode multi-turn utterances along with knowledge in the related document(s). One problem of ITEDD is the authors only experiment on the CMUDoG where the document(s) is pre-defined to sections, they did not consider the effect when the doc length increases and the deliberation decoder could not mine enough document(s) information.

\subsubsection{Summarization}

The advantage of incremental modeling is that it can reflect the temporal relationship of dialogue history, but it also loses the ability to filter dialogue history since long-distance dialogue information will be forgotten in encoding process, so the incremental modeling usually adopts the near rounds of dialogue, increasing the rounds of historical dialogue does not improve performance. In contrast, parallel modeling can better filter historical conversation information, but it needs to consider the influence of temporal relationship. In table \ref{parallel_DBD} we sum up the modeling categories in DBD.

\begin{table}[h]
\caption{The categories of modeling in DBD. D is Document, C=$\{U_1, \dots, U_{n-1}\}$ is the utterances history, Bot=$\{Bot_1, \dots, Bot_{n-1}\}$ is the utterances from Bot, User=$\{User_1, \dots, User_{n-1}\}$ is the utterances from User, i=\{1, 2, \dots, n-1\}.}
\label{parallel_DBD}
\begin{tabular}{c|c|c}
\hline
Description & Method & Examples \\
\hline
Concatenating all & $\{D, C, U_n\}$ &\cite{DBLP:conf/emnlp/ZhouPB18}\\
\hline
Ds and Us seperatly & $\{D\}, \{C, U_n\}$ & \cite{DBLP:journals/corr/abs-1906-06685,DBLP:conf/acl/QinGBLGDCG19,DBLP:journals/corr/abs-1908-09528,DBLP:journals/corr/abs-1908-06449,DBLP:conf/ijcai/ZhaoTWX0Y19,DBLP:journals/corr/abs-1903-10245}\\
\hline
User and Bot seperatly & $\{D\}, \{User_n\}, \{Bot\}, \{User\}$ &\cite{DBLP:conf/naacl/AroraKR19}\\
\hline
D, U, C seperatly & $\{D\}, \{U_n\}, \{C\}$ &\cite{DBLP:conf/ksem/TangH19}\\
\hline
D, U, $U_i$ seperatly & $\{D\}, \{U_n\}, \{U_i\}$ &\cite{DBLP:conf/acl/LiNMFLZ19}\\
\hline
\end{tabular}
\end{table}

\subsection{Knowledge Selection in the DBD}
Unlike the CRC, the KS in the DBD usually pays little attention on context while focus on document(s).
We divide the KS of the DBD into selection and reasoning.

\subsubsection{Selection}
In DGDS, \citet{DBLP:conf/acl/QinGBLGDCG19} presented an architecture combining MRC technique \cite{DBLP:conf/acl/GaoDLS18} into next utterance generation with external web-text. They pre-selected some candidate segments for encoding which improves training efficiency. To improve the process of using background knowledge, \citet{DBLP:journals/corr/abs-1906-06685} focused on generation-based methods and propose Context-aware Knowledge pre-selection (CaKe). The main contribution of CaKe is also a knowledge pre-selection module. They firstly adopted the encoder state of the utterance history context as a query to select the most relevant knowledge, then employed a modified version of BiDAF to point out the most relevant token positions of the background sequence. A limitation of CaKe is that the performance decreases when the background document becomes longer. 

\citet{DBLP:journals/corr/abs-1908-09528} argued that previous KS mechanisms have used a local perspective, i.e., choosing a token at a time based solely on the current decoding state. They adopted a global perspective, i.e., pre-selecting some text fragments from the background knowledge that could help determine the topic of the next response. Their model firstly learned a topic transition vector to encode the most likely text fragments to be used in the next response, then used the vector to guide the local KS at each decoding timestamp.

Models focusing on evidence selection \cite{DBLP:conf/acl/QinGBLGDCG19,DBLP:journals/corr/abs-1906-06685,DBLP:journals/corr/abs-1908-06449,DBLP:journals/corr/abs-1908-09528} and concentrating on feature extraction \cite{DBLP:conf/naacl/AroraKR19} have no explicit reasoning path, they are all belong to the selection method.

\subsubsection{Reasoning}

In the DGDS, \citet{DBLP:journals/corr/abs-1903-10245} claimed the first to unify knowledge triples and long texts as a graph. Then employed a reinforce learning process in the flexible multi-hop knowledge graph reasoning process, called AKGCM. They conducted experiments on Holl-E and WoW. One deficiency is that the reinforcement learning policy is doubtable since they chose a labeled state as a reward. They used the top $1$ accuracy to evaluate the performance of knowledge selection.

Approaches \cite{DBLP:conf/acl/LiNMFLZ19,DBLP:conf/ksem/TangH19} gradually integrating historical information during incremental modeling are also considered to be reasoning because of the existence of the implicit inference path.

\subsection{Response Generation in the DBD}
We analyze the RG from $3$ categories: the direct generation that generates the response word by word, the indirect generation that predicts a span or ranks candidates, and the hybrid generation that combines indirect and direct.

\subsubsection{Direct}
In order to improve the consistency of context and the correctness of knowledge, the ITEDD model \cite{DBLP:conf/acl/LiNMFLZ19} employed a two-way deliberation decoder \cite{DBLP:conf/nips/XiaTWLQYL17} for response generation. The first-level decoder took the representation of last utterance and last encoding state as input to generate responses contextual coherently. The second-level decoder took the first-level decoding results and the grounded document as input to guide the generation. The CMR model \cite{DBLP:conf/acl/QinGBLGDCG19} chose an attentional recurrent neural network decoder after a memory that summarized the salient information from both context and document. The DialogTransformer model \cite{DBLP:conf/ksem/TangH19} used a one-layer Transformer as decoder.

\subsubsection{Indirect}
\citet{DBLP:conf/ijcai/ZhaoTWX0Y19} presented a retrieval-based model that fused information in the document(s) and context into representations of each other. They dynamically determined whether the grounding was necessary, and weighted the importance of different parts of the document and context through hierarchical interaction with a response at the matching step. \citet{DBLP:conf/naacl/AroraKR19} regarded Holl-E as a span prediction task, they argued that models computing rich representations for the document(s) and utterance suffered space and time when dealing with long text. Hence they adopted a knowledge distillation method to train a simple model that learned to mimic certain characteristics of a complex span prediction teacher model.

\subsubsection{Hybrid}
\citet{DBLP:conf/acl/QinGBLGDCG19} firstly selected the document segment(s) then perform generation. The DialogTransformerX model \cite{DBLP:conf/ksem/TangH19} combines three methods for DG: (1) Generating a word, (2) Copying a word from dialogue utterance, (3) Copying a word from unstructured external knowledge. Combining the advantages of generative and extractive models. \citet{DBLP:journals/corr/abs-1908-06449} propose RefNet model. At the decoding step, a decoding switcher predicts the probabilities of executing the reference decoding or generation decoding. Reference decoder learns to directly select a semantic unit (e.g., a span containing complete semantic information) from the background while generation decoder predicts words one by one. One disadvantage of the RefNet is that it needs labeled data for span prediction, hence it is not a general model for DGDS. Another disadvantage is that the emphasis of the RefNet is on the decoding part, not much on the joint modeling part where the importance of different utterances history should be considered.

\subsection{Evaluation Method in the DBD}

At present, the evaluation indexes of the DGDS can be divided into two categories: one is manual evaluation, the other is objective auto evaluation.

\subsubsection{Human Evaluation}
We divide human evaluations into the grading method and the comparison method.

\textbf{Grading.} Workers were asked to rate the responses generated by the model. \citet{DBLP:conf/emnlp/MogheABK18} let workers rate the response on a scale of 1 to 5 (with 1 being the worst) on the following four metrics: (1) Fluency, (2) appropriateness/relevance of the response under the current context, (3) humanness of the response, i.e., whether the responses look as if they were generated by a human (4) and specificity of the response, i.e., whether the model produced movie-specific responses or generic responses such as “This movie is amazing”. \citet{DBLP:journals/corr/abs-1908-06449,DBLP:journals/corr/abs-1908-09528} had workers annotate whether the response was good in terms of four aspects: (1) Naturalness (N), i.e., whether the responses are conversational, natural and fluent; (2) Informativeness (I), i.e., whether the responses use some background information; (3) Appropriateness (A), i.e., whether the responses are appropriate/relevant to the given context; and (4) Humanness (H), i.e., whether the responses look like they are written by a human. \citet{DBLP:conf/acl/LiNMFLZ19} defined three metrics - fluency, knowledge relevance \cite{DBLP:conf/acl/LiuFCRYL18} and context coherence. All these metrics are scored $0/1/2$. Fluency is whether the response is natural and fluent. Knowledge relevance is whether the response uses relevant and correct knowledge. Context coherence is whether the response is coherent with the context and guides the following utterances. Human Judgment employed by \citet{DBLP:conf/ksem/TangH19} was to rate the response on a scale of $1$ to $5$ on the Fluency and Knowledge Capacity of the generated response. \citet{gopalakrishnan2019topical} asked two humans to separately annotate (possible values in parentheses) whether the response is comprehensible (0/1), on-topic (0/1), and interesting (0/1). They also asked workers to annotate how effectively the knowledge is used in response (0-3) and if they would like to continue the conversation after the generated response (0/1). 

\textbf{Comparison.} Workers were required to directly compare the generation response of the model with the generation of the reference target or other baseline models, and choose the preferred one. In human evaluation, \citet{DBLP:conf/acl/QinGBLGDCG19,DBLP:journals/corr/abs-1903-10245} tested the preferences between the response from their model and comparator model. Outputs from systems to be compared were presented pairwise to judges from a crowdsourcing service.

\subsubsection{Retrieval Auto Ealuation}

The ranking of candidate answers is the core of the retrieval DS. \citet{DBLP:conf/ijcai/ZhaoTWX0Y19} used R@N which evaluates whether the correct candidate is retrieved in the top N results. \citet{DBLP:journals/corr/abs-1903-10245} adopted a Hit@1 for the accuracy of selecting the right knowledge. 

\subsubsection{Generative Auto Evaluation}
In this section, we introduce some automatic evaluation metrics currently used in the DGDS.

\textbf{Perplexity \cite{DBLP:conf/nips/BengioDV00}}. In the language model, the Perplexity (PPL) is usually employed to measure the probability of the occurrence of a sentence. While in the DS, the PPL measures how well the model predicts a response, a lower perplexity score indicates better generation performance. The disadvantage of the PPL is that it can not evaluate the relevance between the response and the context.

\textbf{Entropy \cite{DBLP:conf/nips/ZhangGGGLBD18}}. Entropy reflects how evenly the empirical n-gram distribution is for a given sentence.

\textbf{Distinct \cite{DBLP:conf/naacl/LiGBGD16}}. Distinct measures the diversity of reply by calculating the proportion of 1\&2-grams in the total number of generated words to solve the problem of universal reply in DS.

\textbf{F1 \cite{DBLP:conf/emnlp/RajpurkarZLL16}}. Word-level $F1$ treats the prediction and ground truth as bags of tokens and measures the average overlap between the prediction and ground truth answer.

\textbf{BLEU \cite{DBLP:conf/acl/PapineniRWZ02}}. Measuring the similarity by calculating the geometric average of the accuracy of n-gram between the generated response and the golden response.

\textbf{ROUGE \cite{lin2004rouge}}. ROUGE is based on the calculation of the recall rate of the longest common subsequence of generating response and the real one.

\textbf{METEOR \cite{DBLP:conf/wmt/LavieA07}}. In order to improve BLEU, METEOR further considers the alignment between the generated and the real responses. WordNet is adopted to calculate the matching relationship among specific sequence matching, synonym, root, interpretation, etc.

\textbf{NIST \cite{doddington2002automatic}}. NIST is an improvement of BLEU by summing up each information weighted co-occurrence n-gram segments, then dividing it by the total number of n-gram segments.

There are some other metrics employed in the DGDS, for instance, \citet{DBLP:conf/acl/QinGBLGDCG19} computed a '\#match' score which is the number of non-stop word tokens in the response that are present in the document but not present in the context of the conversation, they also measure the average length of the utterance the CMR model generated. \citet{DBLP:conf/acl/ZhongZ19} showed a combined metric ("Comb."), which is the product between the macro-averaged accuracy and the BLEU-4 score on ShARC dataset. \citet{DBLP:journals/corr/abs-1902-00821} adopted Exact Match (EM) in all dialogues. EM requires the answers to have an exact string match with human-annotated answer spans.

\subsection{Summarization}

We present some evaluation metrics in the current DBD models in Table \ref{DBD_metrics}. It should be pointed out that these experimental values can only be used as a reference due to the differences in data processing and experimental environment, more constraints are needed for a fair comparison.

\begin{table}[h]
\caption{The metric employed in the DBD models. * means the test set has two version (Frequent/Rare), we only show the Rare version.}
\label{DBD_metrics}
\begin{tabular}{l|c|c|c|c|c|c}
\hline
Model & Dataset & F1 & BLEU-4 & ROUGE-(1/2/L) & Distinct-(1/2) & PPL \\
\hline
QANeT\cite{DBLP:conf/naacl/AroraKR19} & Holl-E & 47.67 & & & & \\
\hline
AKGCM\cite{DBLP:journals/corr/abs-1903-10245} & Holl-E && 30.84 & --- --- / 29.29 / 34.72 & & \\
\hline
CaKe\cite{DBLP:journals/corr/abs-1906-06685} & Holl-E & & 31.16 & 48.65 / 36.54 / 43.21 & & \\
\hline
RefNet\cite{DBLP:journals/corr/abs-1908-06449} & Holl-E & 48.81& 33.65& 49.64 / 38.15 / 43.77 & & \\
\hline
GLKS\cite{DBLP:journals/corr/abs-1908-09528} & Holl-E & & & 50.67 / 39.20 / 45.64 & & \\
\hline
BiDAF\cite{DBLP:conf/emnlp/MogheABK18} & Holl-E & 51.35 &39.39 &50.73 / 45.01 / 46.95 & & \\
\hline
SEQS\cite{DBLP:conf/emnlp/ZhouPB18} & CMUDoG & & & & & 10.11 \\
\hline
ITEDD\cite{DBLP:conf/acl/LiNMFLZ19} & CMUDoG & & 0.95 & & & 15.11 \\
\hline
DialogT\cite{DBLP:conf/ksem/TangH19} & CMUDoG & & 1.28 & & & 50.3 \\
\hline
TF\cite{gopalakrishnan2019topical} & T-Chat* & 0.20 & & &0.83 / 0.81& 43.6 \\
\hline
CMR\cite{DBLP:conf/acl/QinGBLGDCG19}& CbR & & 1.38 & & 0.052 / 0.283 & \\
\hline
\end{tabular}
\end{table}

\section{Future Work}
In this chapter, we discuss some fundamental and technical problems for the future development of the DGDS. Understanding fundamental problems can provide guidence for solving technical problems.

\subsection{Fundamental problems}
The fundamental problems of the DGDS are listed as below:

\begin{itemize}
\item {\textbf{What is the function of the DGDS?}} We divided the function of the DGDS into three levels: the first level is to mine information in document(s), such as the CRC; the second level is to use unstructured document(s) as external knowledge for generating more informative responses, such as the Holl-E; the third level is to take document(s) as discussion objects and express opinions on the contents of document(s), such as the CMUDoG. These three levels can be compared with the process of human learning, using and creating knowledge. The final form of the DGDS should be: the system can refer to the knowledge of the unstructured document(s) or express appropriate views on the document(s) without restraint, and maintain a conversation with users in line with the real human conversation scenario. 

\item {\textbf{How to verify the NLU problem?}} Do the DGDS models understand the dialogues and the document(s) or they just pick up some unified potential pattern to form a response? \citet{DBLP:journals/corr/abs-1909-10743} argue that the current CRC models on the QuAC and the CoQA do not well reflect comprehension on conversation content and cross sentence information is not that important. The same question exists in the DBD models. To verify whether the system understands the language and how the system performs the reasoning, we need the DGDS to show a reasonable path of exploring knowledge, rather than simply giving a dialogue reply. 

\item {\textbf{How to evaluate the NLG problem?}} At present, the NLG tasks usually focus on personalization, diversification, stylization, consistency, etc. We believe that one-to-many problems deserve more attention, especially in the DGDS where entities are restricted to a specific range. We define "one-to-many" as multiple replies involving different knowledge in the document(s) meet the requirements of the dialog context, which contrasts with the normal one-to-one setting. As shown in Table \ref{one2many}, the second utterance of the bot can be each of the four candidates. If candidates A/B/C is the references, D is the generated one, we should train the model to judge its rationality through the utterance history, rather than reduce the probability of its generation because it is different from the other three reference answers. The DGDS datasets \cite{DBLP:conf/emnlp/ChoiHIYYCLZ18,DBLP:conf/emnlp/MogheABK18} nowadays only set multiple references in the test set, failed to train the model with one-to-many properties. The one-to-many problems can better verify the system's understanding and application of natural language. Comparing with the common open domain DS, the diversity of reply in the DGDS can limit to a certain number of entities, which can more realistically achieve one-to-many training and evaluation. 

\item {\textbf{Lifelong learning problem.}} Most recently, the concept of lifelong learning in the machine learning system has been widely concerned, which requires that the deployed machine learning system continue to improve through interaction with the environment. The idea of lifelong learning is also applicable to the DGDS because it needs to solve the problem of transfer learning not only with different document sources but also with a different task. For example, when the document(s) are news reports, commodity reviews, or novels that come from various data collection sources, the data distribution, syntax structure are different, and the information may also be multimodal. The ShARC task and the CMUDoG task are of some unneglected distinction. Therefore, the DGDS needs to retrain and adjust its strategic components in the deployment process, so that it can automatically learn how to deal with the problem of multi-source heterogeneous data and the distinction between tasks that cannot be completely solved in the training. 
\end{itemize}

\subsection{Technical Problems}
The technical problems of the DGDS based on the fundamental challenges are listed below:

\subsubsection{Memory Ability} Document(s) and historical dialogues should be stored in a long-term memory way. The memory of the document(s) is helpful to judge the utilization of document information in the process of dialogue. The memory of historical dialogue could be used to judge the relationship between historical dialogue information and current dialogue. A multi-turn DS should maintain the memory ability of these two aspects at least, and should not re-model the document(s) and dialogue history every new dialog round. The use of memory ability in the model is not enough.

\citet{DBLP:journals/corr/abs-1903-10245} created an augmented KG with knowledge triples and long texts, which can be regarded as a memory component. They took a factoid knowledge graph (KG) as the backbone, and aligned unstructured sentences of non-factoid knowledge with the factoid KG by linking entities from these sentences to vertices (containing entities) of the KG, augmenting the factoid KG with non-factoid knowledge while retaining its graph structure. Apart from the ability to remember historical conversations and documents, in order to achieve the final form of DGDS, commonsense knowledge \cite{DBLP:conf/aaai/YoungCCZBH18,DBLP:conf/semeval/OstermannRMTP18,DBLP:conf/ijcai/ZhouYHZXZ18} should also exist in memory as a necessary component, so as to form a reasonable dialogue logic and make an appropriate response to the entity which never seen before. Furthermore, we need to store the knowledge representation for lifelong learning.

\subsubsection{Reasoning Ability} To verify the NLU, we need the system to be able to show an interpretable reasoning path. It is also the requirement of current AI ethics to be able to reasonably explain the choice of knowledge and reasoning path. Therefore, we believe that the future development trend will depend more on the graph structure which can clearly show the reasoning path and the reinforcement learning method which can explicitly stipulate the reward. There exist several different datasets that require reasoning in multiple steps in literature, for example the google BABI \cite{DBLP:journals/corr/WestonBCM15}, MultiRC \cite{DBLP:conf/naacl/KhashabiCRUR18}and Open-BookQA \cite{DBLP:conf/emnlp/MihaylovCKS18}, which are sentence-level reasoning. \citet{DBLP:journals/tacl/WelblSR18} and \citet{DBLP:conf/emnlp/Yang0ZBCSM18} introduced multi-document(s) RC dataset WikiHOP and HotpotQA which need multi-hop reasoning to integrate multi-evidence to locate the target. These studies and the research of the DGDS should benefit each other.

For example, when reasoning in long documents and multiple documents, we can benefit from the current graph-based MRC models. \citet{DBLP:journals/corr/abs-1809-02040} used graph convolutional network (GCN) and graph recurrent network (GRN) to better utilize global evidence in WikiHop\cite{DBLP:journals/tacl/WelblSR18}. \citet{DBLP:conf/naacl/CaoAT19} directly adopted candidates found in documents as GNN nodes and calculated classification scores over them. \citet{DBLP:conf/acl/TuWHTHZ19} introduced the HDE graph containing different types of query aware nodes that represented different granularity levels of information (candidates, documents, and entities) for a multi-choice task. They used GNN based message-passing algorithms to accumulate knowledge on the HDE graph. \citet{DBLP:conf/sigir/LuPRAWW19} proposed a KG based QUEST model that computes direct answers to complex questions by dynamically tapping arbitrary text sources and joining sub-results from multiple documents. 

\subsubsection{One-to-many Porblem} The one-to-one training model using golden reference loses its diversity and generalization, and the current evaluation metrics do not reflect the quality of the model well. In one-to-many training, we need to give appropriate rewards and scores to the generated responses that are not in the reference answers but are correct, which could assist the model to understand natural language and judge the generation ability of the model better. To address the "one-to-many" problem, we need to consider three aspects: (1) Dataset. The multiple reference replies should use different knowledge in the document(s). (2) Training loss function. A good training loss function should have inner relationships with evaluation methods to accomplish better performance. (3) Evaluation. This evaluation metric should meet two requirements: one is to be able to give a reasonable score when the generated reply and multiple reference replies respectively refer to different knowledge sources; the other is to be able to keep consistent with human evaluation. The current methods such as PPL are not suitable for one-to-many settings. Other metrics like BLEU can refer to multiple replies at the same time, but the scenarios in machine translation tasks are usually different from DS.

\subsubsection{Model Generalization} In the process of lifelong learning, we need to consider knowledge retention and knowledge transfer. Knowledge retention is defined as the retention of historical experience. Knowledge transfer is defined as the ability to take advantage of the historical experience when dealing with a new type of documents. This requires us to link the modeling process and memory ability, preserve experience knowledge, distinguish good experience from bad one, update the outdated knowledge, and establish new knowledge in the face of unseen tasks, etc. We believe that according to the characteristics of the DGDS tasks and different stages of processing, it is the future trend to build the model on multiple subtasks with multiple levels. Subtasks in different levels learn the commonness among different types of documents. Subtasks in the same level investigate the differences among all types of documents, for example, merchandise reviews and news report have different text structures, news report can also include multimedia information such as voice, picture, video, etc. There must be different sub-functions to solve different text structures and different data forms.

\section{Conclusion}
The Document Grounded Dialogue System (DGDS) can mine document(s) information and discuss specific document(s) in a real human conversation. We believe that extracting unstructured document(s) information in dialogue is the future trend of the DS because a large amount of human knowledge is contained in these document(s). The research of the DGDS not only possesses a broad application prospect but also facilitates the DS to better understand human knowledge and natural language. This article introduces the DGDS, defines the related concepts, analyzes the current datasets and models, and provides views on future research trends in this field, hoping to be helpful for the community.

\bibliographystyle{ACM-Reference-Format}
\bibliography{sample-base}


\begin{thebibliography}{141}


\ifx \showCODEN    \undefined \def \showCODEN     #1{\unskip}     \fi
\ifx \showDOI      \undefined \def \showDOI       #1{#1}\fi
\ifx \showISBNx    \undefined \def \showISBNx     #1{\unskip}     \fi
\ifx \showISBNxiii \undefined \def \showISBNxiii  #1{\unskip}     \fi
\ifx \showISSN     \undefined \def \showISSN      #1{\unskip}     \fi
\ifx \showLCCN     \undefined \def \showLCCN      #1{\unskip}     \fi
\ifx \shownote     \undefined \def \shownote      #1{#1}          \fi
\ifx \showarticletitle \undefined \def \showarticletitle #1{#1}   \fi
\ifx \showURL      \undefined \def \showURL       {\relax}        \fi
\providecommand\bibfield[2]{#2}
\providecommand\bibinfo[2]{#2}
\providecommand\natexlab[1]{#1}
\providecommand\showeprint[2][]{arXiv:#2}

\bibitem[\protect\citeauthoryear{Akasaki and Kaji}{Akasaki and Kaji}{2017}]%
        {DBLP:conf/acl/AkasakiK17}
\bibfield{author}{\bibinfo{person}{Satoshi Akasaki} {and}
  \bibinfo{person}{Nobuhiro Kaji}.} \bibinfo{year}{2017}\natexlab{}.
\newblock \showarticletitle{Chat Detection in an Intelligent Assistant:
  Combining Task-oriented and Non-task-oriented Spoken Dialogue Systems}. In
  \bibinfo{booktitle}{\emph{{ACL} {(1)}}}. \bibinfo{publisher}{Association for
  Computational Linguistics}, \bibinfo{pages}{1308--1319}.
\newblock


\bibitem[\protect\citeauthoryear{Akasaki and Kaji}{Akasaki and Kaji}{2019}]%
        {DBLP:conf/naacl/AkasakiK19}
\bibfield{author}{\bibinfo{person}{Satoshi Akasaki} {and}
  \bibinfo{person}{Nobuhiro Kaji}.} \bibinfo{year}{2019}\natexlab{}.
\newblock \showarticletitle{Conversation Initiation by Diverse News Contents
  Introduction}. In \bibinfo{booktitle}{\emph{{NAACL-HLT} {(1)}}}.
  \bibinfo{publisher}{Association for Computational Linguistics},
  \bibinfo{pages}{3988--3998}.
\newblock


\bibitem[\protect\citeauthoryear{Arora, Khapra, and Ramaswamy}{Arora
  et~al\mbox{.}}{2019}]%
        {DBLP:conf/naacl/AroraKR19}
\bibfield{author}{\bibinfo{person}{Siddhartha Arora},
  \bibinfo{person}{Mitesh~M. Khapra}, {and} \bibinfo{person}{Harish~G.
  Ramaswamy}.} \bibinfo{year}{2019}\natexlab{}.
\newblock \showarticletitle{On Knowledge distillation from complex networks for
  response prediction}. In \bibinfo{booktitle}{\emph{{NAACL-HLT} {(1)}}}.
  \bibinfo{publisher}{Association for Computational Linguistics},
  \bibinfo{pages}{3813--3822}.
\newblock


\bibitem[\protect\citeauthoryear{Bengio, Ducharme, and Vincent}{Bengio
  et~al\mbox{.}}{2000}]%
        {DBLP:conf/nips/BengioDV00}
\bibfield{author}{\bibinfo{person}{Yoshua Bengio},
  \bibinfo{person}{R{\'{e}}jean Ducharme}, {and} \bibinfo{person}{Pascal
  Vincent}.} \bibinfo{year}{2000}\natexlab{}.
\newblock \showarticletitle{A Neural Probabilistic Language Model}. In
  \bibinfo{booktitle}{\emph{{NIPS}}}. \bibinfo{publisher}{{MIT} Press},
  \bibinfo{pages}{932--938}.
\newblock


\bibitem[\protect\citeauthoryear{Cao, Aziz, and Titov}{Cao
  et~al\mbox{.}}{2019}]%
        {DBLP:conf/naacl/CaoAT19}
\bibfield{author}{\bibinfo{person}{Nicola~De Cao}, \bibinfo{person}{Wilker
  Aziz}, {and} \bibinfo{person}{Ivan Titov}.} \bibinfo{year}{2019}\natexlab{}.
\newblock \showarticletitle{Question Answering by Reasoning Across Documents
  with Graph Convolutional Networks}. In \bibinfo{booktitle}{\emph{{NAACL-HLT}
  {(1)}}}. \bibinfo{publisher}{Association for Computational Linguistics},
  \bibinfo{pages}{2306--2317}.
\newblock


\bibitem[\protect\citeauthoryear{Chen, Fisch, Weston, and Bordes}{Chen
  et~al\mbox{.}}{2017a}]%
        {DBLP:conf/acl/ChenFWB17}
\bibfield{author}{\bibinfo{person}{Danqi Chen}, \bibinfo{person}{Adam Fisch},
  \bibinfo{person}{Jason Weston}, {and} \bibinfo{person}{Antoine Bordes}.}
  \bibinfo{year}{2017}\natexlab{a}.
\newblock \showarticletitle{Reading Wikipedia to Answer Open-Domain Questions}.
  In \bibinfo{booktitle}{\emph{{ACL} {(1)}}}. \bibinfo{publisher}{Association
  for Computational Linguistics}, \bibinfo{pages}{1870--1879}.
\newblock


\bibitem[\protect\citeauthoryear{Chen, Liu, Yin, and Tang}{Chen
  et~al\mbox{.}}{2017b}]%
        {DBLP:journals/sigkdd/ChenLYT17}
\bibfield{author}{\bibinfo{person}{Hongshen Chen}, \bibinfo{person}{Xiaorui
  Liu}, \bibinfo{person}{Dawei Yin}, {and} \bibinfo{person}{Jiliang Tang}.}
  \bibinfo{year}{2017}\natexlab{b}.
\newblock \showarticletitle{A Survey on Dialogue Systems: Recent Advances and
  New Frontiers}.
\newblock \bibinfo{journal}{\emph{{SIGKDD} Explorations}} \bibinfo{volume}{19},
  \bibinfo{number}{2} (\bibinfo{year}{2017}), \bibinfo{pages}{25--35}.
\newblock


\bibitem[\protect\citeauthoryear{Chen, Wu, and Zaki}{Chen
  et~al\mbox{.}}{2019}]%
        {DBLP:journals/corr/abs-1908-00059}
\bibfield{author}{\bibinfo{person}{Yu Chen}, \bibinfo{person}{Lingfei Wu},
  {and} \bibinfo{person}{Mohammed~J. Zaki}.} \bibinfo{year}{2019}\natexlab{}.
\newblock \showarticletitle{GraphFlow: Exploiting Conversation Flow with Graph
  Neural Networks for Conversational Machine Comprehension}.
\newblock \bibinfo{journal}{\emph{CoRR}}  \bibinfo{volume}{abs/1908.00059}
  (\bibinfo{year}{2019}).
\newblock
\showeprint[arxiv]{1908.00059}
\urldef\tempurl%
\url{http://arxiv.org/abs/1908.00059}
\showURL{%
\tempurl}


\bibitem[\protect\citeauthoryear{Chiang, Ye, and Chen}{Chiang
  et~al\mbox{.}}{2019}]%
        {DBLP:journals/corr/abs-1909-10743}
\bibfield{author}{\bibinfo{person}{Ting{-}Rui Chiang},
  \bibinfo{person}{Hao{-}Tong Ye}, {and} \bibinfo{person}{Yun{-}Nung Chen}.}
  \bibinfo{year}{2019}\natexlab{}.
\newblock \showarticletitle{An Empirical Study of Content Understanding in
  Conversational Question Answering}.
\newblock \bibinfo{journal}{\emph{CoRR}}  \bibinfo{volume}{abs/1909.10743}
  (\bibinfo{year}{2019}).
\newblock
\showeprint[arxiv]{1909.10743}
\urldef\tempurl%
\url{http://arxiv.org/abs/1909.10743}
\showURL{%
\tempurl}


\bibitem[\protect\citeauthoryear{Cho, van Merrienboer, G{\"{u}}l{\c{c}}ehre,
  Bahdanau, Bougares, Schwenk, and Bengio}{Cho et~al\mbox{.}}{2014}]%
        {DBLP:conf/emnlp/ChoMGBBSB14}
\bibfield{author}{\bibinfo{person}{Kyunghyun Cho}, \bibinfo{person}{Bart van
  Merrienboer}, \bibinfo{person}{{\c{C}}aglar G{\"{u}}l{\c{c}}ehre},
  \bibinfo{person}{Dzmitry Bahdanau}, \bibinfo{person}{Fethi Bougares},
  \bibinfo{person}{Holger Schwenk}, {and} \bibinfo{person}{Yoshua Bengio}.}
  \bibinfo{year}{2014}\natexlab{}.
\newblock \showarticletitle{Learning Phrase Representations using {RNN}
  Encoder-Decoder for Statistical Machine Translation}. In
  \bibinfo{booktitle}{\emph{{EMNLP}}}. \bibinfo{publisher}{{ACL}},
  \bibinfo{pages}{1724--1734}.
\newblock


\bibitem[\protect\citeauthoryear{Choi, He, Iyyer, Yatskar, Yih, Choi, Liang,
  and Zettlemoyer}{Choi et~al\mbox{.}}{2018}]%
        {DBLP:conf/emnlp/ChoiHIYYCLZ18}
\bibfield{author}{\bibinfo{person}{Eunsol Choi}, \bibinfo{person}{He He},
  \bibinfo{person}{Mohit Iyyer}, \bibinfo{person}{Mark Yatskar},
  \bibinfo{person}{Wen{-}tau Yih}, \bibinfo{person}{Yejin Choi},
  \bibinfo{person}{Percy Liang}, {and} \bibinfo{person}{Luke Zettlemoyer}.}
  \bibinfo{year}{2018}\natexlab{}.
\newblock \showarticletitle{QuAC: Question Answering in Context}. In
  \bibinfo{booktitle}{\emph{{EMNLP}}}. \bibinfo{publisher}{Association for
  Computational Linguistics}, \bibinfo{pages}{2174--2184}.
\newblock


\bibitem[\protect\citeauthoryear{Christmann, Roy, Abujabal, Singh, and
  Weikum}{Christmann et~al\mbox{.}}{2019}]%
        {DBLP:conf/cikm/ChristmannRASW19}
\bibfield{author}{\bibinfo{person}{Philipp Christmann},
  \bibinfo{person}{Rishiraj~Saha Roy}, \bibinfo{person}{Abdalghani Abujabal},
  \bibinfo{person}{Jyotsna Singh}, {and} \bibinfo{person}{Gerhard Weikum}.}
  \bibinfo{year}{2019}\natexlab{}.
\newblock \showarticletitle{Look before you Hop: Conversational Question
  Answering over Knowledge Graphs Using Judicious Context Expansion}. In
  \bibinfo{booktitle}{\emph{Proceedings of the 28th {ACM} International
  Conference on Information and Knowledge Management, {CIKM} 2019, Beijing,
  China, November 3-7, 2019}}, \bibfield{editor}{\bibinfo{person}{Wenwu Zhu},
  \bibinfo{person}{Dacheng Tao}, \bibinfo{person}{Xueqi Cheng},
  \bibinfo{person}{Peng Cui}, \bibinfo{person}{Elke~A. Rundensteiner},
  \bibinfo{person}{David Carmel}, \bibinfo{person}{Qi~He}, {and}
  \bibinfo{person}{Jeffrey~Xu Yu}} (Eds.). \bibinfo{publisher}{{ACM}},
  \bibinfo{pages}{729--738}.
\newblock
\urldef\tempurl%
\url{https://doi.org/10.1145/3357384.3358016}
\showDOI{\tempurl}


\bibitem[\protect\citeauthoryear{Clark and Gardner}{Clark and Gardner}{2018}]%
        {DBLP:conf/acl/GardnerC18}
\bibfield{author}{\bibinfo{person}{Christopher Clark} {and}
  \bibinfo{person}{Matt Gardner}.} \bibinfo{year}{2018}\natexlab{}.
\newblock \showarticletitle{Simple and Effective Multi-Paragraph Reading
  Comprehension}. In \bibinfo{booktitle}{\emph{{ACL} {(1)}}}.
  \bibinfo{publisher}{Association for Computational Linguistics},
  \bibinfo{pages}{845--855}.
\newblock


\bibitem[\protect\citeauthoryear{Clark}{Clark}{2015}]%
        {DBLP:conf/aaai/Clark15}
\bibfield{author}{\bibinfo{person}{Peter Clark}.}
  \bibinfo{year}{2015}\natexlab{}.
\newblock \showarticletitle{Elementary School Science and Math Tests as a
  Driver for {AI:} Take the Aristo Challenge!}. In
  \bibinfo{booktitle}{\emph{{AAAI}}}. \bibinfo{publisher}{{AAAI} Press},
  \bibinfo{pages}{4019--4021}.
\newblock


\bibitem[\protect\citeauthoryear{Colby, Weber, and Hilf}{Colby
  et~al\mbox{.}}{1971}]%
        {DBLP:journals/ai/ColbyWH71}
\bibfield{author}{\bibinfo{person}{Kenneth~Mark Colby}, \bibinfo{person}{Sylvia
  Weber}, {and} \bibinfo{person}{Franklin~Dennis Hilf}.}
  \bibinfo{year}{1971}\natexlab{}.
\newblock \showarticletitle{Artificial Paranoia}.
\newblock \bibinfo{journal}{\emph{Artif. Intell.}} \bibinfo{volume}{2},
  \bibinfo{number}{1} (\bibinfo{year}{1971}), \bibinfo{pages}{1--25}.
\newblock


\bibitem[\protect\citeauthoryear{Deriu, Rodrigo, Otegi, Echegoyen, Rosset,
  Agirre, and Cieliebak}{Deriu et~al\mbox{.}}{2019}]%
        {DBLP:journals/corr/abs-1905-04071}
\bibfield{author}{\bibinfo{person}{Jan Deriu}, \bibinfo{person}{{\'{A}}lvaro
  Rodrigo}, \bibinfo{person}{Arantxa Otegi}, \bibinfo{person}{Guillermo
  Echegoyen}, \bibinfo{person}{Sophie Rosset}, \bibinfo{person}{Eneko Agirre},
  {and} \bibinfo{person}{Mark Cieliebak}.} \bibinfo{year}{2019}\natexlab{}.
\newblock \showarticletitle{Survey on Evaluation Methods for Dialogue Systems}.
\newblock \bibinfo{journal}{\emph{CoRR}}  \bibinfo{volume}{abs/1905.04071}
  (\bibinfo{year}{2019}).
\newblock
\showeprint[arxiv]{1905.04071}
\urldef\tempurl%
\url{http://arxiv.org/abs/1905.04071}
\showURL{%
\tempurl}


\bibitem[\protect\citeauthoryear{Devlin, Chang, Lee, and Toutanova}{Devlin
  et~al\mbox{.}}{2019}]%
        {DBLP:conf/naacl/DevlinCLT19}
\bibfield{author}{\bibinfo{person}{Jacob Devlin}, \bibinfo{person}{Ming{-}Wei
  Chang}, \bibinfo{person}{Kenton Lee}, {and} \bibinfo{person}{Kristina
  Toutanova}.} \bibinfo{year}{2019}\natexlab{}.
\newblock \showarticletitle{{BERT:} Pre-training of Deep Bidirectional
  Transformers for Language Understanding}. In
  \bibinfo{booktitle}{\emph{Proceedings of the 2019 Conference of the North
  American Chapter of the Association for Computational Linguistics: Human
  Language Technologies, {NAACL-HLT} 2019, Minneapolis, MN, USA, June 2-7,
  2019, Volume 1 (Long and Short Papers)}},
  \bibfield{editor}{\bibinfo{person}{Jill Burstein}, \bibinfo{person}{Christy
  Doran}, {and} \bibinfo{person}{Thamar Solorio}} (Eds.).
  \bibinfo{publisher}{Association for Computational Linguistics},
  \bibinfo{pages}{4171--4186}.
\newblock
\urldef\tempurl%
\url{https://www.aclweb.org/anthology/N19-1423/}
\showURL{%
\tempurl}


\bibitem[\protect\citeauthoryear{Dinan, Roller, Shuster, Fan, Auli, and
  Weston}{Dinan et~al\mbox{.}}{2018}]%
        {DBLP:journals/corr/abs-1811-01241}
\bibfield{author}{\bibinfo{person}{Emily Dinan}, \bibinfo{person}{Stephen
  Roller}, \bibinfo{person}{Kurt Shuster}, \bibinfo{person}{Angela Fan},
  \bibinfo{person}{Michael Auli}, {and} \bibinfo{person}{Jason Weston}.}
  \bibinfo{year}{2018}\natexlab{}.
\newblock \showarticletitle{Wizard of Wikipedia: Knowledge-Powered
  Conversational agents}.
\newblock \bibinfo{journal}{\emph{CoRR}}  \bibinfo{volume}{abs/1811.01241}
  (\bibinfo{year}{2018}).
\newblock
\showeprint[arxiv]{1811.01241}
\urldef\tempurl%
\url{http://arxiv.org/abs/1811.01241}
\showURL{%
\tempurl}


\bibitem[\protect\citeauthoryear{Doddington}{Doddington}{2002}]%
        {doddington2002automatic}
\bibfield{author}{\bibinfo{person}{George Doddington}.}
  \bibinfo{year}{2002}\natexlab{}.
\newblock \showarticletitle{Automatic evaluation of machine translation quality
  using n-gram co-occurrence statistics}. In
  \bibinfo{booktitle}{\emph{Proceedings of the second international conference
  on Human Language Technology Research}}. Morgan Kaufmann Publishers Inc.,
  \bibinfo{pages}{138--145}.
\newblock


\bibitem[\protect\citeauthoryear{Dodge, Gane, Zhang, Bordes, Chopra, Miller,
  Szlam, and Weston}{Dodge et~al\mbox{.}}{2016}]%
        {DBLP:journals/corr/DodgeGZBCMSW15}
\bibfield{author}{\bibinfo{person}{Jesse Dodge}, \bibinfo{person}{Andreea
  Gane}, \bibinfo{person}{Xiang Zhang}, \bibinfo{person}{Antoine Bordes},
  \bibinfo{person}{Sumit Chopra}, \bibinfo{person}{Alexander~H. Miller},
  \bibinfo{person}{Arthur Szlam}, {and} \bibinfo{person}{Jason Weston}.}
  \bibinfo{year}{2016}\natexlab{}.
\newblock \showarticletitle{Evaluating Prerequisite Qualities for Learning
  End-to-End Dialog Systems}. In \bibinfo{booktitle}{\emph{4th International
  Conference on Learning Representations, {ICLR} 2016, San Juan, Puerto Rico,
  May 2-4, 2016, Conference Track Proceedings}},
  \bibfield{editor}{\bibinfo{person}{Yoshua Bengio} {and} \bibinfo{person}{Yann
  LeCun}} (Eds.).
\newblock
\urldef\tempurl%
\url{http://arxiv.org/abs/1511.06931}
\showURL{%
\tempurl}


\bibitem[\protect\citeauthoryear{Dua, Wang, Dasigi, Stanovsky, Singh, and
  Gardner}{Dua et~al\mbox{.}}{2019}]%
        {DBLP:conf/naacl/DuaWDSS019}
\bibfield{author}{\bibinfo{person}{Dheeru Dua}, \bibinfo{person}{Yizhong Wang},
  \bibinfo{person}{Pradeep Dasigi}, \bibinfo{person}{Gabriel Stanovsky},
  \bibinfo{person}{Sameer Singh}, {and} \bibinfo{person}{Matt Gardner}.}
  \bibinfo{year}{2019}\natexlab{}.
\newblock \showarticletitle{{DROP:} {A} Reading Comprehension Benchmark
  Requiring Discrete Reasoning Over Paragraphs}. In
  \bibinfo{booktitle}{\emph{{NAACL-HLT} {(1)}}}.
  \bibinfo{publisher}{Association for Computational Linguistics},
  \bibinfo{pages}{2368--2378}.
\newblock


\bibitem[\protect\citeauthoryear{Elgohary, Zhao, and Boyd{-}Graber}{Elgohary
  et~al\mbox{.}}{2018}]%
        {DBLP:conf/emnlp/ElgoharyZB18}
\bibfield{author}{\bibinfo{person}{Ahmed Elgohary}, \bibinfo{person}{Chen
  Zhao}, {and} \bibinfo{person}{Jordan~L. Boyd{-}Graber}.}
  \bibinfo{year}{2018}\natexlab{}.
\newblock \showarticletitle{A dataset and baselines for sequential open-domain
  question answering}. In \bibinfo{booktitle}{\emph{{EMNLP}}}.
  \bibinfo{publisher}{Association for Computational Linguistics},
  \bibinfo{pages}{1077--1083}.
\newblock


\bibitem[\protect\citeauthoryear{Fischer and Grodzinsky}{Fischer and
  Grodzinsky}{1993}]%
        {DBLP:books/daglib/0072558}
\bibfield{author}{\bibinfo{person}{Alice~E. Fischer} {and}
  \bibinfo{person}{Frances~S. Grodzinsky}.} \bibinfo{year}{1993}\natexlab{}.
\newblock \bibinfo{booktitle}{\emph{The anatomy of programming languages}}.
\newblock \bibinfo{publisher}{Prentice Hall}.
\newblock


\bibitem[\protect\citeauthoryear{Fu, Chang, and Danescu{-}Niculescu{-}Mizil}{Fu
  et~al\mbox{.}}{2019}]%
        {DBLP:conf/naacl/FuCD19}
\bibfield{author}{\bibinfo{person}{Liye Fu}, \bibinfo{person}{Jonathan~P.
  Chang}, {and} \bibinfo{person}{Cristian Danescu{-}Niculescu{-}Mizil}.}
  \bibinfo{year}{2019}\natexlab{}.
\newblock \showarticletitle{Asking the Right Question: Inferring Advice-Seeking
  Intentions from Personal Narratives}. In
  \bibinfo{booktitle}{\emph{{NAACL-HLT} {(1)}}}.
  \bibinfo{publisher}{Association for Computational Linguistics},
  \bibinfo{pages}{528--541}.
\newblock


\bibitem[\protect\citeauthoryear{Furlanello, Lipton, Tschannen, Itti, and
  Anandkumar}{Furlanello et~al\mbox{.}}{2018}]%
        {DBLP:conf/icml/FurlanelloLTIA18}
\bibfield{author}{\bibinfo{person}{Tommaso Furlanello},
  \bibinfo{person}{Zachary~Chase Lipton}, \bibinfo{person}{Michael Tschannen},
  \bibinfo{person}{Laurent Itti}, {and} \bibinfo{person}{Anima Anandkumar}.}
  \bibinfo{year}{2018}\natexlab{}.
\newblock \showarticletitle{Born-Again Neural Networks}. In
  \bibinfo{booktitle}{\emph{{ICML}}} \emph{(\bibinfo{series}{Proceedings of
  Machine Learning Research})}, Vol.~\bibinfo{volume}{80}.
  \bibinfo{publisher}{{PMLR}}, \bibinfo{pages}{1602--1611}.
\newblock


\bibitem[\protect\citeauthoryear{Gao, Galley, and Li}{Gao
  et~al\mbox{.}}{2019}]%
        {DBLP:journals/ftir/GaoGL19}
\bibfield{author}{\bibinfo{person}{Jianfeng Gao}, \bibinfo{person}{Michel
  Galley}, {and} \bibinfo{person}{Lihong Li}.} \bibinfo{year}{2019}\natexlab{}.
\newblock \showarticletitle{Neural Approaches to Conversational {AI}}.
\newblock \bibinfo{journal}{\emph{Foundations and Trends in Information
  Retrieval}} \bibinfo{volume}{13}, \bibinfo{number}{2-3}
  (\bibinfo{year}{2019}), \bibinfo{pages}{127--298}.
\newblock


\bibitem[\protect\citeauthoryear{Ghazvininejad, Brockett, Chang, Dolan, Gao,
  Yih, and Galley}{Ghazvininejad et~al\mbox{.}}{2018}]%
        {DBLP:conf/aaai/GhazvininejadBC18}
\bibfield{author}{\bibinfo{person}{Marjan Ghazvininejad},
  \bibinfo{person}{Chris Brockett}, \bibinfo{person}{Ming{-}Wei Chang},
  \bibinfo{person}{Bill Dolan}, \bibinfo{person}{Jianfeng Gao},
  \bibinfo{person}{Wen{-}tau Yih}, {and} \bibinfo{person}{Michel Galley}.}
  \bibinfo{year}{2018}\natexlab{}.
\newblock \showarticletitle{A Knowledge-Grounded Neural Conversation Model}. In
  \bibinfo{booktitle}{\emph{{AAAI}}}. \bibinfo{publisher}{{AAAI} Press},
  \bibinfo{pages}{5110--5117}.
\newblock


\bibitem[\protect\citeauthoryear{Goodfellow, Shlens, and Szegedy}{Goodfellow
  et~al\mbox{.}}{2015}]%
        {DBLP:journals/corr/GoodfellowSS14}
\bibfield{author}{\bibinfo{person}{Ian~J. Goodfellow},
  \bibinfo{person}{Jonathon Shlens}, {and} \bibinfo{person}{Christian
  Szegedy}.} \bibinfo{year}{2015}\natexlab{}.
\newblock \showarticletitle{Explaining and Harnessing Adversarial Examples}. In
  \bibinfo{booktitle}{\emph{3rd International Conference on Learning
  Representations, {ICLR} 2015, San Diego, CA, USA, May 7-9, 2015, Conference
  Track Proceedings}}, \bibfield{editor}{\bibinfo{person}{Yoshua Bengio} {and}
  \bibinfo{person}{Yann LeCun}} (Eds.).
\newblock
\urldef\tempurl%
\url{http://arxiv.org/abs/1412.6572}
\showURL{%
\tempurl}


\bibitem[\protect\citeauthoryear{Gopalakrishnan, Hedayatnia, Chen, Gottardi,
  Kwatra, Venkatesh, Gabriel, Hakkani-T{\"u}r, and AI}{Gopalakrishnan
  et~al\mbox{.}}{2019}]%
        {gopalakrishnan2019topical}
\bibfield{author}{\bibinfo{person}{Karthik Gopalakrishnan},
  \bibinfo{person}{Behnam Hedayatnia}, \bibinfo{person}{Qinlang Chen},
  \bibinfo{person}{Anna Gottardi}, \bibinfo{person}{Sanjeev Kwatra},
  \bibinfo{person}{Anu Venkatesh}, \bibinfo{person}{Raefer Gabriel},
  \bibinfo{person}{Dilek Hakkani-T{\"u}r}, {and} \bibinfo{person}{Amazon~Alexa
  AI}.} \bibinfo{year}{2019}\natexlab{}.
\newblock \showarticletitle{Topical-Chat: Towards Knowledge-Grounded
  Open-Domain Conversations}.
\newblock \bibinfo{journal}{\emph{Proc. Interspeech 2019}}
  (\bibinfo{year}{2019}), \bibinfo{pages}{1891--1895}.
\newblock


\bibitem[\protect\citeauthoryear{Gu, Lu, Li, and Li}{Gu et~al\mbox{.}}{2016}]%
        {DBLP:conf/acl/GuLLL16}
\bibfield{author}{\bibinfo{person}{Jiatao Gu}, \bibinfo{person}{Zhengdong Lu},
  \bibinfo{person}{Hang Li}, {and} \bibinfo{person}{Victor O.~K. Li}.}
  \bibinfo{year}{2016}\natexlab{}.
\newblock \showarticletitle{Incorporating Copying Mechanism in
  Sequence-to-Sequence Learning}. In \bibinfo{booktitle}{\emph{{ACL} {(1)}}}.
  \bibinfo{publisher}{The Association for Computer Linguistics}.
\newblock


\bibitem[\protect\citeauthoryear{Gu, Gui, and Lin}{Gu et~al\mbox{.}}{2019}]%
        {DBLP:journals/access/GuGL19}
\bibfield{author}{\bibinfo{person}{Yingjie Gu}, \bibinfo{person}{Xiaolin Gui},
  {and} \bibinfo{person}{Defu Lin}.} \bibinfo{year}{2019}\natexlab{}.
\newblock \showarticletitle{TT-Net: Topic Transfer-Based Neural Network for
  Conversational Reading Comprehension}.
\newblock \bibinfo{journal}{\emph{{IEEE} Access}}  \bibinfo{volume}{7}
  (\bibinfo{year}{2019}), \bibinfo{pages}{116696--116705}.
\newblock


\bibitem[\protect\citeauthoryear{Guo, Wang, Ding, Hao, Sun, and Yu}{Guo
  et~al\mbox{.}}{2019}]%
        {DBLP:journals/corr/abs-1909-03409}
\bibfield{author}{\bibinfo{person}{Bin Guo}, \bibinfo{person}{Hao Wang},
  \bibinfo{person}{Yasan Ding}, \bibinfo{person}{Shaoyang Hao},
  \bibinfo{person}{Yueqi Sun}, {and} \bibinfo{person}{Zhiwen Yu}.}
  \bibinfo{year}{2019}\natexlab{}.
\newblock \showarticletitle{c-TextGen: Conditional Text Generation for
  Harmonious Human-Machine Interaction}.
\newblock \bibinfo{journal}{\emph{CoRR}}  \bibinfo{volume}{abs/1909.03409}
  (\bibinfo{year}{2019}).
\newblock
\showeprint[arxiv]{1909.03409}
\urldef\tempurl%
\url{http://arxiv.org/abs/1909.03409}
\showURL{%
\tempurl}


\bibitem[\protect\citeauthoryear{Guo, Tang, Duan, Zhou, and Yin}{Guo
  et~al\mbox{.}}{2018}]%
        {DBLP:conf/nips/GuoTDZY18}
\bibfield{author}{\bibinfo{person}{Daya Guo}, \bibinfo{person}{Duyu Tang},
  \bibinfo{person}{Nan Duan}, \bibinfo{person}{Ming Zhou}, {and}
  \bibinfo{person}{Jian Yin}.} \bibinfo{year}{2018}\natexlab{}.
\newblock \showarticletitle{Dialog-to-Action: Conversational Question Answering
  Over a Large-Scale Knowledge Base}. In \bibinfo{booktitle}{\emph{NeurIPS}}.
  \bibinfo{pages}{2946--2955}.
\newblock


\bibitem[\protect\citeauthoryear{Hermann, Kocisk{\'{y}}, Grefenstette,
  Espeholt, Kay, Suleyman, and Blunsom}{Hermann et~al\mbox{.}}{2015}]%
        {DBLP:conf/nips/HermannKGEKSB15}
\bibfield{author}{\bibinfo{person}{Karl~Moritz Hermann},
  \bibinfo{person}{Tom{\'{a}}s Kocisk{\'{y}}}, \bibinfo{person}{Edward
  Grefenstette}, \bibinfo{person}{Lasse Espeholt}, \bibinfo{person}{Will Kay},
  \bibinfo{person}{Mustafa Suleyman}, {and} \bibinfo{person}{Phil Blunsom}.}
  \bibinfo{year}{2015}\natexlab{}.
\newblock \showarticletitle{Teaching Machines to Read and Comprehend}. In
  \bibinfo{booktitle}{\emph{{NIPS}}}. \bibinfo{pages}{1693--1701}.
\newblock


\bibitem[\protect\citeauthoryear{Hill, Bordes, Chopra, and Weston}{Hill
  et~al\mbox{.}}{2016}]%
        {DBLP:journals/corr/HillBCW15}
\bibfield{author}{\bibinfo{person}{Felix Hill}, \bibinfo{person}{Antoine
  Bordes}, \bibinfo{person}{Sumit Chopra}, {and} \bibinfo{person}{Jason
  Weston}.} \bibinfo{year}{2016}\natexlab{}.
\newblock \showarticletitle{The Goldilocks Principle: Reading Children's Books
  with Explicit Memory Representations}. In \bibinfo{booktitle}{\emph{4th
  International Conference on Learning Representations, {ICLR} 2016, San Juan,
  Puerto Rico, May 2-4, 2016, Conference Track Proceedings}},
  \bibfield{editor}{\bibinfo{person}{Yoshua Bengio} {and} \bibinfo{person}{Yann
  LeCun}} (Eds.).
\newblock
\urldef\tempurl%
\url{http://arxiv.org/abs/1511.02301}
\showURL{%
\tempurl}


\bibitem[\protect\citeauthoryear{Huang, Choi, and Yih}{Huang
  et~al\mbox{.}}{2018}]%
        {DBLP:journals/corr/abs-1810-06683}
\bibfield{author}{\bibinfo{person}{Hsin{-}Yuan Huang}, \bibinfo{person}{Eunsol
  Choi}, {and} \bibinfo{person}{Wen{-}tau Yih}.}
  \bibinfo{year}{2018}\natexlab{}.
\newblock \showarticletitle{FlowQA: Grasping Flow in History for Conversational
  Machine Comprehension}.
\newblock \bibinfo{journal}{\emph{CoRR}}  \bibinfo{volume}{abs/1810.06683}
  (\bibinfo{year}{2018}).
\newblock
\showeprint[arxiv]{1810.06683}
\urldef\tempurl%
\url{http://arxiv.org/abs/1810.06683}
\showURL{%
\tempurl}


\bibitem[\protect\citeauthoryear{Huang, Zhu, and Gao}{Huang
  et~al\mbox{.}}{2019}]%
        {DBLP:journals/corr/abs-1905-05709}
\bibfield{author}{\bibinfo{person}{Minlie Huang}, \bibinfo{person}{Xiaoyan
  Zhu}, {and} \bibinfo{person}{Jianfeng Gao}.} \bibinfo{year}{2019}\natexlab{}.
\newblock \showarticletitle{Challenges in Building Intelligent Open-domain
  Dialog Systems}.
\newblock \bibinfo{journal}{\emph{CoRR}}  \bibinfo{volume}{abs/1905.05709}
  (\bibinfo{year}{2019}).
\newblock
\showeprint[arxiv]{1905.05709}
\urldef\tempurl%
\url{http://arxiv.org/abs/1905.05709}
\showURL{%
\tempurl}


\bibitem[\protect\citeauthoryear{Iyyer, Yih, and Chang}{Iyyer
  et~al\mbox{.}}{2017}]%
        {DBLP:conf/acl/IyyerYC17}
\bibfield{author}{\bibinfo{person}{Mohit Iyyer}, \bibinfo{person}{Wen{-}tau
  Yih}, {and} \bibinfo{person}{Ming{-}Wei Chang}.}
  \bibinfo{year}{2017}\natexlab{}.
\newblock \showarticletitle{Search-based Neural Structured Learning for
  Sequential Question Answering}. In \bibinfo{booktitle}{\emph{{ACL} {(1)}}}.
  \bibinfo{publisher}{Association for Computational Linguistics},
  \bibinfo{pages}{1821--1831}.
\newblock


\bibitem[\protect\citeauthoryear{Ju, Zhao, Chen, Zheng, Yang, and Liu}{Ju
  et~al\mbox{.}}{2019}]%
        {DBLP:journals/corr/abs-1909-10772}
\bibfield{author}{\bibinfo{person}{Ying Ju}, \bibinfo{person}{Fubang Zhao},
  \bibinfo{person}{Shijie Chen}, \bibinfo{person}{Bowen Zheng},
  \bibinfo{person}{Xuefeng Yang}, {and} \bibinfo{person}{Yunfeng Liu}.}
  \bibinfo{year}{2019}\natexlab{}.
\newblock \showarticletitle{Technical report on Conversational Question
  Answering}.
\newblock \bibinfo{journal}{\emph{CoRR}}  \bibinfo{volume}{abs/1909.10772}
  (\bibinfo{year}{2019}).
\newblock
\showeprint[arxiv]{1909.10772}
\urldef\tempurl%
\url{http://arxiv.org/abs/1909.10772}
\showURL{%
\tempurl}


\bibitem[\protect\citeauthoryear{Kannan and Vinyals}{Kannan and
  Vinyals}{2017}]%
        {DBLP:journals/corr/KannanV17}
\bibfield{author}{\bibinfo{person}{Anjuli Kannan} {and} \bibinfo{person}{Oriol
  Vinyals}.} \bibinfo{year}{2017}\natexlab{}.
\newblock \showarticletitle{Adversarial Evaluation of Dialogue Models}.
\newblock \bibinfo{journal}{\emph{CoRR}}  \bibinfo{volume}{abs/1701.08198}
  (\bibinfo{year}{2017}).
\newblock
\showeprint[arxiv]{1701.08198}
\urldef\tempurl%
\url{http://arxiv.org/abs/1701.08198}
\showURL{%
\tempurl}


\bibitem[\protect\citeauthoryear{Khashabi, Chaturvedi, Roth, Upadhyay, and
  Roth}{Khashabi et~al\mbox{.}}{2018}]%
        {DBLP:conf/naacl/KhashabiCRUR18}
\bibfield{author}{\bibinfo{person}{Daniel Khashabi}, \bibinfo{person}{Snigdha
  Chaturvedi}, \bibinfo{person}{Michael Roth}, \bibinfo{person}{Shyam
  Upadhyay}, {and} \bibinfo{person}{Dan Roth}.}
  \bibinfo{year}{2018}\natexlab{}.
\newblock \showarticletitle{Looking Beyond the Surface: {A} Challenge Set for
  Reading Comprehension over Multiple Sentences}. In
  \bibinfo{booktitle}{\emph{{NAACL-HLT}}}. \bibinfo{publisher}{Association for
  Computational Linguistics}, \bibinfo{pages}{252--262}.
\newblock


\bibitem[\protect\citeauthoryear{Krause, Damonte, Dobre, Duma, Fainberg,
  Fancellu, Kahembwe, Cheng, and Webber}{Krause et~al\mbox{.}}{2017}]%
        {DBLP:journals/corr/abs-1709-09816}
\bibfield{author}{\bibinfo{person}{Ben Krause}, \bibinfo{person}{Marco
  Damonte}, \bibinfo{person}{Mihai Dobre}, \bibinfo{person}{Daniel Duma},
  \bibinfo{person}{Joachim Fainberg}, \bibinfo{person}{Federico Fancellu},
  \bibinfo{person}{Emmanuel Kahembwe}, \bibinfo{person}{Jianpeng Cheng}, {and}
  \bibinfo{person}{Bonnie~L. Webber}.} \bibinfo{year}{2017}\natexlab{}.
\newblock \showarticletitle{Edina: Building an Open Domain Socialbot with
  Self-dialogues}.
\newblock \bibinfo{journal}{\emph{CoRR}}  \bibinfo{volume}{abs/1709.09816}
  (\bibinfo{year}{2017}).
\newblock
\showeprint[arxiv]{1709.09816}
\urldef\tempurl%
\url{http://arxiv.org/abs/1709.09816}
\showURL{%
\tempurl}


\bibitem[\protect\citeauthoryear{Krishna and Iyyer}{Krishna and Iyyer}{2019}]%
        {DBLP:conf/acl/KrishnaI19}
\bibfield{author}{\bibinfo{person}{Kalpesh Krishna} {and}
  \bibinfo{person}{Mohit Iyyer}.} \bibinfo{year}{2019}\natexlab{}.
\newblock \showarticletitle{Generating Question-Answer Hierarchies}. In
  \bibinfo{booktitle}{\emph{{ACL} {(1)}}}. \bibinfo{publisher}{Association for
  Computational Linguistics}, \bibinfo{pages}{2321--2334}.
\newblock


\bibitem[\protect\citeauthoryear{Kwiatkowski, Palomaki, Redfield, Collins,
  Parikh, Alberti, Epstein, Polosukhin, Devlin, Lee, Toutanova, Jones, Kelcey,
  Chang, Dai, Uszkoreit, Le, and Petrov}{Kwiatkowski et~al\mbox{.}}{2019}]%
        {DBLP:journals/tacl/KwiatkowskiPRCP19}
\bibfield{author}{\bibinfo{person}{Tom Kwiatkowski},
  \bibinfo{person}{Jennimaria Palomaki}, \bibinfo{person}{Olivia Redfield},
  \bibinfo{person}{Michael Collins}, \bibinfo{person}{Ankur~P. Parikh},
  \bibinfo{person}{Chris Alberti}, \bibinfo{person}{Danielle Epstein},
  \bibinfo{person}{Illia Polosukhin}, \bibinfo{person}{Jacob Devlin},
  \bibinfo{person}{Kenton Lee}, \bibinfo{person}{Kristina Toutanova},
  \bibinfo{person}{Llion Jones}, \bibinfo{person}{Matthew Kelcey},
  \bibinfo{person}{Ming{-}Wei Chang}, \bibinfo{person}{Andrew~M. Dai},
  \bibinfo{person}{Jakob Uszkoreit}, \bibinfo{person}{Quoc Le}, {and}
  \bibinfo{person}{Slav Petrov}.} \bibinfo{year}{2019}\natexlab{}.
\newblock \showarticletitle{Natural Questions: a Benchmark for Question
  Answering Research}.
\newblock \bibinfo{journal}{\emph{{TACL}}}  \bibinfo{volume}{7}
  (\bibinfo{year}{2019}), \bibinfo{pages}{452--466}.
\newblock


\bibitem[\protect\citeauthoryear{Lavie and Agarwal}{Lavie and Agarwal}{2007}]%
        {DBLP:conf/wmt/LavieA07}
\bibfield{author}{\bibinfo{person}{Alon Lavie} {and} \bibinfo{person}{Abhaya
  Agarwal}.} \bibinfo{year}{2007}\natexlab{}.
\newblock \showarticletitle{{METEOR:} An Automatic Metric for {MT} Evaluation
  with High Levels of Correlation with Human Judgments}. In
  \bibinfo{booktitle}{\emph{WMT@ACL}}. \bibinfo{publisher}{Association for
  Computational Linguistics}, \bibinfo{pages}{228--231}.
\newblock


\bibitem[\protect\citeauthoryear{Lawrence, Kotnis, and Niepert}{Lawrence
  et~al\mbox{.}}{2019}]%
        {DBLP:conf/emnlp/LawrenceKN19}
\bibfield{author}{\bibinfo{person}{Carolin Lawrence}, \bibinfo{person}{Bhushan
  Kotnis}, {and} \bibinfo{person}{Mathias Niepert}.}
  \bibinfo{year}{2019}\natexlab{}.
\newblock \showarticletitle{Attending to Future Tokens for Bidirectional
  Sequence Generation}. In \bibinfo{booktitle}{\emph{Proceedings of the 2019
  Conference on Empirical Methods in Natural Language Processing and the 9th
  International Joint Conference on Natural Language Processing, {EMNLP-IJCNLP}
  2019, Hong Kong, China, November 3-7, 2019}},
  \bibfield{editor}{\bibinfo{person}{Kentaro Inui}, \bibinfo{person}{Jing
  Jiang}, \bibinfo{person}{Vincent Ng}, {and} \bibinfo{person}{Xiaojun Wan}}
  (Eds.). \bibinfo{publisher}{Association for Computational Linguistics},
  \bibinfo{pages}{1--10}.
\newblock
\urldef\tempurl%
\url{https://doi.org/10.18653/v1/D19-1001}
\showDOI{\tempurl}


\bibitem[\protect\citeauthoryear{Li, Galley, Brockett, Gao, and Dolan}{Li
  et~al\mbox{.}}{2016}]%
        {DBLP:conf/naacl/LiGBGD16}
\bibfield{author}{\bibinfo{person}{Jiwei Li}, \bibinfo{person}{Michel Galley},
  \bibinfo{person}{Chris Brockett}, \bibinfo{person}{Jianfeng Gao}, {and}
  \bibinfo{person}{Bill Dolan}.} \bibinfo{year}{2016}\natexlab{}.
\newblock \showarticletitle{A Diversity-Promoting Objective Function for Neural
  Conversation Models}. In \bibinfo{booktitle}{\emph{{HLT-NAACL}}}.
  \bibinfo{publisher}{The Association for Computational Linguistics},
  \bibinfo{pages}{110--119}.
\newblock


\bibitem[\protect\citeauthoryear{Li, Niu, Meng, Feng, Li, and Zhou}{Li
  et~al\mbox{.}}{2019}]%
        {DBLP:conf/acl/LiNMFLZ19}
\bibfield{author}{\bibinfo{person}{Zekang Li}, \bibinfo{person}{Cheng Niu},
  \bibinfo{person}{Fandong Meng}, \bibinfo{person}{Yang Feng},
  \bibinfo{person}{Qian Li}, {and} \bibinfo{person}{Jie Zhou}.}
  \bibinfo{year}{2019}\natexlab{}.
\newblock \showarticletitle{Incremental Transformer with Deliberation Decoder
  for Document Grounded Conversations}. In \bibinfo{booktitle}{\emph{{ACL}
  {(1)}}}. \bibinfo{publisher}{Association for Computational Linguistics},
  \bibinfo{pages}{12--21}.
\newblock


\bibitem[\protect\citeauthoryear{Lin}{Lin}{2004}]%
        {lin2004rouge}
\bibfield{author}{\bibinfo{person}{Chin-Yew Lin}.}
  \bibinfo{year}{2004}\natexlab{}.
\newblock \showarticletitle{Rouge: A package for automatic evaluation of
  summaries}. In \bibinfo{booktitle}{\emph{Text summarization branches out}}.
  \bibinfo{pages}{74--81}.
\newblock


\bibitem[\protect\citeauthoryear{Liu, Lowe, Serban, Noseworthy, Charlin, and
  Pineau}{Liu et~al\mbox{.}}{2016}]%
        {DBLP:conf/emnlp/LiuLSNCP16}
\bibfield{author}{\bibinfo{person}{Chia{-}Wei Liu}, \bibinfo{person}{Ryan
  Lowe}, \bibinfo{person}{Iulian Serban}, \bibinfo{person}{Michael Noseworthy},
  \bibinfo{person}{Laurent Charlin}, {and} \bibinfo{person}{Joelle Pineau}.}
  \bibinfo{year}{2016}\natexlab{}.
\newblock \showarticletitle{How {NOT} To Evaluate Your Dialogue System: An
  Empirical Study of Unsupervised Evaluation Metrics for Dialogue Response
  Generation}. In \bibinfo{booktitle}{\emph{{EMNLP}}}. \bibinfo{publisher}{The
  Association for Computational Linguistics}, \bibinfo{pages}{2122--2132}.
\newblock


\bibitem[\protect\citeauthoryear{Liu, Chen, Ren, Feng, Liu, and Yin}{Liu
  et~al\mbox{.}}{2018a}]%
        {DBLP:conf/acl/LiuFCRYL18}
\bibfield{author}{\bibinfo{person}{Shuman Liu}, \bibinfo{person}{Hongshen
  Chen}, \bibinfo{person}{Zhaochun Ren}, \bibinfo{person}{Yang Feng},
  \bibinfo{person}{Qun Liu}, {and} \bibinfo{person}{Dawei Yin}.}
  \bibinfo{year}{2018}\natexlab{a}.
\newblock \showarticletitle{Knowledge Diffusion for Neural Dialogue
  Generation}. In \bibinfo{booktitle}{\emph{{ACL} {(1)}}}.
  \bibinfo{publisher}{Association for Computational Linguistics},
  \bibinfo{pages}{1489--1498}.
\newblock


\bibitem[\protect\citeauthoryear{Liu, Zhang, Zhang, Wang, and Zhang}{Liu
  et~al\mbox{.}}{2019c}]%
        {DBLP:journals/corr/abs-1907-01118}
\bibfield{author}{\bibinfo{person}{Shanshan Liu}, \bibinfo{person}{Xin Zhang},
  \bibinfo{person}{Sheng Zhang}, \bibinfo{person}{Hui Wang}, {and}
  \bibinfo{person}{Weiming Zhang}.} \bibinfo{year}{2019}\natexlab{c}.
\newblock \showarticletitle{Neural Machine Reading Comprehension: Methods and
  Trends}.
\newblock \bibinfo{journal}{\emph{CoRR}}  \bibinfo{volume}{abs/1907.01118}
  (\bibinfo{year}{2019}).
\newblock
\showeprint[arxiv]{1907.01118}
\urldef\tempurl%
\url{http://arxiv.org/abs/1907.01118}
\showURL{%
\tempurl}


\bibitem[\protect\citeauthoryear{Liu, Shen, Duh, and Gao}{Liu
  et~al\mbox{.}}{2018b}]%
        {DBLP:conf/acl/GaoDLS18}
\bibfield{author}{\bibinfo{person}{Xiaodong Liu}, \bibinfo{person}{Yelong
  Shen}, \bibinfo{person}{Kevin Duh}, {and} \bibinfo{person}{Jianfeng Gao}.}
  \bibinfo{year}{2018}\natexlab{b}.
\newblock \showarticletitle{Stochastic Answer Networks for Machine Reading
  Comprehension}. In \bibinfo{booktitle}{\emph{{ACL} {(1)}}}.
  \bibinfo{publisher}{Association for Computational Linguistics},
  \bibinfo{pages}{1694--1704}.
\newblock


\bibitem[\protect\citeauthoryear{Liu, Ott, Goyal, Du, Joshi, Chen, Levy, Lewis,
  Zettlemoyer, and Stoyanov}{Liu et~al\mbox{.}}{2019b}]%
        {DBLP:journals/corr/abs-1907-11692}
\bibfield{author}{\bibinfo{person}{Yinhan Liu}, \bibinfo{person}{Myle Ott},
  \bibinfo{person}{Naman Goyal}, \bibinfo{person}{Jingfei Du},
  \bibinfo{person}{Mandar Joshi}, \bibinfo{person}{Danqi Chen},
  \bibinfo{person}{Omer Levy}, \bibinfo{person}{Mike Lewis},
  \bibinfo{person}{Luke Zettlemoyer}, {and} \bibinfo{person}{Veselin
  Stoyanov}.} \bibinfo{year}{2019}\natexlab{b}.
\newblock \showarticletitle{RoBERTa: {A} Robustly Optimized {BERT} Pretraining
  Approach}.
\newblock \bibinfo{journal}{\emph{CoRR}}  \bibinfo{volume}{abs/1907.11692}
  (\bibinfo{year}{2019}).
\newblock
\showeprint[arxiv]{1907.11692}
\urldef\tempurl%
\url{http://arxiv.org/abs/1907.11692}
\showURL{%
\tempurl}


\bibitem[\protect\citeauthoryear{Liu, Niu, Wu, and Wang}{Liu
  et~al\mbox{.}}{2019a}]%
        {DBLP:journals/corr/abs-1903-10245}
\bibfield{author}{\bibinfo{person}{Zhibin Liu}, \bibinfo{person}{Zheng{-}Yu
  Niu}, \bibinfo{person}{Hua Wu}, {and} \bibinfo{person}{Haifeng Wang}.}
  \bibinfo{year}{2019}\natexlab{a}.
\newblock \showarticletitle{Knowledge Aware Conversation Generation with
  Reasoning on Augmented Graph}.
\newblock \bibinfo{journal}{\emph{CoRR}}  \bibinfo{volume}{abs/1903.10245}
  (\bibinfo{year}{2019}).
\newblock
\showeprint[arxiv]{1903.10245}
\urldef\tempurl%
\url{http://arxiv.org/abs/1903.10245}
\showURL{%
\tempurl}


\bibitem[\protect\citeauthoryear{Long, Wang, Xu, Wang, Wang, and Wang}{Long
  et~al\mbox{.}}{2017}]%
        {long2017knowledge}
\bibfield{author}{\bibinfo{person}{Yinong Long}, \bibinfo{person}{Jianan Wang},
  \bibinfo{person}{Zhen Xu}, \bibinfo{person}{Zongsheng Wang},
  \bibinfo{person}{Baoxun Wang}, {and} \bibinfo{person}{Zhuoran Wang}.}
  \bibinfo{year}{2017}\natexlab{}.
\newblock \showarticletitle{A knowledge enhanced generative conversational
  service agent}. In \bibinfo{booktitle}{\emph{DSTC6 Workshop}}.
\newblock


\bibitem[\protect\citeauthoryear{Lowe, Noseworthy, Serban, Angelard{-}Gontier,
  Bengio, and Pineau}{Lowe et~al\mbox{.}}{2017}]%
        {DBLP:conf/acl/LoweNSABP17}
\bibfield{author}{\bibinfo{person}{Ryan Lowe}, \bibinfo{person}{Michael
  Noseworthy}, \bibinfo{person}{Iulian~Vlad Serban}, \bibinfo{person}{Nicolas
  Angelard{-}Gontier}, \bibinfo{person}{Yoshua Bengio}, {and}
  \bibinfo{person}{Joelle Pineau}.} \bibinfo{year}{2017}\natexlab{}.
\newblock \showarticletitle{Towards an Automatic Turing Test: Learning to
  Evaluate Dialogue Responses}. In \bibinfo{booktitle}{\emph{{ACL} {(1)}}}.
  \bibinfo{publisher}{Association for Computational Linguistics},
  \bibinfo{pages}{1116--1126}.
\newblock


\bibitem[\protect\citeauthoryear{Lowe, Pow, Serban, Charlin, and Pineau}{Lowe
  et~al\mbox{.}}{2015b}]%
        {lowe2015incorporating}
\bibfield{author}{\bibinfo{person}{Ryan Lowe}, \bibinfo{person}{Nissan Pow},
  \bibinfo{person}{Iulian Serban}, \bibinfo{person}{Laurent Charlin}, {and}
  \bibinfo{person}{Joelle Pineau}.} \bibinfo{year}{2015}\natexlab{b}.
\newblock \showarticletitle{Incorporating unstructured textual knowledge
  sources into neural dialogue systems}. In \bibinfo{booktitle}{\emph{Neural
  information processing systems workshop on machine learning for spoken
  language understanding}}.
\newblock


\bibitem[\protect\citeauthoryear{Lowe, Pow, Serban, and Pineau}{Lowe
  et~al\mbox{.}}{2015a}]%
        {DBLP:conf/sigdial/LowePSP15}
\bibfield{author}{\bibinfo{person}{Ryan Lowe}, \bibinfo{person}{Nissan Pow},
  \bibinfo{person}{Iulian Serban}, {and} \bibinfo{person}{Joelle Pineau}.}
  \bibinfo{year}{2015}\natexlab{a}.
\newblock \showarticletitle{The Ubuntu Dialogue Corpus: {A} Large Dataset for
  Research in Unstructured Multi-Turn Dialogue Systems}. In
  \bibinfo{booktitle}{\emph{{SIGDIAL} Conference}}. \bibinfo{publisher}{The
  Association for Computer Linguistics}, \bibinfo{pages}{285--294}.
\newblock


\bibitem[\protect\citeauthoryear{Lu, Pramanik, Roy, Abujabal, Wang, and
  Weikum}{Lu et~al\mbox{.}}{2019}]%
        {DBLP:conf/sigir/LuPRAWW19}
\bibfield{author}{\bibinfo{person}{Xiaolu Lu}, \bibinfo{person}{Soumajit
  Pramanik}, \bibinfo{person}{Rishiraj~Saha Roy}, \bibinfo{person}{Abdalghani
  Abujabal}, \bibinfo{person}{Yafang Wang}, {and} \bibinfo{person}{Gerhard
  Weikum}.} \bibinfo{year}{2019}\natexlab{}.
\newblock \showarticletitle{Answering Complex Questions by Joining
  Multi-Document Evidence with Quasi Knowledge Graphs}. In
  \bibinfo{booktitle}{\emph{{SIGIR}}}. \bibinfo{publisher}{{ACM}},
  \bibinfo{pages}{105--114}.
\newblock


\bibitem[\protect\citeauthoryear{Ma, Jurczyk, and Choi}{Ma
  et~al\mbox{.}}{2018}]%
        {DBLP:conf/naacl/MaJC18}
\bibfield{author}{\bibinfo{person}{Kaixin Ma}, \bibinfo{person}{Tomasz
  Jurczyk}, {and} \bibinfo{person}{Jinho~D. Choi}.}
  \bibinfo{year}{2018}\natexlab{}.
\newblock \showarticletitle{Challenging Reading Comprehension on Daily
  Conversation: Passage Completion on Multiparty Dialog}. In
  \bibinfo{booktitle}{\emph{{NAACL-HLT}}}. \bibinfo{publisher}{Association for
  Computational Linguistics}, \bibinfo{pages}{2039--2048}.
\newblock


\bibitem[\protect\citeauthoryear{Meng, Ren, Chen, Monz, Ma, and de~Rijke}{Meng
  et~al\mbox{.}}{2019}]%
        {DBLP:journals/corr/abs-1908-06449}
\bibfield{author}{\bibinfo{person}{Chuan Meng}, \bibinfo{person}{Pengjie Ren},
  \bibinfo{person}{Zhumin Chen}, \bibinfo{person}{Christof Monz},
  \bibinfo{person}{Jun Ma}, {and} \bibinfo{person}{Maarten de Rijke}.}
  \bibinfo{year}{2019}\natexlab{}.
\newblock \showarticletitle{RefNet: {A} Reference-aware Network for Background
  Based Conversation}.
\newblock \bibinfo{journal}{\emph{CoRR}}  \bibinfo{volume}{abs/1908.06449}
  (\bibinfo{year}{2019}).
\newblock
\showeprint[arxiv]{1908.06449}
\urldef\tempurl%
\url{http://arxiv.org/abs/1908.06449}
\showURL{%
\tempurl}


\bibitem[\protect\citeauthoryear{Mihaylov, Clark, Khot, and Sabharwal}{Mihaylov
  et~al\mbox{.}}{2018}]%
        {DBLP:conf/emnlp/MihaylovCKS18}
\bibfield{author}{\bibinfo{person}{Todor Mihaylov}, \bibinfo{person}{Peter
  Clark}, \bibinfo{person}{Tushar Khot}, {and} \bibinfo{person}{Ashish
  Sabharwal}.} \bibinfo{year}{2018}\natexlab{}.
\newblock \showarticletitle{Can a Suit of Armor Conduct Electricity? {A} New
  Dataset for Open Book Question Answering}. In
  \bibinfo{booktitle}{\emph{{EMNLP}}}. \bibinfo{publisher}{Association for
  Computational Linguistics}, \bibinfo{pages}{2381--2391}.
\newblock


\bibitem[\protect\citeauthoryear{Mikolov, Karafi{\'{a}}t, Burget,
  Cernock{\'{y}}, and Khudanpur}{Mikolov et~al\mbox{.}}{2010}]%
        {DBLP:conf/interspeech/MikolovKBCK10}
\bibfield{author}{\bibinfo{person}{Tomas Mikolov}, \bibinfo{person}{Martin
  Karafi{\'{a}}t}, \bibinfo{person}{Luk{\'{a}}s Burget}, \bibinfo{person}{Jan
  Cernock{\'{y}}}, {and} \bibinfo{person}{Sanjeev Khudanpur}.}
  \bibinfo{year}{2010}\natexlab{}.
\newblock \showarticletitle{Recurrent neural network based language model}. In
  \bibinfo{booktitle}{\emph{{INTERSPEECH}}}. \bibinfo{publisher}{{ISCA}},
  \bibinfo{pages}{1045--1048}.
\newblock


\bibitem[\protect\citeauthoryear{Miller, Feng, Batra, Bordes, Fisch, Lu,
  Parikh, and Weston}{Miller et~al\mbox{.}}{2017}]%
        {DBLP:conf/emnlp/MillerFBBFLPW17}
\bibfield{author}{\bibinfo{person}{Alexander~H. Miller}, \bibinfo{person}{Will
  Feng}, \bibinfo{person}{Dhruv Batra}, \bibinfo{person}{Antoine Bordes},
  \bibinfo{person}{Adam Fisch}, \bibinfo{person}{Jiasen Lu},
  \bibinfo{person}{Devi Parikh}, {and} \bibinfo{person}{Jason Weston}.}
  \bibinfo{year}{2017}\natexlab{}.
\newblock \showarticletitle{ParlAI: {A} Dialog Research Software Platform}. In
  \bibinfo{booktitle}{\emph{Proceedings of the 2017 Conference on Empirical
  Methods in Natural Language Processing, {EMNLP} 2017, Copenhagen, Denmark,
  September 9-11, 2017 - System Demonstrations}},
  \bibfield{editor}{\bibinfo{person}{Lucia Specia}, \bibinfo{person}{Matt
  Post}, {and} \bibinfo{person}{Michael Paul}} (Eds.).
  \bibinfo{publisher}{Association for Computational Linguistics},
  \bibinfo{pages}{79--84}.
\newblock
\urldef\tempurl%
\url{https://www.aclweb.org/anthology/D17-2014/}
\showURL{%
\tempurl}


\bibitem[\protect\citeauthoryear{Moghe, Arora, Banerjee, and Khapra}{Moghe
  et~al\mbox{.}}{2018}]%
        {DBLP:conf/emnlp/MogheABK18}
\bibfield{author}{\bibinfo{person}{Nikita Moghe}, \bibinfo{person}{Siddhartha
  Arora}, \bibinfo{person}{Suman Banerjee}, {and} \bibinfo{person}{Mitesh~M.
  Khapra}.} \bibinfo{year}{2018}\natexlab{}.
\newblock \showarticletitle{Towards Exploiting Background Knowledge for
  Building Conversation Systems}. In \bibinfo{booktitle}{\emph{{EMNLP}}}.
  \bibinfo{publisher}{Association for Computational Linguistics},
  \bibinfo{pages}{2322--2332}.
\newblock


\bibitem[\protect\citeauthoryear{Nguyen, Rosenberg, Song, Gao, Tiwary,
  Majumder, and Deng}{Nguyen et~al\mbox{.}}{2016}]%
        {DBLP:conf/nips/NguyenRSGTMD16}
\bibfield{author}{\bibinfo{person}{Tri Nguyen}, \bibinfo{person}{Mir
  Rosenberg}, \bibinfo{person}{Xia Song}, \bibinfo{person}{Jianfeng Gao},
  \bibinfo{person}{Saurabh Tiwary}, \bibinfo{person}{Rangan Majumder}, {and}
  \bibinfo{person}{Li Deng}.} \bibinfo{year}{2016}\natexlab{}.
\newblock \showarticletitle{{MS} {MARCO:} {A} Human Generated MAchine Reading
  COmprehension Dataset}. In \bibinfo{booktitle}{\emph{Proceedings of the
  Workshop on Cognitive Computation: Integrating neural and symbolic approaches
  2016 co-located with the 30th Annual Conference on Neural Information
  Processing Systems {(NIPS} 2016), Barcelona, Spain, December 9, 2016}}
  \emph{(\bibinfo{series}{{CEUR} Workshop Proceedings})},
  \bibfield{editor}{\bibinfo{person}{Tarek~Richard Besold},
  \bibinfo{person}{Antoine Bordes}, \bibinfo{person}{Artur~S. d'Avila Garcez},
  {and} \bibinfo{person}{Greg Wayne}} (Eds.), Vol.~\bibinfo{volume}{1773}.
  \bibinfo{publisher}{CEUR-WS.org}.
\newblock
\urldef\tempurl%
\url{http://ceur-ws.org/Vol-1773/CoCoNIPS\_2016\_paper9.pdf}
\showURL{%
\tempurl}


\bibitem[\protect\citeauthoryear{Ohsugi, Saito, Nishida, Asano, and
  Tomita}{Ohsugi et~al\mbox{.}}{2019}]%
        {DBLP:journals/corr/abs-1905-12848}
\bibfield{author}{\bibinfo{person}{Yasuhito Ohsugi}, \bibinfo{person}{Itsumi
  Saito}, \bibinfo{person}{Kyosuke Nishida}, \bibinfo{person}{Hisako Asano},
  {and} \bibinfo{person}{Junji Tomita}.} \bibinfo{year}{2019}\natexlab{}.
\newblock \showarticletitle{A Simple but Effective Method to Incorporate
  Multi-turn Context with {BERT} for Conversational Machine Comprehension}.
\newblock \bibinfo{journal}{\emph{CoRR}}  \bibinfo{volume}{abs/1905.12848}
  (\bibinfo{year}{2019}).
\newblock
\showeprint[arxiv]{1905.12848}
\urldef\tempurl%
\url{http://arxiv.org/abs/1905.12848}
\showURL{%
\tempurl}


\bibitem[\protect\citeauthoryear{Ostermann, Roth, Modi, Thater, and
  Pinkal}{Ostermann et~al\mbox{.}}{2018}]%
        {DBLP:conf/semeval/OstermannRMTP18}
\bibfield{author}{\bibinfo{person}{Simon Ostermann}, \bibinfo{person}{Michael
  Roth}, \bibinfo{person}{Ashutosh Modi}, \bibinfo{person}{Stefan Thater},
  {and} \bibinfo{person}{Manfred Pinkal}.} \bibinfo{year}{2018}\natexlab{}.
\newblock \showarticletitle{SemEval-2018 Task 11: Machine Comprehension Using
  Commonsense Knowledge}. In \bibinfo{booktitle}{\emph{SemEval@NAACL-HLT}}.
  \bibinfo{publisher}{Association for Computational Linguistics},
  \bibinfo{pages}{747--757}.
\newblock


\bibitem[\protect\citeauthoryear{Paek}{Paek}{2001}]%
        {DBLP:conf/sigdial/Paek01}
\bibfield{author}{\bibinfo{person}{Tim Paek}.} \bibinfo{year}{2001}\natexlab{}.
\newblock \showarticletitle{Empirical Methods for Evaluating Dialog Systems}.
  In \bibinfo{booktitle}{\emph{{SIGDIAL} Workshop}}. \bibinfo{publisher}{The
  Association for Computer Linguistics}.
\newblock


\bibitem[\protect\citeauthoryear{Papineni, Roukos, Ward, and Zhu}{Papineni
  et~al\mbox{.}}{2002}]%
        {DBLP:conf/acl/PapineniRWZ02}
\bibfield{author}{\bibinfo{person}{Kishore Papineni}, \bibinfo{person}{Salim
  Roukos}, \bibinfo{person}{Todd Ward}, {and} \bibinfo{person}{Wei{-}Jing
  Zhu}.} \bibinfo{year}{2002}\natexlab{}.
\newblock \showarticletitle{Bleu: a Method for Automatic Evaluation of Machine
  Translation}. In \bibinfo{booktitle}{\emph{{ACL}}}.
  \bibinfo{publisher}{{ACL}}, \bibinfo{pages}{311--318}.
\newblock


\bibitem[\protect\citeauthoryear{Peters, Neumann, Iyyer, Gardner, Clark, Lee,
  and Zettlemoyer}{Peters et~al\mbox{.}}{2018}]%
        {DBLP:conf/naacl/PetersNIGCLZ18}
\bibfield{author}{\bibinfo{person}{Matthew~E. Peters}, \bibinfo{person}{Mark
  Neumann}, \bibinfo{person}{Mohit Iyyer}, \bibinfo{person}{Matt Gardner},
  \bibinfo{person}{Christopher Clark}, \bibinfo{person}{Kenton Lee}, {and}
  \bibinfo{person}{Luke Zettlemoyer}.} \bibinfo{year}{2018}\natexlab{}.
\newblock \showarticletitle{Deep Contextualized Word Representations}. In
  \bibinfo{booktitle}{\emph{{NAACL-HLT}}}. \bibinfo{publisher}{Association for
  Computational Linguistics}, \bibinfo{pages}{2227--2237}.
\newblock


\bibitem[\protect\citeauthoryear{Qin, Galley, Brockett, Liu, Gao, Dolan, Choi,
  and Gao}{Qin et~al\mbox{.}}{2019}]%
        {DBLP:conf/acl/QinGBLGDCG19}
\bibfield{author}{\bibinfo{person}{Lianhui Qin}, \bibinfo{person}{Michel
  Galley}, \bibinfo{person}{Chris Brockett}, \bibinfo{person}{Xiaodong Liu},
  \bibinfo{person}{Xiang Gao}, \bibinfo{person}{Bill Dolan},
  \bibinfo{person}{Yejin Choi}, {and} \bibinfo{person}{Jianfeng Gao}.}
  \bibinfo{year}{2019}\natexlab{}.
\newblock \showarticletitle{Conversing by Reading: Contentful Neural
  Conversation with On-demand Machine Reading}. In
  \bibinfo{booktitle}{\emph{{ACL} {(1)}}}. \bibinfo{publisher}{Association for
  Computational Linguistics}, \bibinfo{pages}{5427--5436}.
\newblock


\bibitem[\protect\citeauthoryear{Qiu, Chen, Xu, and Sun}{Qiu
  et~al\mbox{.}}{2019}]%
        {DBLP:journals/corr/abs-1906-03824}
\bibfield{author}{\bibinfo{person}{Boyu Qiu}, \bibinfo{person}{Xu Chen},
  \bibinfo{person}{Jungang Xu}, {and} \bibinfo{person}{Yingfei Sun}.}
  \bibinfo{year}{2019}\natexlab{}.
\newblock \showarticletitle{A Survey on Neural Machine Reading Comprehension}.
\newblock \bibinfo{journal}{\emph{CoRR}}  \bibinfo{volume}{abs/1906.03824}
  (\bibinfo{year}{2019}).
\newblock
\showeprint[arxiv]{1906.03824}
\urldef\tempurl%
\url{http://arxiv.org/abs/1906.03824}
\showURL{%
\tempurl}


\bibitem[\protect\citeauthoryear{Qu, Yang, Qiu, Croft, Zhang, and Iyyer}{Qu
  et~al\mbox{.}}{2019a}]%
        {DBLP:conf/sigir/Qu0QCZI19}
\bibfield{author}{\bibinfo{person}{Chen Qu}, \bibinfo{person}{Liu Yang},
  \bibinfo{person}{Minghui Qiu}, \bibinfo{person}{W.~Bruce Croft},
  \bibinfo{person}{Yongfeng Zhang}, {and} \bibinfo{person}{Mohit Iyyer}.}
  \bibinfo{year}{2019}\natexlab{a}.
\newblock \showarticletitle{{BERT} with History Answer Embedding for
  Conversational Question Answering}. In \bibinfo{booktitle}{\emph{{SIGIR}}}.
  \bibinfo{publisher}{{ACM}}, \bibinfo{pages}{1133--1136}.
\newblock


\bibitem[\protect\citeauthoryear{Qu, Yang, Qiu, Zhang, Chen, Croft, and
  Iyyer}{Qu et~al\mbox{.}}{2019b}]%
        {DBLP:conf/cikm/QuYQZCCI19}
\bibfield{author}{\bibinfo{person}{Chen Qu}, \bibinfo{person}{Liu Yang},
  \bibinfo{person}{Minghui Qiu}, \bibinfo{person}{Yongfeng Zhang},
  \bibinfo{person}{Cen Chen}, \bibinfo{person}{W.~Bruce Croft}, {and}
  \bibinfo{person}{Mohit Iyyer}.} \bibinfo{year}{2019}\natexlab{b}.
\newblock \showarticletitle{Attentive History Selection for Conversational
  Question Answering}. In \bibinfo{booktitle}{\emph{Proceedings of the 28th
  {ACM} International Conference on Information and Knowledge Management,
  {CIKM} 2019, Beijing, China, November 3-7, 2019}},
  \bibfield{editor}{\bibinfo{person}{Wenwu Zhu}, \bibinfo{person}{Dacheng Tao},
  \bibinfo{person}{Xueqi Cheng}, \bibinfo{person}{Peng Cui},
  \bibinfo{person}{Elke~A. Rundensteiner}, \bibinfo{person}{David Carmel},
  \bibinfo{person}{Qi~He}, {and} \bibinfo{person}{Jeffrey~Xu Yu}} (Eds.).
  \bibinfo{publisher}{{ACM}}, \bibinfo{pages}{1391--1400}.
\newblock
\urldef\tempurl%
\url{https://doi.org/10.1145/3357384.3357905}
\showDOI{\tempurl}


\bibitem[\protect\citeauthoryear{Radford, Narasimhan, Salimans, and
  Sutskever}{Radford et~al\mbox{.}}{2018}]%
        {radford2018improving}
\bibfield{author}{\bibinfo{person}{Alec Radford}, \bibinfo{person}{Karthik
  Narasimhan}, \bibinfo{person}{Tim Salimans}, {and} \bibinfo{person}{Ilya
  Sutskever}.} \bibinfo{year}{2018}\natexlab{}.
\newblock \showarticletitle{Improving language understanding by generative
  pre-training}.
\newblock \bibinfo{journal}{\emph{URL https://s3-us-west-2. amazonaws.
  com/openai-assets/researchcovers/languageunsupervised/language understanding
  paper. pdf}} (\bibinfo{year}{2018}).
\newblock


\bibitem[\protect\citeauthoryear{Radlinski, Balog, Byrne, and
  Krishnamoorthi}{Radlinski et~al\mbox{.}}{2019}]%
        {radlinski2019coached}
\bibfield{author}{\bibinfo{person}{Filip Radlinski}, \bibinfo{person}{Krisztian
  Balog}, \bibinfo{person}{Bill Byrne}, {and} \bibinfo{person}{Karthik
  Krishnamoorthi}.} \bibinfo{year}{2019}\natexlab{}.
\newblock \showarticletitle{Coached Conversational Preference Elicitation: A
  Case Study in Understanding Movie Preferences}.
\newblock  (\bibinfo{year}{2019}).
\newblock


\bibitem[\protect\citeauthoryear{Rajpurkar, Zhang, Lopyrev, and
  Liang}{Rajpurkar et~al\mbox{.}}{2016}]%
        {DBLP:conf/emnlp/RajpurkarZLL16}
\bibfield{author}{\bibinfo{person}{Pranav Rajpurkar}, \bibinfo{person}{Jian
  Zhang}, \bibinfo{person}{Konstantin Lopyrev}, {and} \bibinfo{person}{Percy
  Liang}.} \bibinfo{year}{2016}\natexlab{}.
\newblock \showarticletitle{SQuAD: 100, 000+ Questions for Machine
  Comprehension of Text}. In \bibinfo{booktitle}{\emph{{EMNLP}}}.
  \bibinfo{publisher}{The Association for Computational Linguistics},
  \bibinfo{pages}{2383--2392}.
\newblock


\bibitem[\protect\citeauthoryear{Ram, Prasad, Khatri, Venkatesh, Gabriel, Liu,
  Nunn, Hedayatnia, Cheng, Nagar, King, Bland, Wartick, Pan, Song, Jayadevan,
  Hwang, and Pettigrue}{Ram et~al\mbox{.}}{2018}]%
        {DBLP:journals/corr/abs-1801-03604}
\bibfield{author}{\bibinfo{person}{Ashwin Ram}, \bibinfo{person}{Rohit Prasad},
  \bibinfo{person}{Chandra Khatri}, \bibinfo{person}{Anu Venkatesh},
  \bibinfo{person}{Raefer Gabriel}, \bibinfo{person}{Qing Liu},
  \bibinfo{person}{Jeff Nunn}, \bibinfo{person}{Behnam Hedayatnia},
  \bibinfo{person}{Ming Cheng}, \bibinfo{person}{Ashish Nagar},
  \bibinfo{person}{Eric King}, \bibinfo{person}{Kate Bland},
  \bibinfo{person}{Amanda Wartick}, \bibinfo{person}{Yi Pan},
  \bibinfo{person}{Han Song}, \bibinfo{person}{Sk Jayadevan},
  \bibinfo{person}{Gene Hwang}, {and} \bibinfo{person}{Art Pettigrue}.}
  \bibinfo{year}{2018}\natexlab{}.
\newblock \showarticletitle{Conversational {AI:} The Science Behind the Alexa
  Prize}.
\newblock \bibinfo{journal}{\emph{CoRR}}  \bibinfo{volume}{abs/1801.03604}
  (\bibinfo{year}{2018}).
\newblock
\showeprint[arxiv]{1801.03604}
\urldef\tempurl%
\url{http://arxiv.org/abs/1801.03604}
\showURL{%
\tempurl}


\bibitem[\protect\citeauthoryear{Reddy, Chen, and Manning}{Reddy
  et~al\mbox{.}}{2019}]%
        {DBLP:journals/tacl/ReddyCM19}
\bibfield{author}{\bibinfo{person}{Siva Reddy}, \bibinfo{person}{Danqi Chen},
  {and} \bibinfo{person}{Christopher~D. Manning}.}
  \bibinfo{year}{2019}\natexlab{}.
\newblock \showarticletitle{CoQA: {A} Conversational Question Answering
  Challenge}.
\newblock \bibinfo{journal}{\emph{{TACL}}}  \bibinfo{volume}{7}
  (\bibinfo{year}{2019}), \bibinfo{pages}{249--266}.
\newblock


\bibitem[\protect\citeauthoryear{Ren, Chen, Monz, Ma, and de~Rijke}{Ren
  et~al\mbox{.}}{2019}]%
        {DBLP:journals/corr/abs-1908-09528}
\bibfield{author}{\bibinfo{person}{Pengjie Ren}, \bibinfo{person}{Zhumin Chen},
  \bibinfo{person}{Christof Monz}, \bibinfo{person}{Jun Ma}, {and}
  \bibinfo{person}{Maarten de Rijke}.} \bibinfo{year}{2019}\natexlab{}.
\newblock \showarticletitle{Thinking Globally, Acting Locally: Distantly
  Supervised Global-to-Local Knowledge Selection for Background Based
  Conversation}.
\newblock \bibinfo{journal}{\emph{CoRR}}  \bibinfo{volume}{abs/1908.09528}
  (\bibinfo{year}{2019}).
\newblock
\showeprint[arxiv]{1908.09528}
\urldef\tempurl%
\url{http://arxiv.org/abs/1908.09528}
\showURL{%
\tempurl}


\bibitem[\protect\citeauthoryear{Saeidi, Bartolo, Lewis, Singh,
  Rockt{\"{a}}schel, Sheldon, Bouchard, and Riedel}{Saeidi
  et~al\mbox{.}}{2018}]%
        {DBLP:conf/emnlp/SaeidiBL0RSB018}
\bibfield{author}{\bibinfo{person}{Marzieh Saeidi}, \bibinfo{person}{Max
  Bartolo}, \bibinfo{person}{Patrick Lewis}, \bibinfo{person}{Sameer Singh},
  \bibinfo{person}{Tim Rockt{\"{a}}schel}, \bibinfo{person}{Mike Sheldon},
  \bibinfo{person}{Guillaume Bouchard}, {and} \bibinfo{person}{Sebastian
  Riedel}.} \bibinfo{year}{2018}\natexlab{}.
\newblock \showarticletitle{Interpretation of Natural Language Rules in
  Conversational Machine Reading}. In \bibinfo{booktitle}{\emph{{EMNLP}}}.
  \bibinfo{publisher}{Association for Computational Linguistics},
  \bibinfo{pages}{2087--2097}.
\newblock


\bibitem[\protect\citeauthoryear{Saha, Pahuja, Khapra, Sankaranarayanan, and
  Chandar}{Saha et~al\mbox{.}}{2018}]%
        {DBLP:conf/aaai/SahaPKSC18}
\bibfield{author}{\bibinfo{person}{Amrita Saha}, \bibinfo{person}{Vardaan
  Pahuja}, \bibinfo{person}{Mitesh~M. Khapra}, \bibinfo{person}{Karthik
  Sankaranarayanan}, {and} \bibinfo{person}{Sarath Chandar}.}
  \bibinfo{year}{2018}\natexlab{}.
\newblock \showarticletitle{Complex Sequential Question Answering: Towards
  Learning to Converse Over Linked Question Answer Pairs with a Knowledge
  Graph}. In \bibinfo{booktitle}{\emph{{AAAI}}}. \bibinfo{publisher}{{AAAI}
  Press}, \bibinfo{pages}{705--713}.
\newblock


\bibitem[\protect\citeauthoryear{Sai, Gupta, Khapra, and Srinivasan}{Sai
  et~al\mbox{.}}{2019}]%
        {DBLP:conf/aaai/SaiGKS19}
\bibfield{author}{\bibinfo{person}{Ananya~B. Sai}, \bibinfo{person}{Mithun~Das
  Gupta}, \bibinfo{person}{Mitesh~M. Khapra}, {and} \bibinfo{person}{Mukundhan
  Srinivasan}.} \bibinfo{year}{2019}\natexlab{}.
\newblock \showarticletitle{Re-Evaluating {ADEM:} {A} Deeper Look at Scoring
  Dialogue Responses}. In \bibinfo{booktitle}{\emph{{AAAI}}}.
  \bibinfo{publisher}{{AAAI} Press}, \bibinfo{pages}{6220--6227}.
\newblock


\bibitem[\protect\citeauthoryear{Santhanam and Shaikh}{Santhanam and
  Shaikh}{2019}]%
        {DBLP:journals/corr/abs-1906-00500}
\bibfield{author}{\bibinfo{person}{Sashank Santhanam} {and}
  \bibinfo{person}{Samira Shaikh}.} \bibinfo{year}{2019}\natexlab{}.
\newblock \showarticletitle{A Survey of Natural Language Generation Techniques
  with a Focus on Dialogue Systems - Past, Present and Future Directions}.
\newblock \bibinfo{journal}{\emph{CoRR}}  \bibinfo{volume}{abs/1906.00500}
  (\bibinfo{year}{2019}).
\newblock
\showeprint[arxiv]{1906.00500}
\urldef\tempurl%
\url{http://arxiv.org/abs/1906.00500}
\showURL{%
\tempurl}


\bibitem[\protect\citeauthoryear{Sarikaya}{Sarikaya}{2017}]%
        {DBLP:journals/spm/Sarikaya17}
\bibfield{author}{\bibinfo{person}{Ruhi Sarikaya}.}
  \bibinfo{year}{2017}\natexlab{}.
\newblock \showarticletitle{The Technology Behind Personal Digital Assistants:
  An overview of the system architecture and key components}.
\newblock \bibinfo{journal}{\emph{{IEEE} Signal Process. Mag.}}
  \bibinfo{volume}{34}, \bibinfo{number}{1} (\bibinfo{year}{2017}),
  \bibinfo{pages}{67--81}.
\newblock
\urldef\tempurl%
\url{https://doi.org/10.1109/MSP.2016.2617341}
\showDOI{\tempurl}


\bibitem[\protect\citeauthoryear{See, Liu, and Manning}{See
  et~al\mbox{.}}{2017}]%
        {DBLP:conf/acl/SeeLM17}
\bibfield{author}{\bibinfo{person}{Abigail See}, \bibinfo{person}{Peter~J.
  Liu}, {and} \bibinfo{person}{Christopher~D. Manning}.}
  \bibinfo{year}{2017}\natexlab{}.
\newblock \showarticletitle{Get To The Point: Summarization with
  Pointer-Generator Networks}. In \bibinfo{booktitle}{\emph{{ACL} {(1)}}}.
  \bibinfo{publisher}{Association for Computational Linguistics},
  \bibinfo{pages}{1073--1083}.
\newblock


\bibitem[\protect\citeauthoryear{Seo, Kembhavi, Farhadi, and Hajishirzi}{Seo
  et~al\mbox{.}}{2016}]%
        {DBLP:journals/corr/SeoKFH16}
\bibfield{author}{\bibinfo{person}{Min~Joon Seo}, \bibinfo{person}{Aniruddha
  Kembhavi}, \bibinfo{person}{Ali Farhadi}, {and} \bibinfo{person}{Hannaneh
  Hajishirzi}.} \bibinfo{year}{2016}\natexlab{}.
\newblock \showarticletitle{Bidirectional Attention Flow for Machine
  Comprehension}.
\newblock \bibinfo{journal}{\emph{CoRR}}  \bibinfo{volume}{abs/1611.01603}
  (\bibinfo{year}{2016}).
\newblock
\showeprint[arxiv]{1611.01603}
\urldef\tempurl%
\url{http://arxiv.org/abs/1611.01603}
\showURL{%
\tempurl}


\bibitem[\protect\citeauthoryear{Serban, Lowe, Henderson, Charlin, and
  Pineau}{Serban et~al\mbox{.}}{2018}]%
        {DBLP:journals/dad/SerbanLHCP18}
\bibfield{author}{\bibinfo{person}{Iulian~Vlad Serban}, \bibinfo{person}{Ryan
  Lowe}, \bibinfo{person}{Peter Henderson}, \bibinfo{person}{Laurent Charlin},
  {and} \bibinfo{person}{Joelle Pineau}.} \bibinfo{year}{2018}\natexlab{}.
\newblock \showarticletitle{A Survey of Available Corpora For Building
  Data-Driven Dialogue Systems: The Journal Version}.
\newblock \bibinfo{journal}{\emph{D{\&}D}} \bibinfo{volume}{9},
  \bibinfo{number}{1} (\bibinfo{year}{2018}), \bibinfo{pages}{1--49}.
\newblock


\bibitem[\protect\citeauthoryear{Sharma, Contractor, Kumar, and Joshi}{Sharma
  et~al\mbox{.}}{2019}]%
        {DBLP:journals/corr/abs-1909-03759}
\bibfield{author}{\bibinfo{person}{Abhishek Sharma}, \bibinfo{person}{Danish
  Contractor}, \bibinfo{person}{Harshit Kumar}, {and}
  \bibinfo{person}{Sachindra Joshi}.} \bibinfo{year}{2019}\natexlab{}.
\newblock \showarticletitle{Neural Conversational {QA:} Learning to Reason v.s.
  Exploiting Patterns}.
\newblock \bibinfo{journal}{\emph{CoRR}}  \bibinfo{volume}{abs/1909.03759}
  (\bibinfo{year}{2019}).
\newblock
\showeprint[arxiv]{1909.03759}
\urldef\tempurl%
\url{http://arxiv.org/abs/1909.03759}
\showURL{%
\tempurl}


\bibitem[\protect\citeauthoryear{Shen, Huang, Gao, and Chen}{Shen
  et~al\mbox{.}}{2017}]%
        {DBLP:conf/kdd/ShenHGC17}
\bibfield{author}{\bibinfo{person}{Yelong Shen}, \bibinfo{person}{Po{-}Sen
  Huang}, \bibinfo{person}{Jianfeng Gao}, {and} \bibinfo{person}{Weizhu Chen}.}
  \bibinfo{year}{2017}\natexlab{}.
\newblock \showarticletitle{ReasoNet: Learning to Stop Reading in Machine
  Comprehension}. In \bibinfo{booktitle}{\emph{{KDD}}}.
  \bibinfo{publisher}{{ACM}}, \bibinfo{pages}{1047--1055}.
\newblock


\bibitem[\protect\citeauthoryear{Shum, He, and Li}{Shum et~al\mbox{.}}{2018}]%
        {DBLP:journals/jzusc/ShumHL18}
\bibfield{author}{\bibinfo{person}{Heung{-}Yeung Shum},
  \bibinfo{person}{Xiaodong He}, {and} \bibinfo{person}{Di Li}.}
  \bibinfo{year}{2018}\natexlab{}.
\newblock \showarticletitle{From Eliza to XiaoIce: challenges and opportunities
  with social chatbots}.
\newblock \bibinfo{journal}{\emph{Frontiers of {IT} {\&} {EE}}}
  \bibinfo{volume}{19}, \bibinfo{number}{1} (\bibinfo{year}{2018}),
  \bibinfo{pages}{10--26}.
\newblock


\bibitem[\protect\citeauthoryear{Song, Wang, Yu, Zhang, Florian, and
  Gildea}{Song et~al\mbox{.}}{2018}]%
        {DBLP:journals/corr/abs-1809-02040}
\bibfield{author}{\bibinfo{person}{Linfeng Song}, \bibinfo{person}{Zhiguo
  Wang}, \bibinfo{person}{Mo Yu}, \bibinfo{person}{Yue Zhang},
  \bibinfo{person}{Radu Florian}, {and} \bibinfo{person}{Daniel Gildea}.}
  \bibinfo{year}{2018}\natexlab{}.
\newblock \showarticletitle{Exploring Graph-structured Passage Representation
  for Multi-hop Reading Comprehension with Graph Neural Networks}.
\newblock \bibinfo{journal}{\emph{CoRR}}  \bibinfo{volume}{abs/1809.02040}
  (\bibinfo{year}{2018}).
\newblock
\showeprint[arxiv]{1809.02040}
\urldef\tempurl%
\url{http://arxiv.org/abs/1809.02040}
\showURL{%
\tempurl}


\bibitem[\protect\citeauthoryear{Su, Guo, Fan, Lan, Zhang, and Cheng}{Su
  et~al\mbox{.}}{2019}]%
        {DBLP:conf/aaai/SuGFLZC19}
\bibfield{author}{\bibinfo{person}{Lixin Su}, \bibinfo{person}{Jiafeng Guo},
  \bibinfo{person}{Yixing Fan}, \bibinfo{person}{Yanyan Lan},
  \bibinfo{person}{Ruqing Zhang}, {and} \bibinfo{person}{Xueqi Cheng}.}
  \bibinfo{year}{2019}\natexlab{}.
\newblock \showarticletitle{An Adaptive Framework for Conversational Question
  Answering}. In \bibinfo{booktitle}{\emph{{AAAI}}}. \bibinfo{publisher}{{AAAI}
  Press}, \bibinfo{pages}{10041--10042}.
\newblock


\bibitem[\protect\citeauthoryear{Sun, Yu, Chen, Yu, Choi, and Cardie}{Sun
  et~al\mbox{.}}{2019}]%
        {DBLP:journals/tacl/SunYCYCC19}
\bibfield{author}{\bibinfo{person}{Kai Sun}, \bibinfo{person}{Dian Yu},
  \bibinfo{person}{Jianshu Chen}, \bibinfo{person}{Dong Yu},
  \bibinfo{person}{Yejin Choi}, {and} \bibinfo{person}{Claire Cardie}.}
  \bibinfo{year}{2019}\natexlab{}.
\newblock \showarticletitle{{DREAM:} {A} Challenge Dataset and Models for
  Dialogue-Based Reading Comprehension}.
\newblock \bibinfo{journal}{\emph{{TACL}}}  \bibinfo{volume}{7}
  (\bibinfo{year}{2019}), \bibinfo{pages}{217--231}.
\newblock


\bibitem[\protect\citeauthoryear{Sutskever, Vinyals, and Le}{Sutskever
  et~al\mbox{.}}{2014}]%
        {DBLP:conf/nips/SutskeverVL14}
\bibfield{author}{\bibinfo{person}{Ilya Sutskever}, \bibinfo{person}{Oriol
  Vinyals}, {and} \bibinfo{person}{Quoc~V. Le}.}
  \bibinfo{year}{2014}\natexlab{}.
\newblock \showarticletitle{Sequence to Sequence Learning with Neural
  Networks}. In \bibinfo{booktitle}{\emph{{NIPS}}}.
  \bibinfo{pages}{3104--3112}.
\newblock


\bibitem[\protect\citeauthoryear{Talmor and Berant}{Talmor and Berant}{2018}]%
        {DBLP:conf/naacl/TalmorB18}
\bibfield{author}{\bibinfo{person}{Alon Talmor} {and} \bibinfo{person}{Jonathan
  Berant}.} \bibinfo{year}{2018}\natexlab{}.
\newblock \showarticletitle{The Web as a Knowledge-Base for Answering Complex
  Questions}. In \bibinfo{booktitle}{\emph{{NAACL-HLT}}}.
  \bibinfo{publisher}{Association for Computational Linguistics},
  \bibinfo{pages}{641--651}.
\newblock


\bibitem[\protect\citeauthoryear{Talmor, Geva, and Berant}{Talmor
  et~al\mbox{.}}{2017}]%
        {DBLP:conf/starsem/TalmorGB17}
\bibfield{author}{\bibinfo{person}{Alon Talmor}, \bibinfo{person}{Mor Geva},
  {and} \bibinfo{person}{Jonathan Berant}.} \bibinfo{year}{2017}\natexlab{}.
\newblock \showarticletitle{Evaluating Semantic Parsing against a Simple
  Web-based Question Answering Model}. In \bibinfo{booktitle}{\emph{*SEM}}.
  \bibinfo{publisher}{Association for Computational Linguistics},
  \bibinfo{pages}{161--167}.
\newblock


\bibitem[\protect\citeauthoryear{Tang and Hu}{Tang and Hu}{2019}]%
        {DBLP:conf/ksem/TangH19}
\bibfield{author}{\bibinfo{person}{Xiangru Tang} {and} \bibinfo{person}{Po
  Hu}.} \bibinfo{year}{2019}\natexlab{}.
\newblock \showarticletitle{Knowledge-Aware Self-Attention Networks for
  Document Grounded Dialogue Generation}. In \bibinfo{booktitle}{\emph{{KSEM}
  {(2)}}} \emph{(\bibinfo{series}{Lecture Notes in Computer Science})},
  Vol.~\bibinfo{volume}{11776}. \bibinfo{publisher}{Springer},
  \bibinfo{pages}{400--411}.
\newblock


\bibitem[\protect\citeauthoryear{Tao, Mou, Zhao, and Yan}{Tao
  et~al\mbox{.}}{2018}]%
        {DBLP:conf/aaai/TaoMZY18}
\bibfield{author}{\bibinfo{person}{Chongyang Tao}, \bibinfo{person}{Lili Mou},
  \bibinfo{person}{Dongyan Zhao}, {and} \bibinfo{person}{Rui Yan}.}
  \bibinfo{year}{2018}\natexlab{}.
\newblock \showarticletitle{{RUBER:} An Unsupervised Method for Automatic
  Evaluation of Open-Domain Dialog Systems}. In
  \bibinfo{booktitle}{\emph{{AAAI}}}. \bibinfo{publisher}{{AAAI} Press},
  \bibinfo{pages}{722--729}.
\newblock


\bibitem[\protect\citeauthoryear{Trischler, Wang, Yuan, Harris, Sordoni,
  Bachman, and Suleman}{Trischler et~al\mbox{.}}{2017}]%
        {DBLP:conf/rep4nlp/TrischlerWYHSBS17}
\bibfield{author}{\bibinfo{person}{Adam Trischler}, \bibinfo{person}{Tong
  Wang}, \bibinfo{person}{Xingdi Yuan}, \bibinfo{person}{Justin Harris},
  \bibinfo{person}{Alessandro Sordoni}, \bibinfo{person}{Philip Bachman}, {and}
  \bibinfo{person}{Kaheer Suleman}.} \bibinfo{year}{2017}\natexlab{}.
\newblock \showarticletitle{NewsQA: {A} Machine Comprehension Dataset}. In
  \bibinfo{booktitle}{\emph{Rep4NLP@ACL}}. \bibinfo{publisher}{Association for
  Computational Linguistics}, \bibinfo{pages}{191--200}.
\newblock


\bibitem[\protect\citeauthoryear{Tu, Wang, Huang, Tang, He, and Zhou}{Tu
  et~al\mbox{.}}{2019}]%
        {DBLP:conf/acl/TuWHTHZ19}
\bibfield{author}{\bibinfo{person}{Ming Tu}, \bibinfo{person}{Guangtao Wang},
  \bibinfo{person}{Jing Huang}, \bibinfo{person}{Yun Tang},
  \bibinfo{person}{Xiaodong He}, {and} \bibinfo{person}{Bowen Zhou}.}
  \bibinfo{year}{2019}\natexlab{}.
\newblock \showarticletitle{Multi-hop Reading Comprehension across Multiple
  Documents by Reasoning over Heterogeneous Graphs}. In
  \bibinfo{booktitle}{\emph{{ACL} {(1)}}}. \bibinfo{publisher}{Association for
  Computational Linguistics}, \bibinfo{pages}{2704--2713}.
\newblock


\bibitem[\protect\citeauthoryear{Turing}{Turing}{2009}]%
        {turing2009computing}
\bibfield{author}{\bibinfo{person}{Alan~M Turing}.}
  \bibinfo{year}{2009}\natexlab{}.
\newblock \showarticletitle{Computing machinery and intelligence}.
\newblock In \bibinfo{booktitle}{\emph{Parsing the Turing Test}}.
  \bibinfo{publisher}{Springer}, \bibinfo{pages}{23--65}.
\newblock


\bibitem[\protect\citeauthoryear{Vaswani, Shazeer, Parmar, Uszkoreit, Jones,
  Gomez, Kaiser, and Polosukhin}{Vaswani et~al\mbox{.}}{2017}]%
        {DBLP:conf/nips/VaswaniSPUJGKP17}
\bibfield{author}{\bibinfo{person}{Ashish Vaswani}, \bibinfo{person}{Noam
  Shazeer}, \bibinfo{person}{Niki Parmar}, \bibinfo{person}{Jakob Uszkoreit},
  \bibinfo{person}{Llion Jones}, \bibinfo{person}{Aidan~N. Gomez},
  \bibinfo{person}{Lukasz Kaiser}, {and} \bibinfo{person}{Illia Polosukhin}.}
  \bibinfo{year}{2017}\natexlab{}.
\newblock \showarticletitle{Attention is All you Need}. In
  \bibinfo{booktitle}{\emph{{NIPS}}}. \bibinfo{pages}{5998--6008}.
\newblock


\bibitem[\protect\citeauthoryear{Vougiouklis, Hare, and Simperl}{Vougiouklis
  et~al\mbox{.}}{2016}]%
        {DBLP:conf/coling/VougiouklisHS16}
\bibfield{author}{\bibinfo{person}{Pavlos Vougiouklis},
  \bibinfo{person}{Jonathon~S. Hare}, {and} \bibinfo{person}{Elena Simperl}.}
  \bibinfo{year}{2016}\natexlab{}.
\newblock \showarticletitle{A Neural Network Approach for Knowledge-Driven
  Response Generation}. In \bibinfo{booktitle}{\emph{{COLING}}}.
  \bibinfo{publisher}{{ACL}}, \bibinfo{pages}{3370--3380}.
\newblock


\bibitem[\protect\citeauthoryear{Walker, Aberdeen, Boland, Bratt, Garofolo,
  Hirschman, Le, Lee, Narayanan, Papineni, Pellom, Polifroni, Potamianos,
  Prabhu, Rudnicky, Sanders, Seneff, Stallard, and Whittaker}{Walker
  et~al\mbox{.}}{2001a}]%
        {DBLP:conf/interspeech/WalkerABBGHLLNPPPPPRSSSW01}
\bibfield{author}{\bibinfo{person}{Marilyn~A. Walker}, \bibinfo{person}{John~S.
  Aberdeen}, \bibinfo{person}{Julie~E. Boland}, \bibinfo{person}{Elizabeth~Owen
  Bratt}, \bibinfo{person}{John~S. Garofolo}, \bibinfo{person}{Lynette
  Hirschman}, \bibinfo{person}{Audrey~N. Le}, \bibinfo{person}{Sungbok Lee},
  \bibinfo{person}{Shrikanth Narayanan}, \bibinfo{person}{Kishore Papineni},
  \bibinfo{person}{Bryan~L. Pellom}, \bibinfo{person}{Joseph Polifroni},
  \bibinfo{person}{Alexandros Potamianos}, \bibinfo{person}{P. Prabhu},
  \bibinfo{person}{Alexander~I. Rudnicky}, \bibinfo{person}{Gregory~A.
  Sanders}, \bibinfo{person}{Stephanie Seneff}, \bibinfo{person}{David
  Stallard}, {and} \bibinfo{person}{Steve Whittaker}.}
  \bibinfo{year}{2001}\natexlab{a}.
\newblock \showarticletitle{{DARPA} communicator dialog travel planning
  systems: the june 2000 data collection}. In
  \bibinfo{booktitle}{\emph{{INTERSPEECH}}}. \bibinfo{publisher}{{ISCA}},
  \bibinfo{pages}{1371--1374}.
\newblock


\bibitem[\protect\citeauthoryear{Walker, Passonneau, and Boland}{Walker
  et~al\mbox{.}}{2001b}]%
        {DBLP:conf/acl/WalkerPB01}
\bibfield{author}{\bibinfo{person}{Marilyn~A. Walker},
  \bibinfo{person}{Rebecca~J. Passonneau}, {and} \bibinfo{person}{Julie~E.
  Boland}.} \bibinfo{year}{2001}\natexlab{b}.
\newblock \showarticletitle{Quantitative and Qualitative Evaluation of Darpa
  Communicator Spoken Dialogue Systems}. In \bibinfo{booktitle}{\emph{{ACL}}}.
  \bibinfo{publisher}{Morgan Kaufmann Publishers}, \bibinfo{pages}{515--522}.
\newblock


\bibitem[\protect\citeauthoryear{Walker, Rudnicky, Aberdeen, Bratt, Garofolo,
  Hastie, Le, Pellom, Potamianos, Passonneau, Prasad, Roukos, Sanders, Seneff,
  and Stallard}{Walker et~al\mbox{.}}{2002}]%
        {DBLP:conf/interspeech/WalkerRABGHLPPPPRSSS02}
\bibfield{author}{\bibinfo{person}{Marilyn~A. Walker},
  \bibinfo{person}{Alexander~I. Rudnicky}, \bibinfo{person}{John~S. Aberdeen},
  \bibinfo{person}{Elizabeth~Owen Bratt}, \bibinfo{person}{John~S. Garofolo},
  \bibinfo{person}{Helen~Wright Hastie}, \bibinfo{person}{Audrey~N. Le},
  \bibinfo{person}{Bryan~L. Pellom}, \bibinfo{person}{Alexandros Potamianos},
  \bibinfo{person}{Rebecca~J. Passonneau}, \bibinfo{person}{Rashmi Prasad},
  \bibinfo{person}{Salim Roukos}, \bibinfo{person}{Gregory~A. Sanders},
  \bibinfo{person}{Stephanie Seneff}, {and} \bibinfo{person}{David Stallard}.}
  \bibinfo{year}{2002}\natexlab{}.
\newblock \showarticletitle{{DARPA} communicator evaluation: progress from 2000
  to 2001}. In \bibinfo{booktitle}{\emph{{INTERSPEECH}}}.
  \bibinfo{publisher}{{ISCA}}.
\newblock


\bibitem[\protect\citeauthoryear{Weizenbaum}{Weizenbaum}{1966}]%
        {DBLP:journals/cacm/Weizenbaum66}
\bibfield{author}{\bibinfo{person}{Joseph Weizenbaum}.}
  \bibinfo{year}{1966}\natexlab{}.
\newblock \showarticletitle{{ELIZA} - a computer program for the study of
  natural language communication between man and machine}.
\newblock \bibinfo{journal}{\emph{Commun. {ACM}}} \bibinfo{volume}{9},
  \bibinfo{number}{1} (\bibinfo{year}{1966}), \bibinfo{pages}{36--45}.
\newblock


\bibitem[\protect\citeauthoryear{Welbl, Liu, and Gardner}{Welbl
  et~al\mbox{.}}{2017}]%
        {DBLP:conf/aclnut/WelblLG17}
\bibfield{author}{\bibinfo{person}{Johannes Welbl}, \bibinfo{person}{Nelson~F.
  Liu}, {and} \bibinfo{person}{Matt Gardner}.} \bibinfo{year}{2017}\natexlab{}.
\newblock \showarticletitle{Crowdsourcing Multiple Choice Science Questions}.
  In \bibinfo{booktitle}{\emph{NUT@EMNLP}}. \bibinfo{publisher}{Association for
  Computational Linguistics}, \bibinfo{pages}{94--106}.
\newblock


\bibitem[\protect\citeauthoryear{Welbl, Stenetorp, and Riedel}{Welbl
  et~al\mbox{.}}{2018}]%
        {DBLP:journals/tacl/WelblSR18}
\bibfield{author}{\bibinfo{person}{Johannes Welbl}, \bibinfo{person}{Pontus
  Stenetorp}, {and} \bibinfo{person}{Sebastian Riedel}.}
  \bibinfo{year}{2018}\natexlab{}.
\newblock \showarticletitle{Constructing Datasets for Multi-hop Reading
  Comprehension Across Documents}.
\newblock \bibinfo{journal}{\emph{{TACL}}}  \bibinfo{volume}{6}
  (\bibinfo{year}{2018}), \bibinfo{pages}{287--302}.
\newblock


\bibitem[\protect\citeauthoryear{Weston, Bordes, Chopra, and Mikolov}{Weston
  et~al\mbox{.}}{2016}]%
        {DBLP:journals/corr/WestonBCM15}
\bibfield{author}{\bibinfo{person}{Jason Weston}, \bibinfo{person}{Antoine
  Bordes}, \bibinfo{person}{Sumit Chopra}, {and} \bibinfo{person}{Tomas
  Mikolov}.} \bibinfo{year}{2016}\natexlab{}.
\newblock \showarticletitle{Towards AI-Complete Question Answering: {A} Set of
  Prerequisite Toy Tasks}. In \bibinfo{booktitle}{\emph{4th International
  Conference on Learning Representations, {ICLR} 2016, San Juan, Puerto Rico,
  May 2-4, 2016, Conference Track Proceedings}},
  \bibfield{editor}{\bibinfo{person}{Yoshua Bengio} {and} \bibinfo{person}{Yann
  LeCun}} (Eds.).
\newblock
\urldef\tempurl%
\url{http://arxiv.org/abs/1502.05698}
\showURL{%
\tempurl}


\bibitem[\protect\citeauthoryear{Weston, Chopra, and Bordes}{Weston
  et~al\mbox{.}}{2015}]%
        {DBLP:journals/corr/WestonCB14}
\bibfield{author}{\bibinfo{person}{Jason Weston}, \bibinfo{person}{Sumit
  Chopra}, {and} \bibinfo{person}{Antoine Bordes}.}
  \bibinfo{year}{2015}\natexlab{}.
\newblock \showarticletitle{Memory Networks}. In \bibinfo{booktitle}{\emph{3rd
  International Conference on Learning Representations, {ICLR} 2015, San Diego,
  CA, USA, May 7-9, 2015, Conference Track Proceedings}},
  \bibfield{editor}{\bibinfo{person}{Yoshua Bengio} {and} \bibinfo{person}{Yann
  LeCun}} (Eds.).
\newblock
\urldef\tempurl%
\url{http://arxiv.org/abs/1410.3916}
\showURL{%
\tempurl}


\bibitem[\protect\citeauthoryear{Weston, Dinan, and Miller}{Weston
  et~al\mbox{.}}{2018}]%
        {DBLP:conf/emnlp/WestonDM18}
\bibfield{author}{\bibinfo{person}{Jason Weston}, \bibinfo{person}{Emily
  Dinan}, {and} \bibinfo{person}{Alexander~H. Miller}.}
  \bibinfo{year}{2018}\natexlab{}.
\newblock \showarticletitle{Retrieve and Refine: Improved Sequence Generation
  Models For Dialogue}. In \bibinfo{booktitle}{\emph{Proceedings of the 2nd
  International Workshop on Search-Oriented Conversational AI, SCAI@EMNLP 2018,
  Brussels, Belgium, October 31, 2018}},
  \bibfield{editor}{\bibinfo{person}{Aleksandr Chuklin}, \bibinfo{person}{Jeff
  Dalton}, \bibinfo{person}{Julia Kiseleva}, \bibinfo{person}{Alexey Borisov},
  {and} \bibinfo{person}{Mikhail Burtsev}} (Eds.).
  \bibinfo{publisher}{Association for Computational Linguistics},
  \bibinfo{pages}{87--92}.
\newblock
\urldef\tempurl%
\url{https://www.aclweb.org/anthology/W18-5713/}
\showURL{%
\tempurl}


\bibitem[\protect\citeauthoryear{Wu, Martinez, and Klyen}{Wu
  et~al\mbox{.}}{2018}]%
        {DBLP:conf/naacl/WuMK18}
\bibfield{author}{\bibinfo{person}{Xianchao Wu}, \bibinfo{person}{Ander
  Martinez}, {and} \bibinfo{person}{Momo Klyen}.}
  \bibinfo{year}{2018}\natexlab{}.
\newblock \showarticletitle{Dialog Generation Using Multi-Turn Reasoning Neural
  Networks}. In \bibinfo{booktitle}{\emph{{NAACL-HLT}}}.
  \bibinfo{publisher}{Association for Computational Linguistics},
  \bibinfo{pages}{2049--2059}.
\newblock


\bibitem[\protect\citeauthoryear{Xia, Tian, Wu, Lin, Qin, Yu, and Liu}{Xia
  et~al\mbox{.}}{2017}]%
        {DBLP:conf/nips/XiaTWLQYL17}
\bibfield{author}{\bibinfo{person}{Yingce Xia}, \bibinfo{person}{Fei Tian},
  \bibinfo{person}{Lijun Wu}, \bibinfo{person}{Jianxin Lin},
  \bibinfo{person}{Tao Qin}, \bibinfo{person}{Nenghai Yu}, {and}
  \bibinfo{person}{Tie{-}Yan Liu}.} \bibinfo{year}{2017}\natexlab{}.
\newblock \showarticletitle{Deliberation Networks: Sequence Generation Beyond
  One-Pass Decoding}. In \bibinfo{booktitle}{\emph{{NIPS}}}.
  \bibinfo{pages}{1784--1794}.
\newblock


\bibitem[\protect\citeauthoryear{Xu, Liu, Shu, and Yu}{Xu
  et~al\mbox{.}}{2019}]%
        {DBLP:journals/corr/abs-1902-00821}
\bibfield{author}{\bibinfo{person}{Hu Xu}, \bibinfo{person}{Bing Liu},
  \bibinfo{person}{Lei Shu}, {and} \bibinfo{person}{Philip~S. Yu}.}
  \bibinfo{year}{2019}\natexlab{}.
\newblock \showarticletitle{Review Conversational Reading Comprehension}.
\newblock \bibinfo{journal}{\emph{CoRR}}  \bibinfo{volume}{abs/1902.00821}
  (\bibinfo{year}{2019}).
\newblock
\showeprint[arxiv]{1902.00821}
\urldef\tempurl%
\url{http://arxiv.org/abs/1902.00821}
\showURL{%
\tempurl}


\bibitem[\protect\citeauthoryear{Xu, Jiang, Liu, Rong, Wu, Wang, Wang, and
  Wang}{Xu et~al\mbox{.}}{2018}]%
        {DBLP:conf/naacl/XuJLRWWWW18}
\bibfield{author}{\bibinfo{person}{Zhen Xu}, \bibinfo{person}{Nan Jiang},
  \bibinfo{person}{Bingquan Liu}, \bibinfo{person}{Wenge Rong},
  \bibinfo{person}{Bowen Wu}, \bibinfo{person}{Baoxun Wang},
  \bibinfo{person}{Zhuoran Wang}, {and} \bibinfo{person}{Xiaolong Wang}.}
  \bibinfo{year}{2018}\natexlab{}.
\newblock \showarticletitle{{LSDSCC:} a Large Scale Domain-Specific
  Conversational Corpus for Response Generation with Diversity Oriented
  Evaluation Metrics}. In \bibinfo{booktitle}{\emph{{NAACL-HLT}}}.
  \bibinfo{publisher}{Association for Computational Linguistics},
  \bibinfo{pages}{2070--2080}.
\newblock


\bibitem[\protect\citeauthoryear{Yan}{Yan}{2018}]%
        {DBLP:conf/ijcai/Yan18}
\bibfield{author}{\bibinfo{person}{Rui Yan}.} \bibinfo{year}{2018}\natexlab{}.
\newblock \showarticletitle{"Chitty-Chitty-Chat Bot": Deep Learning for
  Conversational {AI}}. In \bibinfo{booktitle}{\emph{{IJCAI}}}.
  \bibinfo{publisher}{ijcai.org}, \bibinfo{pages}{5520--5526}.
\newblock


\bibitem[\protect\citeauthoryear{Yan, Duan, Chen, Zhou, Zhou, and Li}{Yan
  et~al\mbox{.}}{2017}]%
        {DBLP:conf/aaai/YanDCZZL17}
\bibfield{author}{\bibinfo{person}{Zhao Yan}, \bibinfo{person}{Nan Duan},
  \bibinfo{person}{Peng Chen}, \bibinfo{person}{Ming Zhou},
  \bibinfo{person}{Jianshe Zhou}, {and} \bibinfo{person}{Zhoujun Li}.}
  \bibinfo{year}{2017}\natexlab{}.
\newblock \showarticletitle{Building Task-Oriented Dialogue Systems for Online
  Shopping}. In \bibinfo{booktitle}{\emph{{AAAI}}}. \bibinfo{publisher}{{AAAI}
  Press}, \bibinfo{pages}{4618--4626}.
\newblock


\bibitem[\protect\citeauthoryear{Yang, Hu, Qiu, Qu, Gao, Croft, Liu, Shen, and
  Liu}{Yang et~al\mbox{.}}{2019b}]%
        {DBLP:conf/cikm/0005HQQGCLSL19}
\bibfield{author}{\bibinfo{person}{Liu Yang}, \bibinfo{person}{Junjie Hu},
  \bibinfo{person}{Minghui Qiu}, \bibinfo{person}{Chen Qu},
  \bibinfo{person}{Jianfeng Gao}, \bibinfo{person}{W.~Bruce Croft},
  \bibinfo{person}{Xiaodong Liu}, \bibinfo{person}{Yelong Shen}, {and}
  \bibinfo{person}{Jingjing Liu}.} \bibinfo{year}{2019}\natexlab{b}.
\newblock \showarticletitle{A Hybrid Retrieval-Generation Neural Conversation
  Model}. In \bibinfo{booktitle}{\emph{{CIKM}}}. \bibinfo{publisher}{{ACM}},
  \bibinfo{pages}{1341--1350}.
\newblock


\bibitem[\protect\citeauthoryear{Yang, Qiu, Qu, Guo, Zhang, Croft, Huang, and
  Chen}{Yang et~al\mbox{.}}{2018b}]%
        {DBLP:conf/sigir/YangQQGZCHC18}
\bibfield{author}{\bibinfo{person}{Liu Yang}, \bibinfo{person}{Minghui Qiu},
  \bibinfo{person}{Chen Qu}, \bibinfo{person}{Jiafeng Guo},
  \bibinfo{person}{Yongfeng Zhang}, \bibinfo{person}{W.~Bruce Croft},
  \bibinfo{person}{Jun Huang}, {and} \bibinfo{person}{Haiqing Chen}.}
  \bibinfo{year}{2018}\natexlab{b}.
\newblock \showarticletitle{Response Ranking with Deep Matching Networks and
  External Knowledge in Information-seeking Conversation Systems}. In
  \bibinfo{booktitle}{\emph{{SIGIR}}}. \bibinfo{publisher}{{ACM}},
  \bibinfo{pages}{245--254}.
\newblock


\bibitem[\protect\citeauthoryear{Yang, Dai, Yang, Carbonell, Salakhutdinov, and
  Le}{Yang et~al\mbox{.}}{2019a}]%
        {DBLP:journals/corr/abs-1906-08237}
\bibfield{author}{\bibinfo{person}{Zhilin Yang}, \bibinfo{person}{Zihang Dai},
  \bibinfo{person}{Yiming Yang}, \bibinfo{person}{Jaime~G. Carbonell},
  \bibinfo{person}{Ruslan Salakhutdinov}, {and} \bibinfo{person}{Quoc~V. Le}.}
  \bibinfo{year}{2019}\natexlab{a}.
\newblock \showarticletitle{XLNet: Generalized Autoregressive Pretraining for
  Language Understanding}.
\newblock \bibinfo{journal}{\emph{CoRR}}  \bibinfo{volume}{abs/1906.08237}
  (\bibinfo{year}{2019}).
\newblock
\showeprint[arxiv]{1906.08237}
\urldef\tempurl%
\url{http://arxiv.org/abs/1906.08237}
\showURL{%
\tempurl}


\bibitem[\protect\citeauthoryear{Yang, Qi, Zhang, Bengio, Cohen, Salakhutdinov,
  and Manning}{Yang et~al\mbox{.}}{2018a}]%
        {DBLP:conf/emnlp/Yang0ZBCSM18}
\bibfield{author}{\bibinfo{person}{Zhilin Yang}, \bibinfo{person}{Peng Qi},
  \bibinfo{person}{Saizheng Zhang}, \bibinfo{person}{Yoshua Bengio},
  \bibinfo{person}{William~W. Cohen}, \bibinfo{person}{Ruslan Salakhutdinov},
  {and} \bibinfo{person}{Christopher~D. Manning}.}
  \bibinfo{year}{2018}\natexlab{a}.
\newblock \showarticletitle{HotpotQA: {A} Dataset for Diverse, Explainable
  Multi-hop Question Answering}. In \bibinfo{booktitle}{\emph{{EMNLP}}}.
  \bibinfo{publisher}{Association for Computational Linguistics},
  \bibinfo{pages}{2369--2380}.
\newblock


\bibitem[\protect\citeauthoryear{Yatskar}{Yatskar}{2019}]%
        {DBLP:conf/naacl/Yatskar19}
\bibfield{author}{\bibinfo{person}{Mark Yatskar}.}
  \bibinfo{year}{2019}\natexlab{}.
\newblock \showarticletitle{A Qualitative Comparison of CoQA, SQuAD 2.0 and
  QuAC}. In \bibinfo{booktitle}{\emph{{NAACL-HLT} {(1)}}}.
  \bibinfo{publisher}{Association for Computational Linguistics},
  \bibinfo{pages}{2318--2323}.
\newblock


\bibitem[\protect\citeauthoryear{Yeh and Chen}{Yeh and Chen}{2019}]%
        {DBLP:journals/corr/abs-1908-05117}
\bibfield{author}{\bibinfo{person}{Yi~Ting Yeh} {and}
  \bibinfo{person}{Yun{-}Nung Chen}.} \bibinfo{year}{2019}\natexlab{}.
\newblock \showarticletitle{FlowDelta: Modeling Flow Information Gain in
  Reasoning for Conversational Machine Comprehension}.
\newblock \bibinfo{journal}{\emph{CoRR}}  \bibinfo{volume}{abs/1908.05117}
  (\bibinfo{year}{2019}).
\newblock
\showeprint[arxiv]{1908.05117}
\urldef\tempurl%
\url{http://arxiv.org/abs/1908.05117}
\showURL{%
\tempurl}


\bibitem[\protect\citeauthoryear{Young, Cambria, Chaturvedi, Zhou, Biswas, and
  Huang}{Young et~al\mbox{.}}{2018}]%
        {DBLP:conf/aaai/YoungCCZBH18}
\bibfield{author}{\bibinfo{person}{Tom Young}, \bibinfo{person}{Erik Cambria},
  \bibinfo{person}{Iti Chaturvedi}, \bibinfo{person}{Hao Zhou},
  \bibinfo{person}{Subham Biswas}, {and} \bibinfo{person}{Minlie Huang}.}
  \bibinfo{year}{2018}\natexlab{}.
\newblock \showarticletitle{Augmenting End-to-End Dialogue Systems With
  Commonsense Knowledge}. In \bibinfo{booktitle}{\emph{{AAAI}}}.
  \bibinfo{publisher}{{AAAI} Press}, \bibinfo{pages}{4970--4977}.
\newblock


\bibitem[\protect\citeauthoryear{Zhang, Dinan, Urbanek, Szlam, Kiela, and
  Weston}{Zhang et~al\mbox{.}}{2018a}]%
        {DBLP:conf/acl/KielaWZDUS18}
\bibfield{author}{\bibinfo{person}{Saizheng Zhang}, \bibinfo{person}{Emily
  Dinan}, \bibinfo{person}{Jack Urbanek}, \bibinfo{person}{Arthur Szlam},
  \bibinfo{person}{Douwe Kiela}, {and} \bibinfo{person}{Jason Weston}.}
  \bibinfo{year}{2018}\natexlab{a}.
\newblock \showarticletitle{Personalizing Dialogue Agents: {I} have a dog, do
  you have pets too?}. In \bibinfo{booktitle}{\emph{{ACL} {(1)}}}.
  \bibinfo{publisher}{Association for Computational Linguistics},
  \bibinfo{pages}{2204--2213}.
\newblock


\bibitem[\protect\citeauthoryear{Zhang, Yang, Li, and Wang}{Zhang
  et~al\mbox{.}}{2019b}]%
        {DBLP:journals/corr/abs-1907-01686}
\bibfield{author}{\bibinfo{person}{Xin Zhang}, \bibinfo{person}{An Yang},
  \bibinfo{person}{Sujian Li}, {and} \bibinfo{person}{Yizhong Wang}.}
  \bibinfo{year}{2019}\natexlab{b}.
\newblock \showarticletitle{Machine Reading Comprehension: a Literature
  Review}.
\newblock \bibinfo{journal}{\emph{CoRR}}  \bibinfo{volume}{abs/1907.01686}
  (\bibinfo{year}{2019}).
\newblock
\showeprint[arxiv]{1907.01686}
\urldef\tempurl%
\url{http://arxiv.org/abs/1907.01686}
\showURL{%
\tempurl}


\bibitem[\protect\citeauthoryear{Zhang, Galley, Gao, Gan, Li, Brockett, and
  Dolan}{Zhang et~al\mbox{.}}{2018b}]%
        {DBLP:conf/nips/ZhangGGGLBD18}
\bibfield{author}{\bibinfo{person}{Yizhe Zhang}, \bibinfo{person}{Michel
  Galley}, \bibinfo{person}{Jianfeng Gao}, \bibinfo{person}{Zhe Gan},
  \bibinfo{person}{Xiujun Li}, \bibinfo{person}{Chris Brockett}, {and}
  \bibinfo{person}{Bill Dolan}.} \bibinfo{year}{2018}\natexlab{b}.
\newblock \showarticletitle{Generating Informative and Diverse Conversational
  Responses via Adversarial Information Maximization}. In
  \bibinfo{booktitle}{\emph{NeurIPS}}. \bibinfo{pages}{1815--1825}.
\newblock


\bibitem[\protect\citeauthoryear{Zhang, Ren, and de~Rijke}{Zhang
  et~al\mbox{.}}{2019a}]%
        {DBLP:journals/corr/abs-1906-06685}
\bibfield{author}{\bibinfo{person}{Yangjun Zhang}, \bibinfo{person}{Pengjie
  Ren}, {and} \bibinfo{person}{Maarten de Rijke}.}
  \bibinfo{year}{2019}\natexlab{a}.
\newblock \showarticletitle{Improving Background Based Conversation with
  Context-aware Knowledge Pre-selection}.
\newblock \bibinfo{journal}{\emph{CoRR}}  \bibinfo{volume}{abs/1906.06685}
  (\bibinfo{year}{2019}).
\newblock
\showeprint[arxiv]{1906.06685}
\urldef\tempurl%
\url{http://arxiv.org/abs/1906.06685}
\showURL{%
\tempurl}


\bibitem[\protect\citeauthoryear{Zhang, Li, Zhu, Zhao, and Liu}{Zhang
  et~al\mbox{.}}{2018c}]%
        {DBLP:conf/coling/ZhangLZZL18}
\bibfield{author}{\bibinfo{person}{Zhuosheng Zhang}, \bibinfo{person}{Jiangtong
  Li}, \bibinfo{person}{Pengfei Zhu}, \bibinfo{person}{Hai Zhao}, {and}
  \bibinfo{person}{Gongshen Liu}.} \bibinfo{year}{2018}\natexlab{c}.
\newblock \showarticletitle{Modeling Multi-turn Conversation with Deep
  Utterance Aggregation}. In \bibinfo{booktitle}{\emph{{COLING}}}.
  \bibinfo{publisher}{Association for Computational Linguistics},
  \bibinfo{pages}{3740--3752}.
\newblock


\bibitem[\protect\citeauthoryear{Zhao, Lu, Lee, and Esk{\'{e}}nazi}{Zhao
  et~al\mbox{.}}{2017}]%
        {DBLP:conf/sigdial/ZhaoLLE17}
\bibfield{author}{\bibinfo{person}{Tiancheng Zhao}, \bibinfo{person}{Allen Lu},
  \bibinfo{person}{Kyusong Lee}, {and} \bibinfo{person}{Maxine
  Esk{\'{e}}nazi}.} \bibinfo{year}{2017}\natexlab{}.
\newblock \showarticletitle{Generative Encoder-Decoder Models for Task-Oriented
  Spoken Dialog Systems with Chatting Capability}. In
  \bibinfo{booktitle}{\emph{{SIGDIAL} Conference}}.
  \bibinfo{publisher}{Association for Computational Linguistics},
  \bibinfo{pages}{27--36}.
\newblock


\bibitem[\protect\citeauthoryear{Zhao, Tao, Wu, Xu, Zhao, and Yan}{Zhao
  et~al\mbox{.}}{2019}]%
        {DBLP:conf/ijcai/ZhaoTWX0Y19}
\bibfield{author}{\bibinfo{person}{Xueliang Zhao}, \bibinfo{person}{Chongyang
  Tao}, \bibinfo{person}{Wei Wu}, \bibinfo{person}{Can Xu},
  \bibinfo{person}{Dongyan Zhao}, {and} \bibinfo{person}{Rui Yan}.}
  \bibinfo{year}{2019}\natexlab{}.
\newblock \showarticletitle{A Document-grounded Matching Network for Response
  Selection in Retrieval-based Chatbots}. In
  \bibinfo{booktitle}{\emph{{IJCAI}}}. \bibinfo{publisher}{ijcai.org},
  \bibinfo{pages}{5443--5449}.
\newblock


\bibitem[\protect\citeauthoryear{Zhong and Zettlemoyer}{Zhong and
  Zettlemoyer}{2019}]%
        {DBLP:conf/acl/ZhongZ19}
\bibfield{author}{\bibinfo{person}{Victor Zhong} {and} \bibinfo{person}{Luke
  Zettlemoyer}.} \bibinfo{year}{2019}\natexlab{}.
\newblock \showarticletitle{{E3:} Entailment-driven Extracting and Editing for
  Conversational Machine Reading}. In \bibinfo{booktitle}{\emph{{ACL} {(1)}}}.
  \bibinfo{publisher}{Association for Computational Linguistics},
  \bibinfo{pages}{2310--2320}.
\newblock


\bibitem[\protect\citeauthoryear{Zhou, Young, Huang, Zhao, Xu, and Zhu}{Zhou
  et~al\mbox{.}}{2018d}]%
        {DBLP:conf/ijcai/ZhouYHZXZ18}
\bibfield{author}{\bibinfo{person}{Hao Zhou}, \bibinfo{person}{Tom Young},
  \bibinfo{person}{Minlie Huang}, \bibinfo{person}{Haizhou Zhao},
  \bibinfo{person}{Jingfang Xu}, {and} \bibinfo{person}{Xiaoyan Zhu}.}
  \bibinfo{year}{2018}\natexlab{d}.
\newblock \showarticletitle{Commonsense Knowledge Aware Conversation Generation
  with Graph Attention}. In \bibinfo{booktitle}{\emph{{IJCAI}}}.
  \bibinfo{publisher}{ijcai.org}, \bibinfo{pages}{4623--4629}.
\newblock


\bibitem[\protect\citeauthoryear{Zhou, Prabhumoye, and Black}{Zhou
  et~al\mbox{.}}{2018c}]%
        {DBLP:conf/emnlp/ZhouPB18}
\bibfield{author}{\bibinfo{person}{Kangyan Zhou}, \bibinfo{person}{Shrimai
  Prabhumoye}, {and} \bibinfo{person}{Alan~W. Black}.}
  \bibinfo{year}{2018}\natexlab{c}.
\newblock \showarticletitle{A Dataset for Document Grounded Conversations}. In
  \bibinfo{booktitle}{\emph{{EMNLP}}}. \bibinfo{publisher}{Association for
  Computational Linguistics}, \bibinfo{pages}{708--713}.
\newblock


\bibitem[\protect\citeauthoryear{Zhou, Gao, Li, and Shum}{Zhou
  et~al\mbox{.}}{2018a}]%
        {DBLP:journals/corr/abs-1812-08989}
\bibfield{author}{\bibinfo{person}{Li Zhou}, \bibinfo{person}{Jianfeng Gao},
  \bibinfo{person}{Di Li}, {and} \bibinfo{person}{Heung{-}Yeung Shum}.}
  \bibinfo{year}{2018}\natexlab{a}.
\newblock \showarticletitle{The Design and Implementation of XiaoIce, an
  Empathetic Social Chatbot}.
\newblock \bibinfo{journal}{\emph{CoRR}}  \bibinfo{volume}{abs/1812.08989}
  (\bibinfo{year}{2018}).
\newblock
\showeprint[arxiv]{1812.08989}
\urldef\tempurl%
\url{http://arxiv.org/abs/1812.08989}
\showURL{%
\tempurl}


\bibitem[\protect\citeauthoryear{Zhou, Li, Dong, Liu, Chen, Zhao, Yu, and
  Wu}{Zhou et~al\mbox{.}}{2018b}]%
        {DBLP:conf/acl/WuLCZDYZL18}
\bibfield{author}{\bibinfo{person}{Xiangyang Zhou}, \bibinfo{person}{Lu Li},
  \bibinfo{person}{Daxiang Dong}, \bibinfo{person}{Yi Liu},
  \bibinfo{person}{Ying Chen}, \bibinfo{person}{Wayne~Xin Zhao},
  \bibinfo{person}{Dianhai Yu}, {and} \bibinfo{person}{Hua Wu}.}
  \bibinfo{year}{2018}\natexlab{b}.
\newblock \showarticletitle{Multi-Turn Response Selection for Chatbots with
  Deep Attention Matching Network}. In \bibinfo{booktitle}{\emph{{ACL} {(1)}}}.
  \bibinfo{publisher}{Association for Computational Linguistics},
  \bibinfo{pages}{1118--1127}.
\newblock


\bibitem[\protect\citeauthoryear{Zhu, Zeng, and Huang}{Zhu
  et~al\mbox{.}}{2018}]%
        {DBLP:journals/corr/abs-1812-03593}
\bibfield{author}{\bibinfo{person}{Chenguang Zhu}, \bibinfo{person}{Michael
  Zeng}, {and} \bibinfo{person}{Xuedong Huang}.}
  \bibinfo{year}{2018}\natexlab{}.
\newblock \showarticletitle{SDNet: Contextualized Attention-based Deep Network
  for Conversational Question Answering}.
\newblock \bibinfo{journal}{\emph{CoRR}}  \bibinfo{volume}{abs/1812.03593}
  (\bibinfo{year}{2018}).
\newblock
\showeprint[arxiv]{1812.03593}
\urldef\tempurl%
\url{http://arxiv.org/abs/1812.03593}
\showURL{%
\tempurl}


\end{thebibliography}

\appendix

\end{document}